%% file: iclr2024_conference.tex
\documentclass{article} 
\usepackage{iclr2024_conference}
\usepackage{times}

\input{math_commands.tex}

\usepackage{graphicx}
\usepackage{array}
\usepackage{booktabs}
\usepackage{graphicx}
\usepackage{subcaption}
\usepackage{tabularx}
\usepackage{algorithm}
\usepackage{tabularx}
\usepackage{arydshln}
\usepackage{longtable}
\usepackage{tabularx}
\usepackage{multirow}
\usepackage{subcaption}
\usepackage{array} 
\usepackage{algpseudocode}
\usepackage{amsmath}
\usepackage{wrapfig}
\usepackage[utf8]{inputenc} 
\usepackage[T1]{fontenc}    
\usepackage{hyperref}       
\usepackage{url}            
\usepackage{booktabs}       
\usepackage{amsfonts}       
\usepackage{nicefrac}       
\usepackage{microtype}      
\usepackage{xcolor}         
\usepackage{enumitem}
\usepackage{listings}

\lstdefinelanguage{SPARQL}{
  morekeywords={BASE,PREFIX,SELECT,WHERE,FILTER,OPTIONAL,DISTINCT,GRAPH,UNION,ASK,CONSTRUCT,DESCRIBE,FROM,NAMED,ORDER,BY,ASC,DESC,LIMIT,OFFSET,BIND,VALUES},
  sensitive=false,
  morecomment=[l]{\#},
  morestring=[b]",
}

\title{Think-on-Graph: Deep and Responsible Reasoning of Large Language Model on Knowledge Graph}

\author{Jiashuo Sun$^{21}$\thanks{Equal contribution.}~\thanks{Work done during internship at IDEA Research.}~~Chengjin Xu$^{1}$\footnotemark[1]~~Lumingyuan Tang$^{31}$\footnotemark[1]~\footnotemark[2]~~Saizhuo Wang$^{41}$\footnotemark[1]~\footnotemark[2]  \\ \bf Chen Lin$^{2}$~~Yeyun Gong$^{6}$~~Lionel M. Ni$^{5}$~Heung-Yeung Shum$^{14}$~Jian Guo$^{15}$\thanks{Corresponding author.}\\
$^1$IDEA Research, International Digital Economy Academy \\
$^2$Xiamen University \\ $^3$University of Southern California \\
$^4$The Hong Kong University of Science and Technology \\
$^5$The Hong Kong University of Science and Technology (Guangzhou) \\
$^6$Microsoft Research Asia \\}

\iclrfinalcopy 
\begin{document}

\maketitle

\begin{abstract}
\label{sec: abstract}
Although large language models (LLMs) have achieved significant success in various tasks, they often struggle with hallucination problems, especially in scenarios requiring deep and responsible reasoning. These issues could be partially addressed by introducing external knowledge graphs (KG) in LLM reasoning. In this paper, we propose a new LLM-KG integrating paradigm ``$\hbox{LLM}\otimes\hbox{KG}$'' which treats the LLM as an agent to interactively explore related entities and relations on KGs and perform reasoning based on the retrieved knowledge. We further implement this paradigm by introducing a new approach called Think-on-Graph (ToG), in which the LLM agent iteratively executes beam search on KG, discovers the most promising reasoning paths, and returns the most likely reasoning results. We use a number of well-designed experiments to examine and illustrate the following advantages of ToG: 1) compared with LLMs, ToG has better deep reasoning power; 2) ToG has the ability of knowledge traceability and knowledge correctability by leveraging LLMs reasoning and expert feedback; 3) ToG provides a flexible plug-and-play framework for different LLMs, KGs and prompting strategies without any additional training cost; 4) the performance of ToG with small LLM models could exceed large LLM such as GPT-4 in certain scenarios and this reduces the cost of LLM deployment and application. As a training-free method with lower computational cost and better generality, ToG achieves overall SOTA in 6 out of 9 datasets where most previous SOTAs rely on additional training. Our code is publicly available at \url{https://github.com/IDEA-FinAI/ToG}.

\end{abstract}

\section{Introduction}
\label{sec: introduction}
\begin{figure}[t]
	\centering
	\includegraphics[width=0.95\columnwidth]{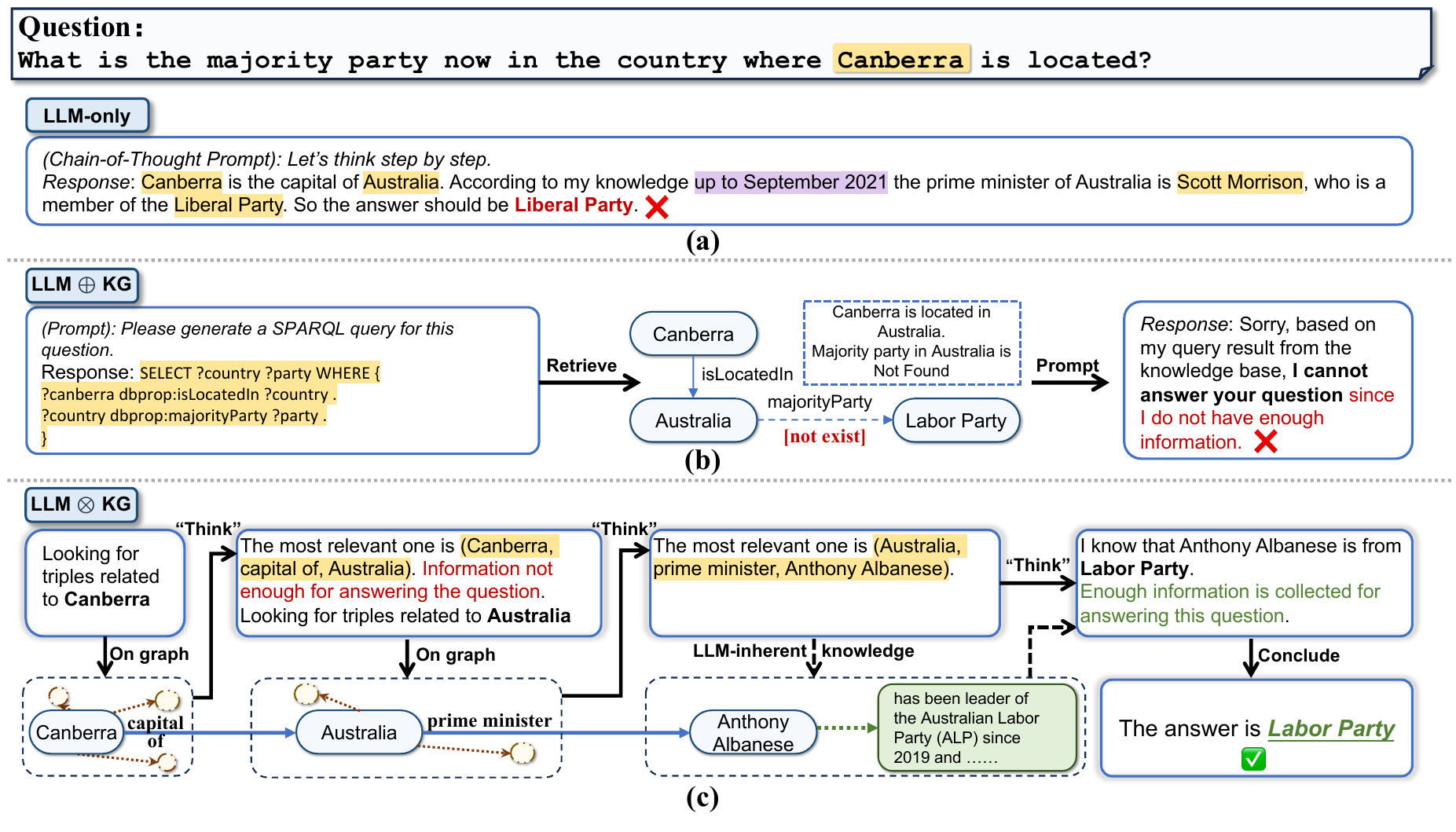}
 \caption{Representative workflow of three LLM reasoning paradigms: (a) LLM-only (e.g., Chain-of-Thought prompting), (b) LLM $\oplus$ KG (e.g., KBQA via LLM-generated SPARQL query), (c) LLM $\otimes$ KG (e.g., Think-on-Graph).}
         \vspace{-10pt}
	\label{fig: example}
\end{figure}

Large language models (LLMs) \citep{gpt3.5, gpt4, lamda, fewshotlearner, palm, llama2} have demonstrated remarkable performance across various natural language processing tasks. These models capitalize on pre-training techniques applied to vast text corpora to generate responses that are coherent and contextually appropriate. 
Despite their impressive performance, LLMs have substantial limitations when facing complex knowledge reasoning tasks \citep{kilt, commonsenseqa, cwq_dataset, zhang2023noisy} that require deep and responsible reasoning. Firstly, LLMs usually fail to provide accurate answers to questions requiring specialized knowledge beyond what was included in the pre-training phase (out-of-date knowledge in Figure \ref{fig: example}a), or to questions requiring long logic chain and multi-hop knowledge reasoning. Secondly, LLMs lack responsibility, explainability and transparency, raising concerns about the risk of hallucinations or toxic texts. Thirdly, the training process for LLMs is often expensive and time-consuming, making it challenging to keep their knowledge up to date.


Recognizing these challenges, a natural and promising solution is to incorporate external knowledge such as knowledge graphs (KGs) to help improve LLM reasoning.
KGs offer structured, explicit, and editable representations of knowledge, presenting a complementary strategy to mitigate the limitations of LLMs \citep{pan2023unifying}.
Researchers \citep{cok, skg, knowledgeaugmented, yang2023chatgpt, wang2023boosting, structgpt} have explored the usage of KGs as external knowledge sources to mitigate hallucination in LLMs. These approaches follow a routine: retrieve information from KGs, augment the prompt accordingly, and feed the increased prompt into LLMs (as illustrated in Figure \ref{fig: example}b). In this paper, we refer to this paradigm as ``$\hbox{LLM}\oplus\hbox{KG}$''.
Although aiming to integrate the power of LLM and KG, in this paradigm, LLM plays the role of translator which transfers input questions to machine-understandable command for KG searching and reasoning, but it does not participate in the graph reasoning process directly. Unfortunately, the loose-coupling $\hbox{LLM}\oplus\hbox{KG}$ paradigm has its own limitations, and its success depends heavily on the completeness and high quality of KG. In Figure \ref{fig: example}b, for example, although LLM successfully identified necessary relation types required to answer the question, the absence of the relation ``majority party'' leads to a failure in retrieving the correct answer.

Building upon these considerations, we propose a new tight-coupling ``$\hbox{LLM} \otimes \hbox{KG}$'' paradigm where KGs and LLMs work in tandem, complementing each other's capabilities in each step of graph reasoning.
Figure \ref{fig: example}c provides an example illustrating the advantage of $\hbox{LLM} \otimes \hbox{KG}$. In this example, the missing relation "majority party" resulting in the failure in Figure \ref{fig: example}b can be complemented by a reference triple $(\textbf{Australia},\textbf{prime minister},\textbf{Anthony Albanese})$ discovered by the LLM agent with dynamic reasoning ability \citep{yao2022react}, as well as the political party membership of \textbf{Anthony Albanese} coming from LLM's inherent knowledge.
In this way, the LLM succeeds in generating the correct answer with reliable knowledge retrieved from KGs.
As an implementation of this paradigm, we propose an algorithmic framework ``Think-on-Graph'' (meaning: LLMs ``Think'' along the reasoning paths ``on'' knowledge ``graph'' step-by-step, abbreviated as ToG below),
for deep, responsible, and efficient LLM reasoning.
Using the beam search algorithm \citep{beamsearch} in KG/LLM reasoning \citep{kg_beam, beamsearchqa, self_eval, liu2024what}, ToG allows LLM to dynamically explore a number of reasoning paths in KG and make decisions accordingly.
Given an input question, ToG first identifies initial entities and then iteratively calls the LLM to retrieve relevant triples from KGs through exploration (looking for relevant triples in KG via ``on graph'' step) and reasoning (deciding on the most relevant triples via ``think'' step) until adequate information through the top-N reasoning paths in beam search is gathered to answer the question (judged by LLMs in "Think" step) or the predefined maximum search depth is reached.

The advantage of ToG can be abbreviated as
(1) \textbf{Deep reasoning:}
ToG extracts diverse and multi-hop reasoning paths from KGs as the basis for LLM reasoning, enhancing LLMs' deep reasoning capabilities for knowledge-intensive tasks.
(2) \textbf{Responsible reasoning:}
Explicit, editable reasoning paths improve the explainability of the reasoning process of LLMs, and enable the tracing and correction of the provenances of models' outputs.
(3) \textbf{Flexibility and efficiency:}
a) ToG is a plug-and-play framework that can be applied to a variety of LLMs and KGs seamlessly.
b) Under ToG framework, knowledge can be updated frequently via KG instead of LLM whose knowledge-update is expensive and slow.
c) ToG enhances the reasoning ability of small LLMs (e.g., LLAMA2-70B) to be competitive with big LLMs (e.g., GPT-4).

\begin{figure}[t]
	\centering
	\includegraphics[width=0.95\columnwidth]{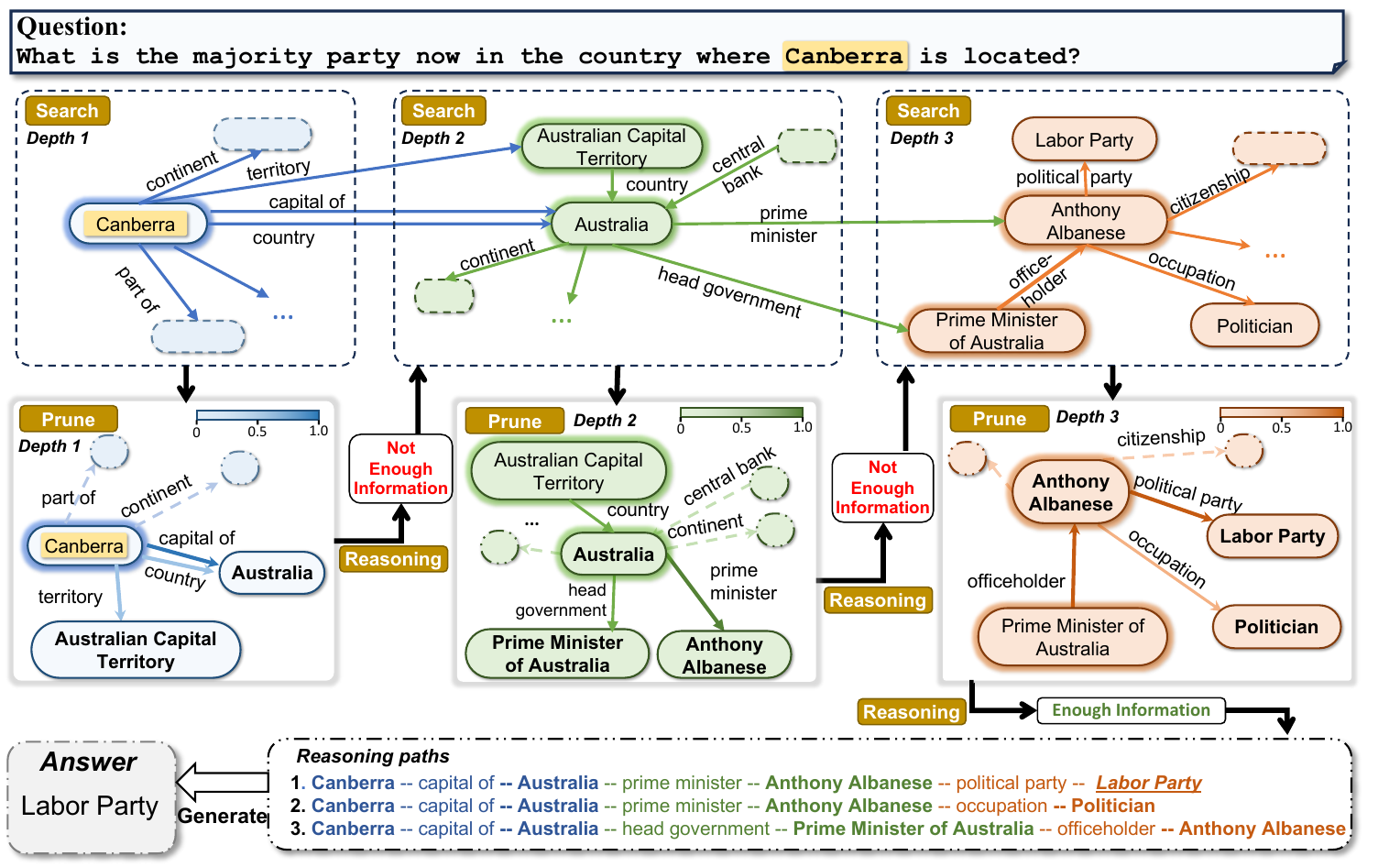}
	\caption{An example workflow of ToG. The glowing entities are the central entities where the search starts at each iteration (depth), and the entities with \textbf{boldface} are the selected central entities for the next iteration after pruning. At each pruning step, the darkness of the edges represents the ranking score given by LLM, and the dashed lines indicate relations that have been pruned due to low evaluation scores.}
	\label{fig: framework}
\end{figure}

\section{Methods}
\label{sec: methods} 
ToG implements the ``$\hbox{LLM}\otimes \hbox{KG}$'' paradigm by asking LLM to perform beam search on knowledge graph. Specifically, it prompts the LLM to iteratively explore multiple possible reasoning paths on KGs until the LLM determines that the question can be answered based on the current reasoning paths. ToG constantly updates and maintains top-$N$ reasoning paths $P=\{p_1,p_2,\ldots, p_N\}$ for the question $x$ after each iteration, where $N$ denotes the width of beam search.
The entire inference process of ToG contains the following 3 phases: initialization, exploration, and reasoning. 

\subsection{Think-on-Graph}

\subsubsection{Initialization of Graph Search}
Given a question, ToG leverages the underlying LLM to localize the initial entity of the reasoning paths on knowledge graph. This phase can be regarded as the initialization of the top-$N$ reasoning paths $P$. ToG first prompts LLMs to automatically extract the topic entities in question and gets the top-$N$ topic entities $E^{0}=\{e^{0}_{1},e^{0}_{2},...,e^{0}_{N}\}$ to the question. Note that the number of topic entities might possibly be less than $N$.

\subsubsection{Exploration}
At the beginning of the $D$-th iteration, each path $p_n$ consists of $D-1$ triples, i.e., $p_n =\{(e^d_{s,n}, r^d_{j,n}, e^d_{o,n})\}^{D-1}_{d=1}$, where $e^d_{s,n}$ and $e^d_{o,n}$ denote subject and object entities, $r^d_{j,n}$ is a specific relation between them, $(e^d_{s,n}, r^d_{j,n}, e^d_{o,n})$ and $(e^{d+1}_{s,n}, r^{d+1}_{j,n}, e^{d+1}_{o,n})$ are connected to each other. The sets of the tail entities and relations in $P$ are denoted as $E^{D-1}=\{e^{D-1}_{1},e^{D-1}_{2},...,e^{D-1}_{N}\}$ and $R^{D-1}=\{r^{D-1}_{1},r^{D-1}_{2},...,r^{D-1}_{N}\}$, respectively. 

The exploration phase in the $D$-th iteration aims to exploit the LLM to identify the most relevant top-$N$ entities $E^{D}$ from the neighboring entities of the current top-$N$ entity set $E^{D-1}$ based on the question $x$ and extend the top-$N$ reasoning paths $P$ with $E^{D}$.
To address the complexity of handling numerous neighboring entities with the LLM, we implement a two-step exploration strategy: first, exploring significant relations, and then using selected relations to guide entity exploration.


\paragraph{Relation Exploration}
\label{sec: relation_exploration}
Relation exploration is a beam search process with the depth of 1 and the width of $N$ from $E^{D-1}$ to $R^{D}$. 
The whole process can be decomposed into two steps: \texttt{Search} and \texttt{Prune}. The LLM serves as an agent to automatically complete this process.
\begin{itemize}[leftmargin=*]

    \item \textbf{Search} 
    At the beginning of the $D$-th iteration, the relation exploration phase first searches out relations $R^{D}_{cand,n}$ linked to the tail entity $e^{D-1}_{n}$ for each reasoning path $p_n$. These relations are aggregated into $R^{D}_{cand}$. In the case of Figure~\ref{fig: framework}, $E^{1}=\{\textbf{Canberra}\}$ and $R^{1}_{cand}$ denotes the set of all relations linked to \textbf{Canberra} inwards or outwards. Notably, the \texttt{Search} procedure can be easily completed by executing two simple pre-defined formal queries shown in Appendix~\ref{sparqls} and~\ref{APIs}, which makes ToG adapt well to different KGs \textbf{without any training cost}.
    
    \item \textbf{Prune} Once we have obtained the candidate relation sets $R^{D}_{cand}$ and the expanded candidate reasoning paths $P_{cand}$ from the relation search, we can utilize the LLM to select out new top-$N$ reasoning paths $P$ ending with the tail relations $R^{D}$ from $P_{cand}$ based on the literal information of the question $x$ and the candidate relations $R^{D}_{cand}$. The prompt used here can be found in Appendix~\ref{prompts4rel}. 
    As shown in Figure~\ref{fig: framework}, the LLM selects top-3 relations $\{\textbf{capital of}, \textbf{country}, \textbf{territory}\}$ out from all relations linked to the entity \textbf{Canberra} in the first iteration. Since \textbf{Canberra} is the only topic entity, the top-3 candidate reasoning paths are updated as $\{(\textbf{Canberra, capital of})$, $(\textbf{Canberra, country})$,$(\textbf{Canberra, territory})\}$.

\end{itemize}

\paragraph{Entity Exploration}
\label{sec: entity_exploration}
Similar to relationship exploration, entity exploration is also a beam search process performed by the LLM from $R^{D}$ to $E^{D}$, and consists of two steps, \texttt{Search} and \texttt{Prune}.
\begin{itemize}[leftmargin=*]
    \item \textbf{Search} Once we have obtained new top-$N$ reasoning paths $P$ and the set of new tail relations $R^{D}$ from relation exploration, for each relation path $p_n\in P$, we can explore a candidate entity set $E^{D}_{cand,n}$ by querying $(e^{D-1}_n, r^{D}_n,?)$ or $(?, r^{D}_n,e^{D-1}_n)$, where $e^{D-1}_n, r_n$ denote the tail entity and relation of $p_n$. We can aggregate $\{E^{D}_{cand,1},E^{D}_{cand,2},...,E^{D}_{cand,N}\}$ into $E^{D}_{cand}$ and expand top-$N$ reasoning paths $P$ to $P_{cand}$ with the tail entities $E^{D}_{cand}$.
    For the shown case, $E^{1}_{cand}$ can be represented as  $\{\textbf{Australia},\textbf{Australia}, \textbf{Australian Capital Territory}\}$.
    \item \textbf{Prune} Since the entities in each candidate set $E^{D}_{cand}$ is expressed in natural language, we can leverage the LLM to select new top-$N$ reasoning paths $P$ ending with the tail entities $E^{D}$ out from $P_{cand}$. The prompt used here can be found in Appendix~\ref{prompts4ent}. As shown in Figure~\ref{fig: framework}, \textbf{Australia} and \textbf{Australian Capital Territory} are scored as 1 since the relations \textbf{capital of}, \textbf{country} and \textbf{territory} are only linked to one tail entity respectively, and the current reasoning paths $p$ are updated as $\{(\textbf{Canberra, capital of, Australia}),(\textbf{Canberra, country, Australia})$, $(\textbf{Canberra, territory, Australian Capital Territory})\}$.
    
\end{itemize}
After executing the two explorations described above, we reconstruct new top-$N$ reasoning paths $P$ where the length of each path increases by 1. Each prune step requires at most $N$ LLM calls.
\subsubsection{Reasoning}
\label{sec: reasoning}
Upon obtaining the current reasoning path $P$ through the exploration process, we prompt the LLM to evaluate whether the current reasoning paths are adequate for generating the answer. If the evaluation yields a positive result, we prompt the LLM to generate the answer using the reasoning paths with the query as inputs as illustrated in Figure~\ref{fig: framework}. The prompt used for evaluation and generation can be found in Appendix~\ref{prompts4rea} and~\ref{prompts4gen}. Conversely, if the evaluation yields a negative result, we repeat the \texttt{Exploration} and \texttt{Reasoning} steps until the evaluation is positive or reaches the maximum search depth $D_{max}$. If the algorithm has not yet concluded, it signifies that even upon reaching the $D_{max}$, ToG remains unable to explore the reasoning paths to resolve the question. In such a scenario, ToG generates the answer exclusively based on the inherent knowledge in the LLM.
The whole inference process of ToG contains $D$ exploration phases and $D$ evaluation steps as well as a generation step, which needs at most $2ND+D+1$ calls to the LLM.

\subsection{Relation-based Think-on-Graph}

Previous KBQA methods, particularly based on semantic parsing, have predominantly relied on relation information in questions to generate formal queries~\citep{complexKBQA}. Inspired by this, we propose relation-based ToG (ToG-R) that explores the top-$N$ relation chains $\{p_n = (e^{0}_n,r^{1}_n,r^{2}_n,...,r^{D}_n)\}^N_{n=1}$ starting with the topic entities $\{e^{0}_n\}^N_{n=1}$ instead of triple-based reasoning paths. ToG-R sequentially performs \text{relation search}, \text{relation prune} and \text{entity search} in each iteration, which is the same as ToG. Then ToG-R performs the \text{reasoning} step based on all candidate reasoning paths ending with $E^{D}_{cand}$ obtained by entity search. If the LLM determines that the retrieved candidate reasoning paths do not contain enough information for the LLM to answer the question, we randomly sample N entities from the candidate entities $E^{D}_{cand}$ and continue to the next iteration. Assuming that entities in each entity set $E^{D}_{cand,n}$ probably belong to the same entity class and have similar neighboring relations, the results of pruning the entity set $\{E^{D}_{cand,n}\}^N_{n=1}$ might have little impact on the following relation exploration. Thus, we use the random beam search instead of the LLM-constrained beam search in ToG for \text{entity prune}, referred to as \textbf{random prune}. Algorithm \ref{algorithm1} and \ref{algorithm2} show the implementation details of the ToG and ToG-R. ToG-R needs at most $ND+D+1$ calls to the LLM.

Compared to ToG, ToG-R offers two key benefits: 1) It eliminates the need for the process of pruning entities using the LLM, thereby reducing the overall cost and reasoning time. 2) ToG-R primarily emphasizes the literal information of relations, mitigating the risk of misguided reasoning when the literal information of intermediate entities is missing or unfamiliar to the LLM.

\section{Experiments}
\label{sec: experiments}
\subsection{Experimental Design}
\subsubsection{Datasets and Evaluation Metrics}
\label{sec: datasets}
In order to test ToG's ability on multi-hop knowledge-intensive reasoning tasks, we evaluate ToG on five KBQA datasets (4 Multi-hop and 1 Single-hop): CWQ \citep{cwq_dataset}, WebQSP \citep{webqsp_dataset}, GrailQA \citep{garilqa_dataset}, QALD10-en \citep{qald10-en_dataset}, Simple Questions \citep{simple_questions_dataset}. Moreover, in order to examine ToG on more generic tasks, we also prepare one open-domain QA dataset: WebQuestions \citep{webq_dataset}; two slot filling datasets: T-REx \citep{trex_dataset} and Zero-Shot RE \citep{kilt}; and one fact-checking dataset: Creak \citep{creak_dataset}. 
Note that, for two big datasets GrailQA and Simple Questions, we only randomly selected 1,000 samples each for testing in order to save computational cost. 
For all datasets, exact match accuracy (Hits@1) is used as our evaluation metric following previous works \citep{cok, knowledgeaugmented, structgpt, DB-BLINDER}.

\begin{table}[t]
\centering
\renewcommand{\arraystretch}{1.3}
\resizebox{\linewidth}{!}{
\begin{tabular}{lccccccccc}
\toprule
\multirow{2}{*}{Method}
     & \multicolumn{4}{c}{Multi-Hop KBQA} & Single-Hop KBQA & Open-Domain QA & \multicolumn{2}{c}{Slot Filling} & Fact Checking \\ 
     \cmidrule(r){2-5}  \cmidrule(r){6-6} \cmidrule(r){7-7} \cmidrule(r){8-9} \cmidrule(r){10-10}
                            & CWQ & WebQSP & GrailQA  & QALD10-en & Simple Questions & WebQuestions & T-REx & Zero-Shot RE & Creak \\ 
\midrule
\multicolumn{10}{c}{\textit{Without external knowledge}}   \\ \midrule
IO prompt w/ChatGPT     & 37.6            & 63.3 & 29.4  & 42.0 & 20.0 & 48.7 & 33.6 & 27.7 & 89.7  \\
CoT w/ChatGPT           & 38.8          & 62.2 & 28.1  & 42.9 & 20.3 & 48.5 & 32.0 & 28.8 & 90.1  \\ 
SC w/ChatGPT           & 45.4          & 61.1 & 29.6  & 45.3 & 18.9 & 50.3 & 41.8 & 45.4 & 90.8  \\ \midrule
\multicolumn{10}{c}{\textit{With external knowledge}}   \\ \midrule
Prior FT SOTA & 70.4$^\alpha$ & 82.1$^\beta$ & 75.4$^\gamma$ & 45.4$^\delta$ & 85.8$^\epsilon$ & 56.3$^\zeta$ & 87.7$^\eta$ & 74.6$^\theta$ & 88.2$^\iota$  \\
Prior Prompting SOTA & - & 74.4$^\kappa$ & 53.2$^\kappa$ & - &- & - & - & - & -  \\ \midrule
ToG-R (Ours) w/ChatGPT        & 58.9 & 75.8 & 56.4 & 48.6 & 45.4 & 53.2 & 75.3 & 86.5 & 93.8 \\
ToG (Ours) w/ChatGPT       & 57.1 & 76.2 & 68.7 & 50.2 & 53.6 & 54.5 & 76.8 & 88.0 & 91.2  \\ 
ToG-R (Ours) w/GPT-4        & \textbf{69.5} & 81.9  & 80.3     & \textbf{54.7}    & 58.6    & 57.1     &  75.5    & 86.9    & 95.4 \\
ToG (Ours) w/GPT-4      & 67.6 & \textbf{82.6}  & \textbf{81.4} & 53.8  & \textbf{66.7}   & \textbf{57.9} & \textbf{77.1} & \textbf{88.3} & \textbf{95.6}  \\ \bottomrule
\end{tabular}
}
\caption{The ToG results for different datasets. The prior FT (Fine-tuned) and prompting SOTA include the best-known results: $\alpha$: \citet{cwq1}; $\beta$: \citet{webqsp1}; $\gamma$: \citet{gu2023dont}; $\delta$: \citet{qald10-en1}; $\epsilon$: \citet{simple_questions1}; $\zeta$: \citet{webq1}; $\eta$: \citet{trex1}; $\theta$: \citet{kilt}; $\iota$: \citet{creak1}; $\kappa$: \citet{DB-BLINDER}.}
\label{table:main}
\end{table}

\subsubsection{Methods Selected for Comparison}
\label{sec: baselines}
We compare with standard prompting (IO prompt) \citep{io_prompt}, Chain-of-Thought prompting (CoT prompt) \citep{COT}, and Self-Consistency \citep{self_consistency} with 6 in-context exemplars and "step-by-step" reasoning chains. Moreover, for each dataset, we pick previous state-of-the-art (SOTA) works for comparison. We notice that fine-tuning methods trained specifically on evaluated datasets usually have an advantage by nature over methods based on prompting without training, but sacrificing the flexibility and generalization on other data. For a fair play, therefore, we compare with previous SOTA among all prompting-based methods and previous SOTA among all methods respectively. Note that the paper \citet{LLMKBQA} is not involved in comparison because its results are not based on standard exact match and thus incomparable.

\begin{wraptable}{r}{0.5\textwidth}
\centering
\vspace{-1cm}
\resizebox{\linewidth}{!}{
\begin{tabular}{lcc}
\hline
\multirow{2}{*}{\textbf{Method}} & \multirow{2}{*}{\textbf{CWQ}} & \multirow{2}{*}{\textbf{WebQSP}} \\
                                 &                               &                              \\ \hline
\multicolumn{3}{c}{\textit{Fine-tuned}}   \\ \midrule
NSM \citep{NSM}   &   53.9    &    74.3   \\
CBR-KBQA \citep{cwq1} & 67.1 & - \\ 
TIARA \citep{grailqa1}                   &   -     &   75.2          \\
DeCAF \citep{webqsp1}   &   70.4    &    82.1          \\ \hline
\multicolumn{3}{c}{\textit{Prompting}}   \\ \midrule
KD-CoT \citep{wang2023knowledgedriven}  &   50.5    &    73.7          \\
StructGPT \citep{structgpt}  &   -    &    72.6          \\
KB-BINDER  \citep{DB-BLINDER}   &   -    &    74.4   \\ \hline
\multicolumn{3}{c}{\textit{LLama2-70B-Chat}}   \\ \midrule
CoT                     &  39.1     &  57.4            \\
ToG-R                   &  \textbf{57.6}     &  \textbf{68.9}            \\
ToG                   &  53.6     &  63.7            \\
\texttt{Gain}  &  \textcolor[RGB]{0,128,0}{(+18.5)}      &  \textcolor[RGB]{0,128,0}{(+11.5)}             \\ \hline
\multicolumn{3}{c}{\textit{ChatGPT}}   \\ \midrule
CoT                     &  38.8     &  62.2             \\
ToG-R                   &  57.1     &  75.8            \\ 
ToG                   &  \textbf{58.9}     &  \textbf{76.2}            \\
\texttt{Gain}                   &  \textcolor[RGB]{0,128,0}{(+20.1)}      &  \textcolor[RGB]{0,128,0}{(+14.0)}             \\ \hline
\multicolumn{3}{c}{\textit{GPT-4}}   \\ \midrule
CoT                     &  46.0     &  67.3           \\
ToG-R                   &  67.6     &   81.9           \\
ToG                   &  \textbf{69.5}     &   \textbf{82.6}           \\
\texttt{Gain} &  \textcolor[RGB]{0,128,0}{(+23.5)}      &  \textcolor[RGB]{0,128,0}{(+15.3)}             \\ \hline
\end{tabular}
}
\caption{Performances of ToG using different backbone models on CWQ and WebQSP.}
\label{tab: cwq_webqsp}
\end{wraptable}

\subsubsection{Experiment Details}
Given the plug-and-play convenience of ToG, we try three LLMs in experiments: ChatGPT, GPT-4 and Llama-2. We use OpenAI API to call ChatGPT (GPT-3.5-turbo) and GPT-4\footnote{GPT-3.5-turbo and GPT-4 is both from \url{https://openai.com/}}. Llama-2-70B-Chat \citep{llama2} runs with 8 A100-40G without quantization, where the temperature parameter is set to 0.4 for exploration process (increasing diversity) and set to 0 for reasoning process (guaranteeing reproducibility). The maximum token length for the generation is set to 256. In all experiments, we set both width $N$ and depth $D_{max}$ to 3 for beam search. Freebase \citep{freebase} is used as KG for CWQ, WebQSP, GrailQA, Simple Questions, and Webquestions, and Wikidata \citep{Wikidata} is used as KG for QALD10-en, T-REx, Zero-Shot RE and Creak. 
We use 5 shots in ToG-reasoning prompts for all the datasets.
\subsection{Main Results}


\subsubsection{Comparison to Other Methods}
\label{sec:main_results}
Since CoT uses external KG to enhance LLM, we first compare it with those methods leveraging external knowledge as well. As we can see in Figure \ref{table:main}, even if ToG is a training-free prompting-based method and has natural disadvantage in comparison with those fine-tuning methods trained with data for evaluation, ToG with GPT-4 still achieves new SOTA performance in 6 out of 9 datasets, including WebQSP, GrailQA, QALD10-en, WebQuestions, Zero-Shot RE and Creak. Even for some dataset without SOTA, e.g., CWQ, the performance of CoT has already been close to SOTA (69.5\% v.s. 70.4\%). If comparing with all promoting-based methods, both ToG with GPT-4 and its weaker version ToG with ChatGPT can win the competition in all datasets. In particular, the improvement of 1.6\% on open-domain QA dataset WebQuestions demonstrates the ToG's generality on open-domain QA tasks. We also notice that the performance of ToG on single-hop KBQA dataset is not as good as its performance on other datasets. These results indicate that ToG is more effective on multi-hop datasets in general, which supports our argument that ToG enhances the deep reasoning capability of LLMs. 

We also see from Figure \ref{table:main} that, compared with those methods without leveraging external knowledge (e.g, IO, CoT and SC prompting methods), the advantage of ToG is more significant. For example, the performance improves 51.8\% and 42.9\% on GrailQA and Zero-Shot RE, respectively. It turns out that benefits from external KG can not be ignored in reasoning. 

ToG outperforms ToG-R on most datasets since the triple-based reasoning paths provide additional intermediate entity information compared to the relation chains retrieved by ToG-R. More detailed analysis of the answers generated by ToG can be checked in Appendix~\ref{sec: analysis}. And the results of previous methods on each dataset are reported in Appendix~\ref{sec: datasets} for better comparison,

\subsubsection{Performances with Different Backbone Models}
\label{sec:backbone}
Given ToG's flexibility of plug-and-play, we evaluate how different backbone models affect its performance on two datasets CWQ and WebQSP. Table \ref{tab: cwq_webqsp} shows that, as we expected, the performance of CoT improves with the size (also reflecting partially the reasoning ability) of backbone models (GPT-4 > ChatGPT > Llama-2). Furthermore, we see that, the larger the backbone model, the larger the gap between CoT and ToG (the gain increases from 18.5\% for Llama-2 to 23.5\% for GPT-4 on CWQ, and from 11.5\% for Llama-2 to 15.3\% for GPT-4 on WebQSP), and this indicates more potential of KG can be mined using a more powerful LLM.  

In addition, even if using the smallest model Llama-2 (70B parameters), ToG outperforms CoT with GPT-4.
This implies a much cheaper technical route for LLM deployment and application, i.e., TOG with cheap small LLM may be a candidate for substituting expensive big LLM, especially in vertical scenarios that external KGs can cover.

\subsubsection{Ablation Study}~\label{sec:ablation1}
\begin{figure}[t]
    \centering
    \begin{subfigure}[b]{0.242\textwidth}
        \includegraphics[width=\textwidth]{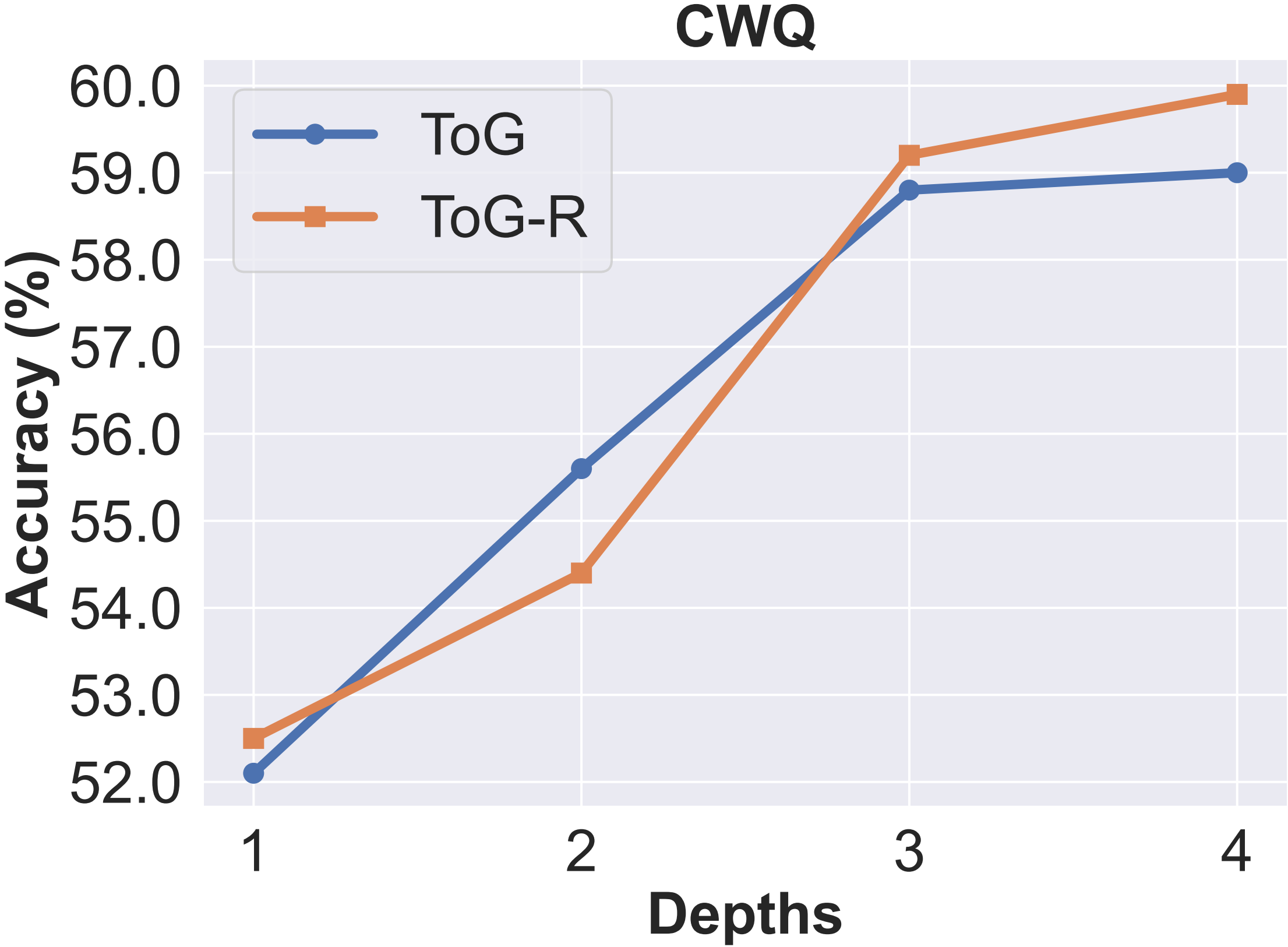}
    \end{subfigure}
    \hfill
    \begin{subfigure}[b]{0.242\textwidth}
        \includegraphics[width=\textwidth]{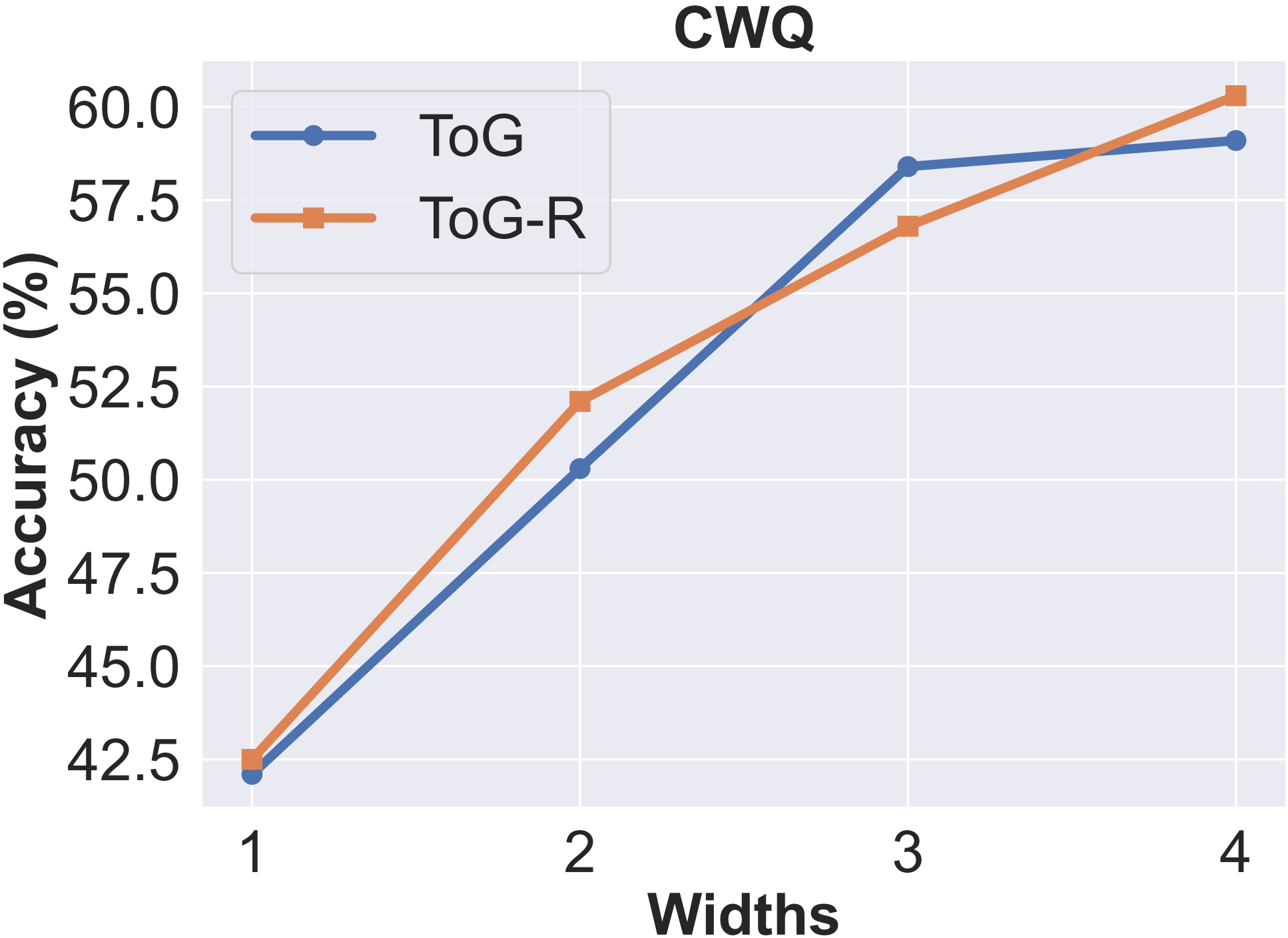}
    \end{subfigure}
    \hfill
    \begin{subfigure}[b]{0.242\textwidth}
        \includegraphics[width=\textwidth]{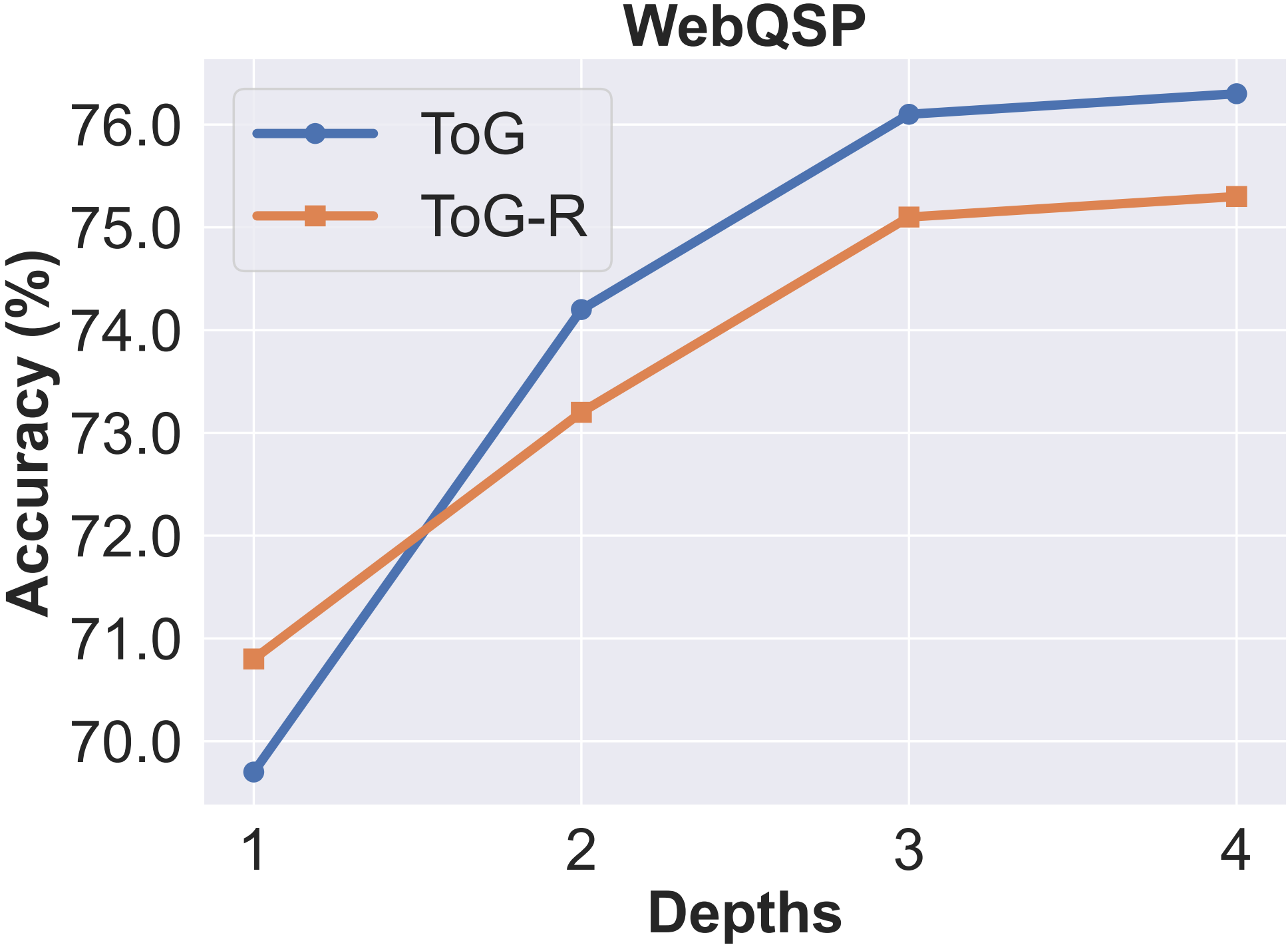}
    \end{subfigure}
    \hfill
    \begin{subfigure}[b]{0.242\textwidth}
        \includegraphics[width=\textwidth]{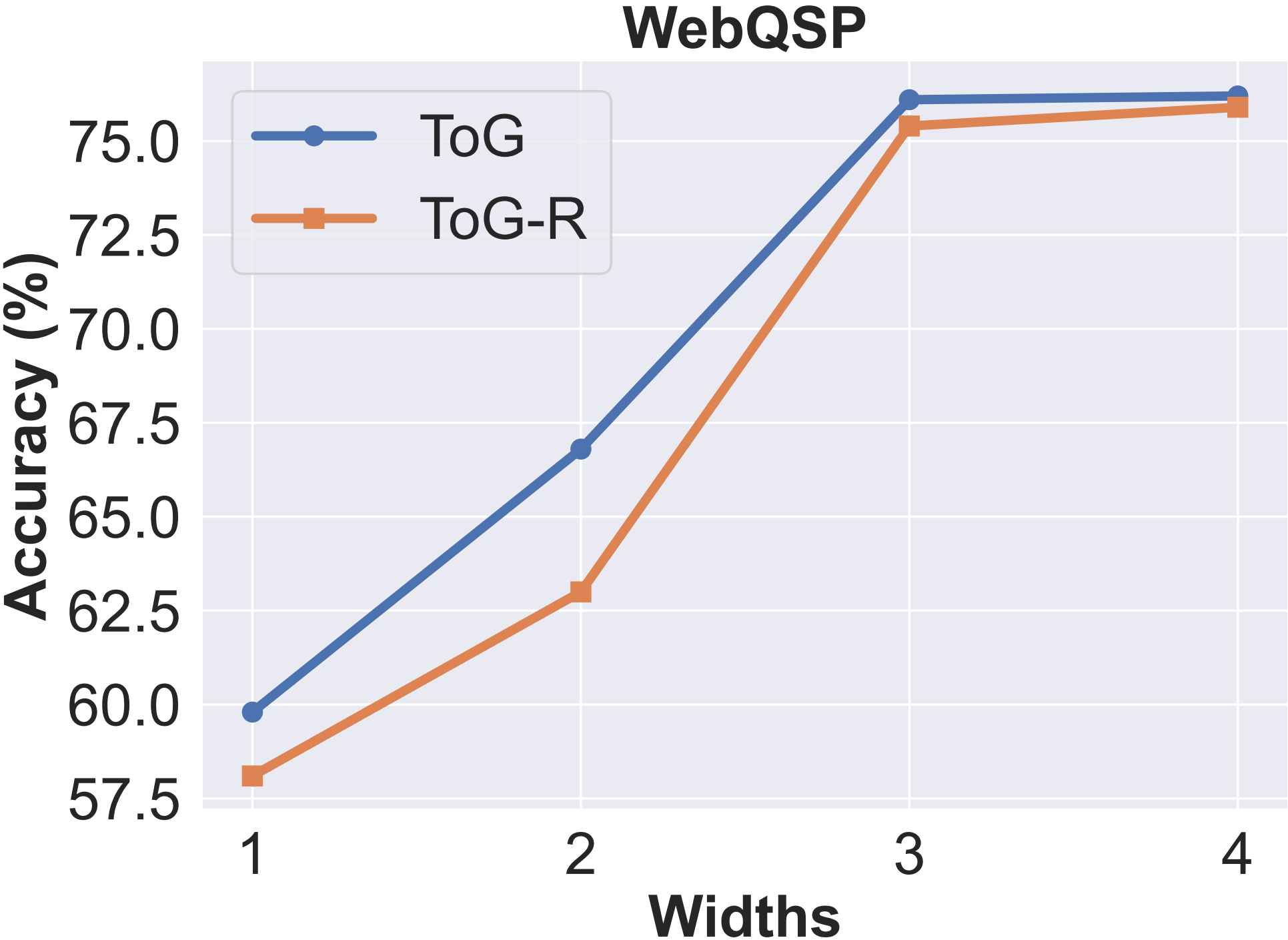}
    \end{subfigure}
    \caption{Performances of ToG with different search depths and widths.}
    \label{fig: depth_width}
\end{figure}

We perform various ablation studies to understand the importance of different factors in ToG. 
We conduct our ablation studies on two subsets of the test sets of CWQ and WebQSP, each of which contains 1,000 randomly sampled questions.
\paragraph{Do search depth and width matter for ToG?}
\label{sec: depth_breadth}
\begin{wraptable}{r}{0.4\textwidth}
\centering
\vspace{-0.4cm}
\resizebox{\linewidth}{!}{
\begin{tabular}{lcc}
\hline
\multirow{2}{*}{\textbf{Method}} & \multirow{2}{*}{\textbf{CWQ}} & \multirow{2}{*}{\textbf{WebQSP}} \\
                                 &                               &                              \\ \hline
\textbf{CoT}                   &   37.6    &  62.0            \\ \hline
\textbf{ToG}                   &       &              \\
w/ Freebase              &  58.8      &    76.2           \\
w/ WikiData              &  54.9      &   68.6            \\ \hline
\textbf{ToG-R}                   &       &               \\
w/ Freebase              &  59.2     &      75.1         \\
w/ WikiData              &    51.9    &    66.7           \\ \hline
\end{tabular}
}
\caption{Performances of ToG using different source KGs on CWQ and WebQSP.}
\label{table: KG}
\end{wraptable}
To explore the influence of the search depth $D_{max}$ and the beam width $N$ on ToG's performance, we conduct experiments under settings with depths ranging from 1 to 4 and widths from 1 to 4. As shown in Figure \ref{fig: depth_width}, ToG's performance improves with the search depth and width. This also implies that ToG's performance could potentially be improved with the increment of the exploration depth and breadth.
However, considering the computational cost (which increases linearly with the depth), we set both the depth and width to 3 as the default experimental setting. On the other hand, the performance growth diminishes when the depth exceeds 3. This is mainly because only a small part of questions have the reasoning depths (based on the number of relations in SPARQL,
as seen in Figure \ref{fig: cal_links} in the Appendix) 
 of greater than 3. 

\paragraph{Do different KGs affect ToG's performance?}


One of the main advantages of ToG is its plug-and-play capabilities. As shown in Table~\ref{table: KG}, ToG achieves significant improvements with different source KGs on CWQ and WebQSP, compared to CoT. On the other hand, different source KGs might have different effects on the performance of ToG. Notably, Freebase brings more significant improvements on CWQ and WebQSP than Wikidata, since both datasets are constructed upon Freebase. Moreover, in a very large KG like Wikidata, the searching and pruning processes are relatively challenging.


\paragraph{How do different prompt designs affect ToG?}
\label{sec: prompt_construction}
\begin{wraptable}{r}{0.4\textwidth}
\centering

\resizebox{\linewidth}{!}{
\begin{tabular}{lcc}
\hline
\multirow{2}{*}{\textbf{Method}} & \multirow{2}{*}{\textbf{CWQ}} & \multirow{2}{*}{\textbf{WebQSP}} \\
                                 &                               &                              \\ \hline
\textbf{ToG}                   &       &              \\
w/ Triples          &   58.8      &   76.2          \\
w/ Sequences              &  57.2      &   73.2          \\
w/ Sentences              &  58.6      &    73          \\ \hline
\textbf{ToG-R}                   &       &             \\
w/ Sequences              & 59.2       &    75.1            \\
w/ Sentences              &  50.1      &   67.3            \\ \hline
\end{tabular}
}
\caption{Performances of ToG using different prompting designs.}
\label{table: prompt}
\end{wraptable}

We perform additional experiments to determine which types of prompt representations can work well for our approach. The results are presented in Table~\ref{table: prompt}. "Triples" denotes using triple formats as prompts to represent multiple paths, such as "(Canberra, capital of, Australia), (Australia, prime minister, Anthony Albanese)". "Sequences" refers to the utilization of a sequence format, as illustrated in Figure~\ref{fig: framework}. "Sentences" involves converting the triples into natural language sentences. For example, "(Canberra, capital of, Australia)" can be converted to "The capital of Canberra is Australia."
The result shows that the utilization of triple-based representations for the reasoning paths yields the highest degree of efficiency and superior performance. Conversely, when considering ToG-R, each reasoning path is a relation chain starting from a topic entity, rendering it incompatible with the triple-based prompt representation. Consequently, the transformation of ToG-R into the natural language form results in excessively lengthy prompts, thereby leading to a notable deterioration in performance.

\paragraph{Comparing the affects from different pruning tools.}
\label{sec: tools}

\begin{wraptable}{r}{0.4\textwidth}
\centering
\vspace{-0.5cm}
\resizebox{\linewidth}{!}{
\begin{tabular}{lcc}
\hline
\multirow{2}{*}{\textbf{Method}} & \multirow{2}{*}{\textbf{CWQ}} & \multirow{2}{*}{\textbf{WebQSP}} \\
                                 &                               &                              \\ \hline
\textbf{ToG}                   &                      &                      \\
w/BM25                           & 51.4                 & 58.7                 \\
w/SentenceBERT                   & 51.7                 & 66.3                 \\
w/ChatGPT                        & 58.8                 & 76.2                 \\ \hline
\textbf{ToG-R}                   & \multicolumn{1}{l}{} & \multicolumn{1}{l}{} \\
w/BM25                           & 49.4                 & 57.3                 \\
w/SentenceBERT                   & 50.1                 & 60.1                 \\
w/ChatGPT                        & 59.2                 & 75.1                 \\ \hline
\end{tabular}
}
\caption{Performances of ToG using different pruning tools.}
\label{table: tools}
\end{wraptable}
Other than the LLM, lightweight models that can measure text similarity like BM25 and SentenceBERT, can be employed as pruning tools in the exploration phase. We can select top-$N$ entities and relations based on their literal similarities with the question. We investigate the impacts of different pruning tools on the performance of the ToG, as demonstrated in Table \ref{table: tools}. The replacement of the LLM with either BM25 or SentenceBERT results in the significant performance degradation of our approach. Concretely, the results on CWQ drop on average by 8.4\%, and the results on WebQSP drop on average by 15.1\%.
The results show that the LLMs perform best as a pruning tool in terms of effectiveness. On the other hand, after utilizing
the BM25 or SentenceBERT, we only need $D + 1$ calls to the LLM instead of $2ND + D + 1$ as we
discuss in Section \ref{sec: reasoning}, which enhances the efficiency of ToG. 

We conduct additional ablation studies on the effect of the number of seed exemplars and the difference between ToG and naive beam search on the KG, which can be seen in Appendix~\ref{sec:ablation2}. 

\subsection{Knowledge Traceability and Correctability in ToG}
\label{sec: application}
\begin{figure}[t]
	\centering
	\includegraphics[width=0.99\columnwidth]{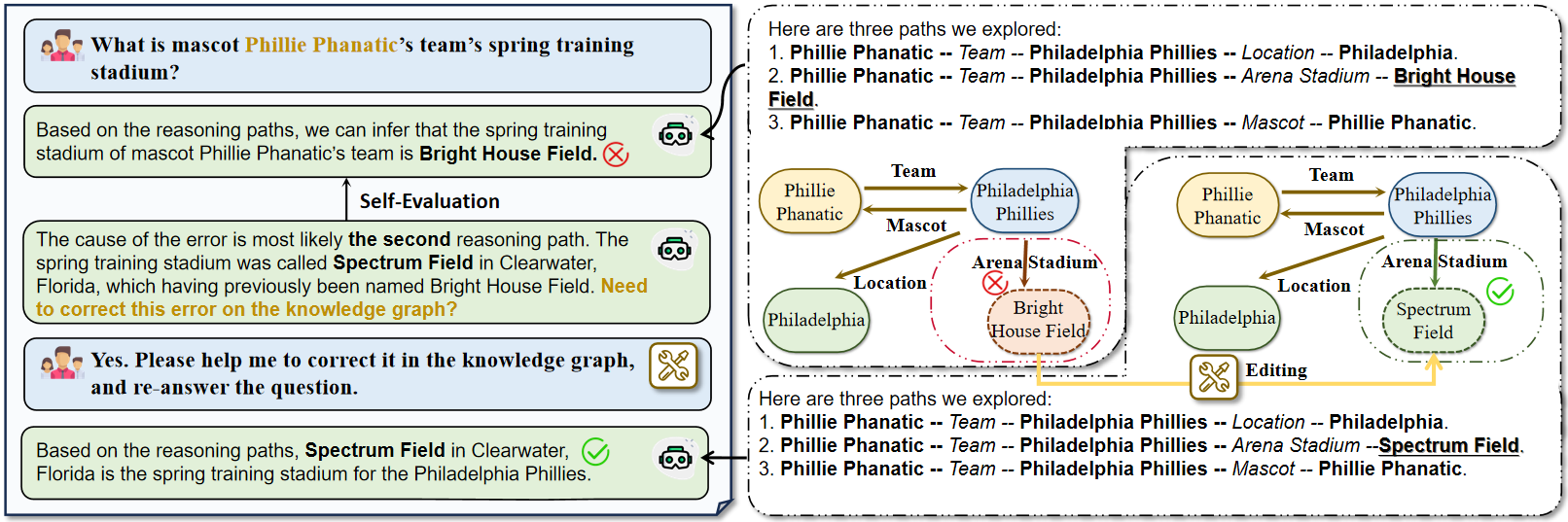}
	\caption{The illustration of knowledge traceability and correctability of ToG.}
	\label{fig: app}
\end{figure}

The quality of KG is very important for correct reasoning by ToG. An interesting feature of ToG is knowledge traceability and knowledge correctability during LLM reasoning, and it provides a way to improve KG's quality using ToG itself and reduce the cost of KG construction and correction. As illustrated in Figure \ref{fig: app}, the explicit reasoning paths of the ToGs can be displayed to users. If potential errors or uncertainties in ToG answers are discovered by human users/experts or other LLMs, ToG has the ability to trace back and examine the reasoning path, find suspicious triples with errors, and correct them. 


Take the case in Figure \ref{fig: app} as an example. Given the input question ``What is mascot Phillie Phanatic's team's spring training stadium?'', ToG outputs the wrong answer ``Bright House Field'' in the first round. Then ToG traces back all reasoning paths, localizes the cause of the error may come from the second reasoning path (Phillie Phanatic $\xrightarrow{\text{Team}}$ Philadelphia Phillies $\xrightarrow{\text{Arena Stadium}}$ Bright House Field), and analyzes that the error comes from the old name ``Specturm Field'' of ``Bright House Field'' in the outdated triple (\textit{Philadelphia Phillies}, \textit{Arena Stadium}, \textit{Bright House Field}). According to the hints from ToG, user can ask LLM to correct this error and answer the same question with correct information. This example reveals that ToG not only enhances LLM with KG, but also improves the quality of KG with LLM, known as knowledge infusion \citep{knowledge_infusion}. 

\section{Related Work}

\paragraph{Reasoning with LLM Prompting}
Chain-of-Thought (CoT) \citep{COT} has been shown to be effective in enhancing LLM reasoning. It creates a series of prompt instances according to reasoning logic under a few-shot learning paradigm in order to improve LLM's performance on complex tasks. The thought of CoT has been improved along different dimensions, including Auto-CoT \citep{auto}, Complex-CoT \citep{complex_cot}, Self-Consistency \citep{self_consistency}, Zero-Shot-CoT \citep{zero_shot_cot}, Iter-CoT \citep{iter-cot}, ToT \citep{tot}, GoT \citep{got} and so on. Given the limitation that all these works only use the knowledge in training data, recent efforts such as ReAct~\citep{yao2022react} attempt to utilize the information from external sources such as Wiki documents to further improve the reasoning performance. 




\paragraph{KG-enhanced LLM} KG has advantages in dynamic, explicit, and structured knowledge representation~\citep{pan2023unifying} and techniques combining LLMs with KGs have been studied. Early studies~\citep{Knowbert,huang2024joint, luo2024rog, zhang2021poolingformer, li2023trea, liu2020reasoning} embed structured knowledge from KGs into the underlying neural networks during the pretraining or fine-tuning process. However, KG embedded in LLM sacrifices its own nature of explainability in knowledge reasoning and efficiency in knowledge updating~\citep{KELLMsurvey}.

Recent works instead combine LLMs with KGs by translating relevant structured knowledge from KGs to textual prompts for LLMs. All the methods follow a fixed pipeline that retrieves extra information from KGs to augment the LLM prompt and they belong to the $\hbox{LLM}\oplus \hbox{KG}$ paradigm we defined in the introduction section. On the other hand, \citet{structgpt} asks LLM to explore KG and so it can be regarded as a special case of ToG, which belongs to the $\hbox{LLM}\otimes \hbox{KG}$ paradigms.

\section{Conclusion}
We introduce the $\hbox{LLM}\otimes \hbox{KG}$ paradigm for integrating LLMs and KGs in a tight-coupling manner, and propose the Think-on-Graph (ToG) algorithmic framework which leverages LLM as a agent participating in KG reasoning for better decision-making. Experimental results demonstrate that ToG outperforms existing fine-tuning-based methods and prompting-based methods without additional training cost and mitigates the hallucination issue of LLMs.

\section{Acknowledgement}
We express our sincere gratitude to the esteemed reviewers for their invaluable feedback and constructive comments, which significantly contributed to the improvement and refinement of this paper. Their insightful suggestions and meticulous attention to detail have played a pivotal role in enhancing the quality and clarity of our research work.

\bibliography{iclr2024_conference}
\bibliographystyle{iclr2024_conference}


\newpage

\appendix

\section{Algorithm for ToG}
We summarize the comprehensive algorithmic procedure of ToG and ToG-R, as shown in Figure Algorithm \ref{algorithm1} and \ref{algorithm2}.

\begin{figure}[htb]
\begin{minipage}[t]{0.5\textwidth}
\begin{algorithm}[H]
\begin{algorithmic}
\Require Input $x$, LLM $\pi$, depth limit $D_{max}$ sample limit $N$.
\State Initialize $E^{0} \gets $ Extract entities on $x$, $P \gets []$, $M \gets 0$.
\While {$D$ $\leq$ $D_{max}$} 
    \State  $R^{D}_{cand}$, $P_{cand}$  $ \gets $ Search($x$, $E^{D-1}$, $P$)
    \State $R^{D}$, $P$  $\gets $ Prune($\pi$, $x$,  $R^{D}_{cand}$, $P_{cand}$) 
    \State $E^{D}_{cand}$, $P_{cand}$  $\gets $ Search($x$, $E^{D-1}$, $R^{D}$, $P$)
    \State $E^{D}$, $P$ $\gets $ Prune($\pi$, $x$,  $E^{D}_{cand}$, $P_{cand}$)
    \If {Reasoning($\pi$, $x$, $P$)} 
        \State Generate($\pi$, $x$, $P$)
        \State \textbf{break}
    \EndIf
    \State Increment $D$ by 1.
\EndWhile%
\If {$D$ $>$ $D_{max}$}
    \State Generate($\pi$, $x$)
\EndIf
\caption{ToG}
\label{algorithm1}
\end{algorithmic}
\end{algorithm}
\end{minipage}
\begin{minipage}[t]{0.5\textwidth}
\begin{algorithm}[H]
\begin{algorithmic}
\Require Input $x$, LLM $\pi$, depth limit $D_{max}$ sample limit $N$.
\State Initialize $E^{0} \gets $ Extract entities on $x$, $P \gets []$, $M \gets 0$.
\While {$D$ $\leq$ $D_{max}$} 
    \State  $R^{D}_{cand}$, $P_{cand}$  $ \gets $ Search($x$, $E^{D-1}$, $P$)
    \State $R^{D}$, $P$  $\gets $ Prune($\pi$, $x$, $R^{D}_{cand}$, $P_{cand}$) 
    \State $E^{D}_{cand}$, $P_{cand}$  $\gets $ Search($x$, $E^{D-1}$, $R^{D}$, $P$)
    \If {Reasoning($\pi$, $x$, $P$, $E^{D}_{cand}$)} 
        \State Generate($\pi$, $x$, $P$, $E^{D}_{cand}$)
        \State \textbf{break}
    \EndIf
    \State $E^{D}$, $P$ $\gets $ Random\_Prune($E^{D}_{cand}$, $P_{cand}$)
    \State Increment $D$ by 1.
\EndWhile%
\If {$D$ $>$ $D_{max}$}
    \State Generate($\pi$, $x$)
\EndIf
\caption{ToG-R}
\label{algorithm2}
\end{algorithmic}
\end{algorithm}
\end{minipage}
\end{figure}

\section{Additional Ablation Study and Experiment Analysis}
In this section, we conduct more experiments for ablation study in addition to Section \ref{sec:ablation1}, and analyze experimental results of ToG in detail.

\subsection{Additional Ablation Study}~\label{sec:ablation2}
\begin{figure}[t]
    \centering
    \begin{subfigure}[b]{0.49\textwidth}
        \includegraphics[width=\textwidth]{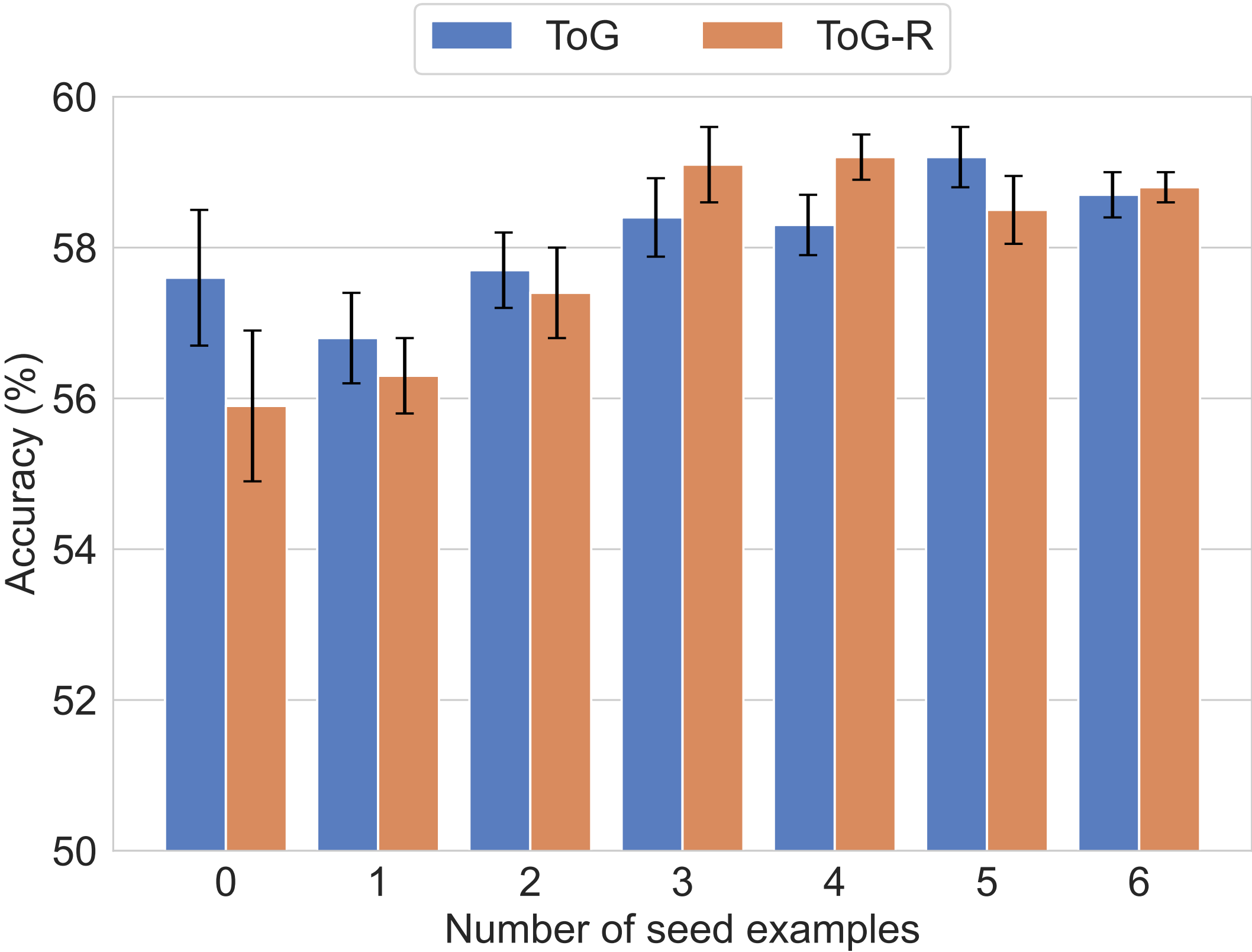}
    \end{subfigure}
    \hfill
    \begin{subfigure}[b]{0.49\textwidth}
        \includegraphics[width=\textwidth]{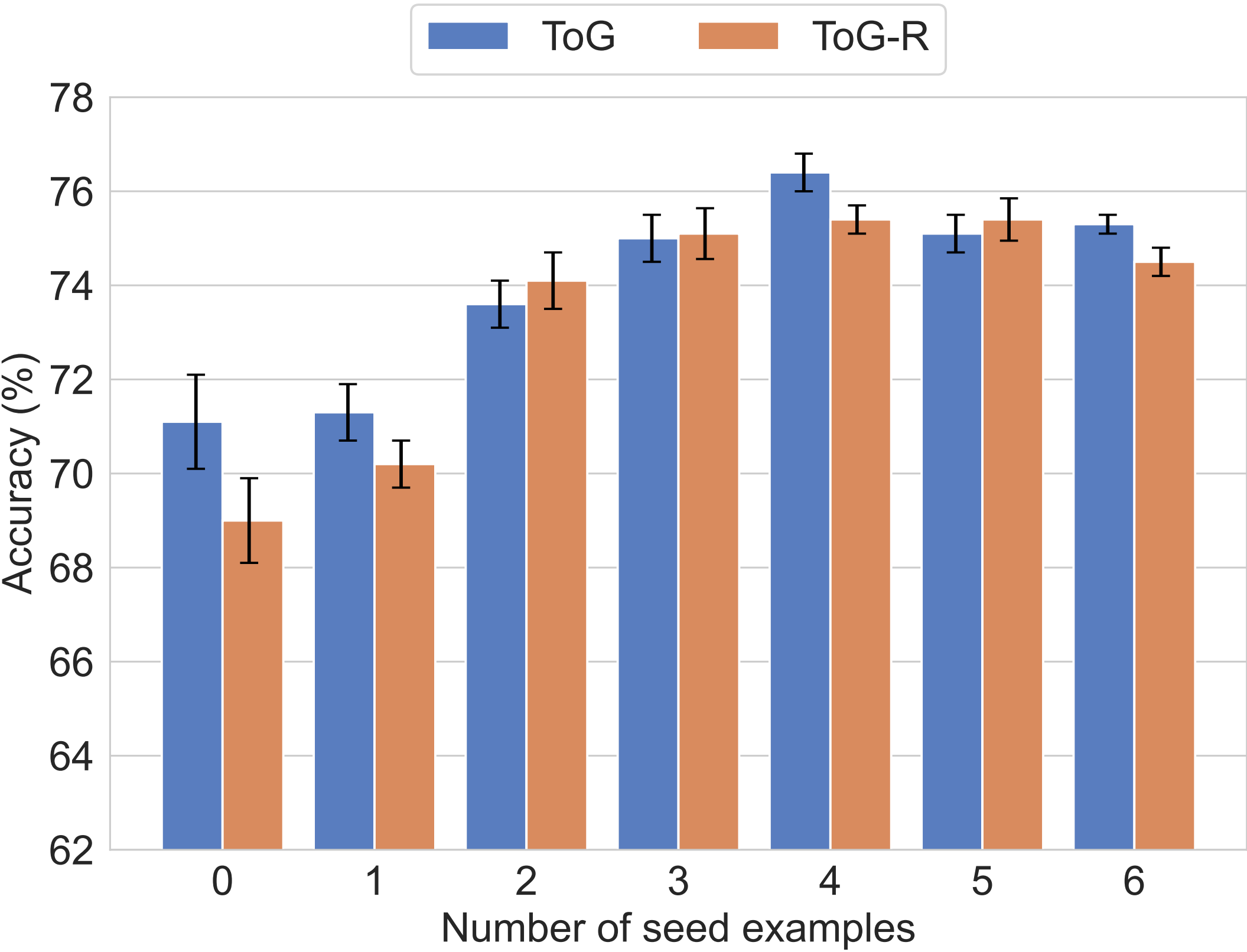}
    \end{subfigure}
    \caption{Exemplar sensitivity analysis for CWQ and WebQSP for ToG, where "0" denotes zero-shot and "k" denotes k-shot.}
    \label{fig: shots}
\end{figure}

\paragraph{Sensitivity to the Number of Seed Examplars}
To better understand how sensitive ToG is sensitivity to the number of seed exemplars, we employ sensitivity analysis shown in Figure \ref{fig: shots}. We conduct zero-shot experiment and select 1-6 examples from the training set as few-shot setting. In the few-shot tests, we randomly chose $M$ of $\{1,2,3,4,6\}$ exemplars as demonstrations and replicated the experiments three times. As the number of examples in the demonstrations increases, the overall performance also generally improves. However, the performance peaks for ToG and ToG-R differ (with the best performance for ToG at 5-shot and for ToG-R at 4-shot). Moreover, ToG's zero-shot performance outpaces ToG-R. This can be attributed to ToG having fully completely explored paths, ensuring commendable performance even in zero-shot. In contrast, ToG-R omits entities in the path, but its average performance with demonstrations is superior to ToG.

\begin{figure}[t]
    \centering
    \begin{subfigure}[b]{0.32\textwidth}
        \includegraphics[width=\textwidth]{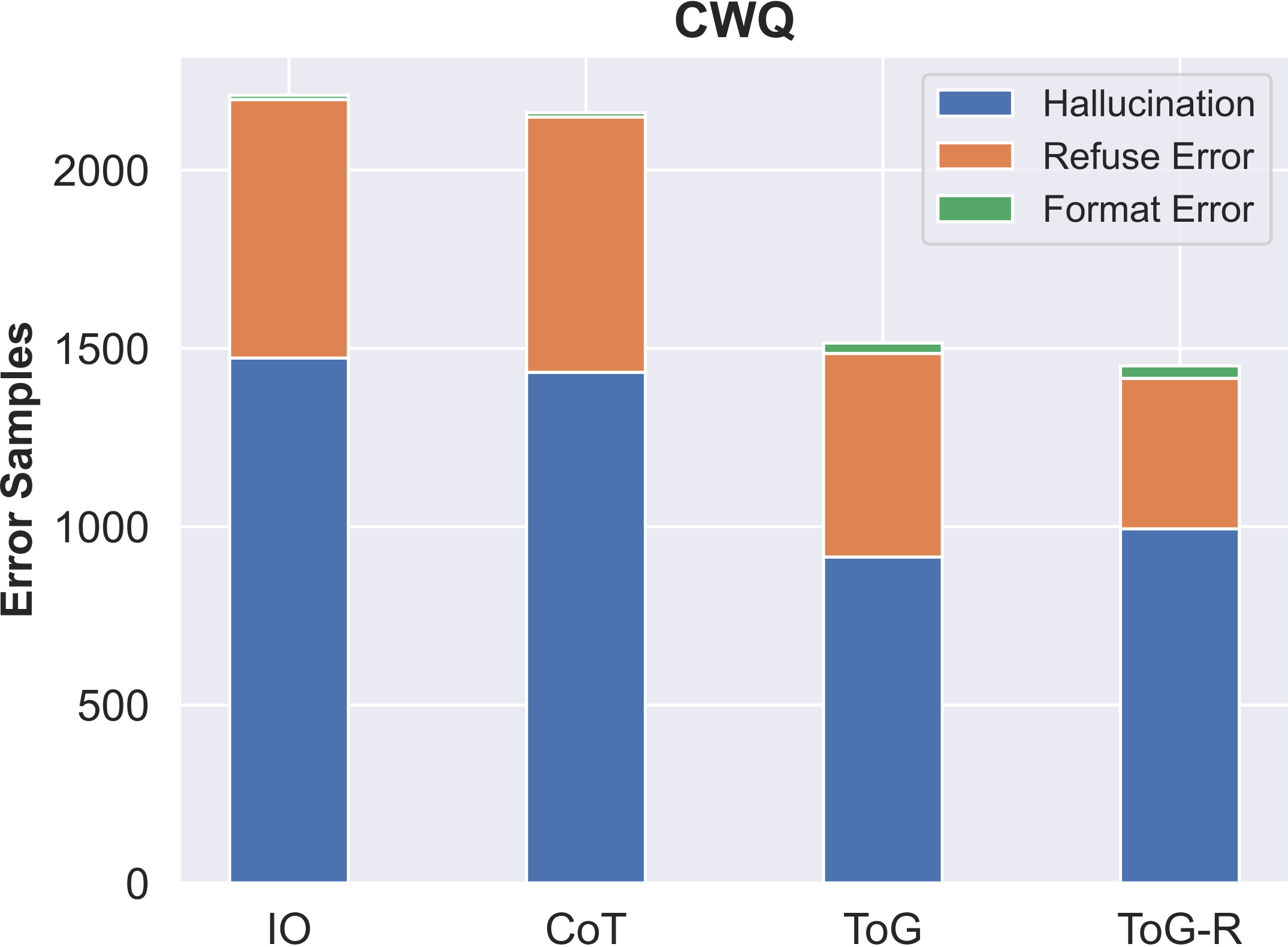}
    \end{subfigure}
    \hfill
    \begin{subfigure}[b]{0.32\textwidth}
        \includegraphics[width=\textwidth]{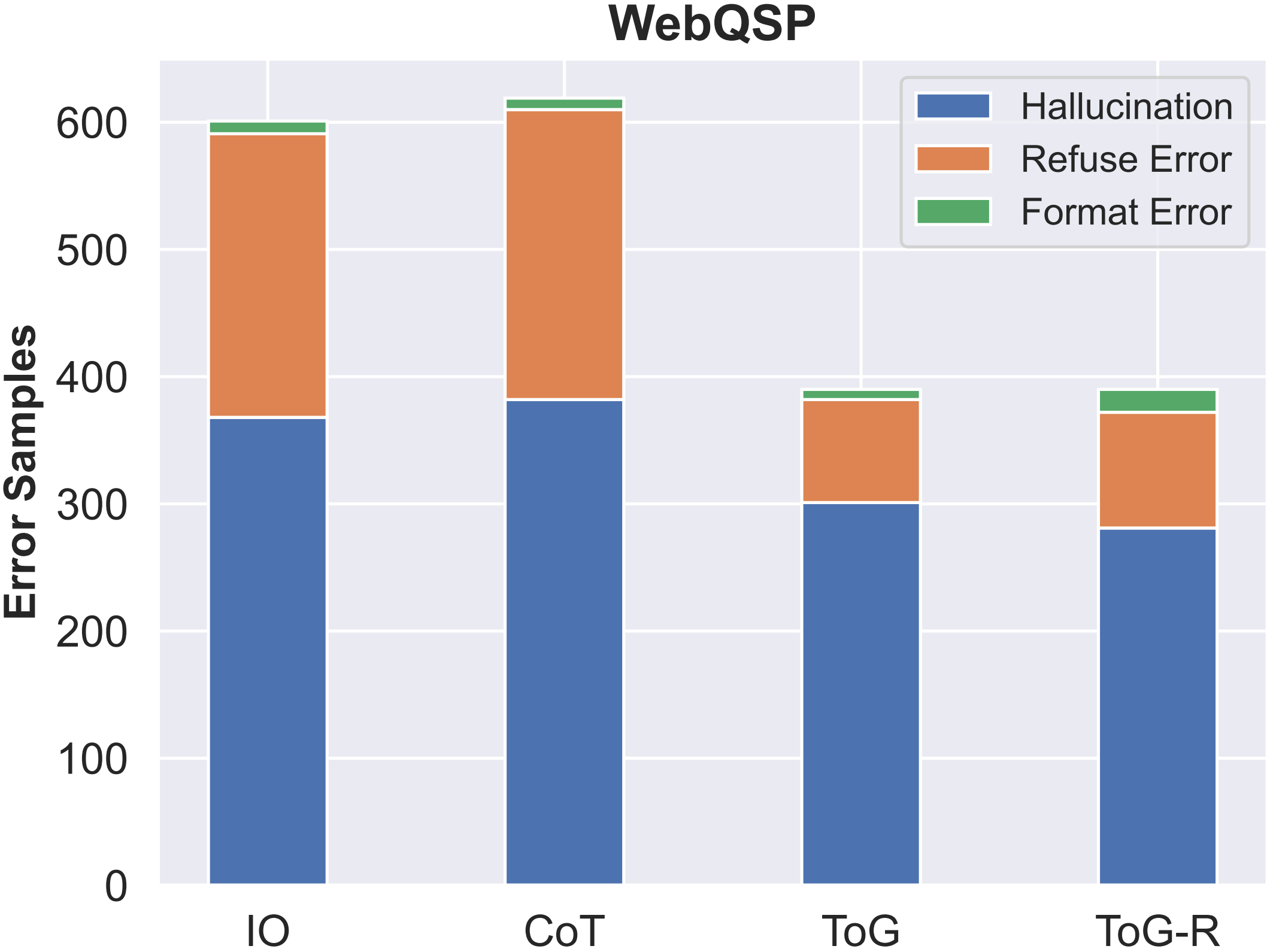}
    \end{subfigure}
    \hfill
    \begin{subfigure}[b]{0.32\textwidth}
        \includegraphics[width=\textwidth]{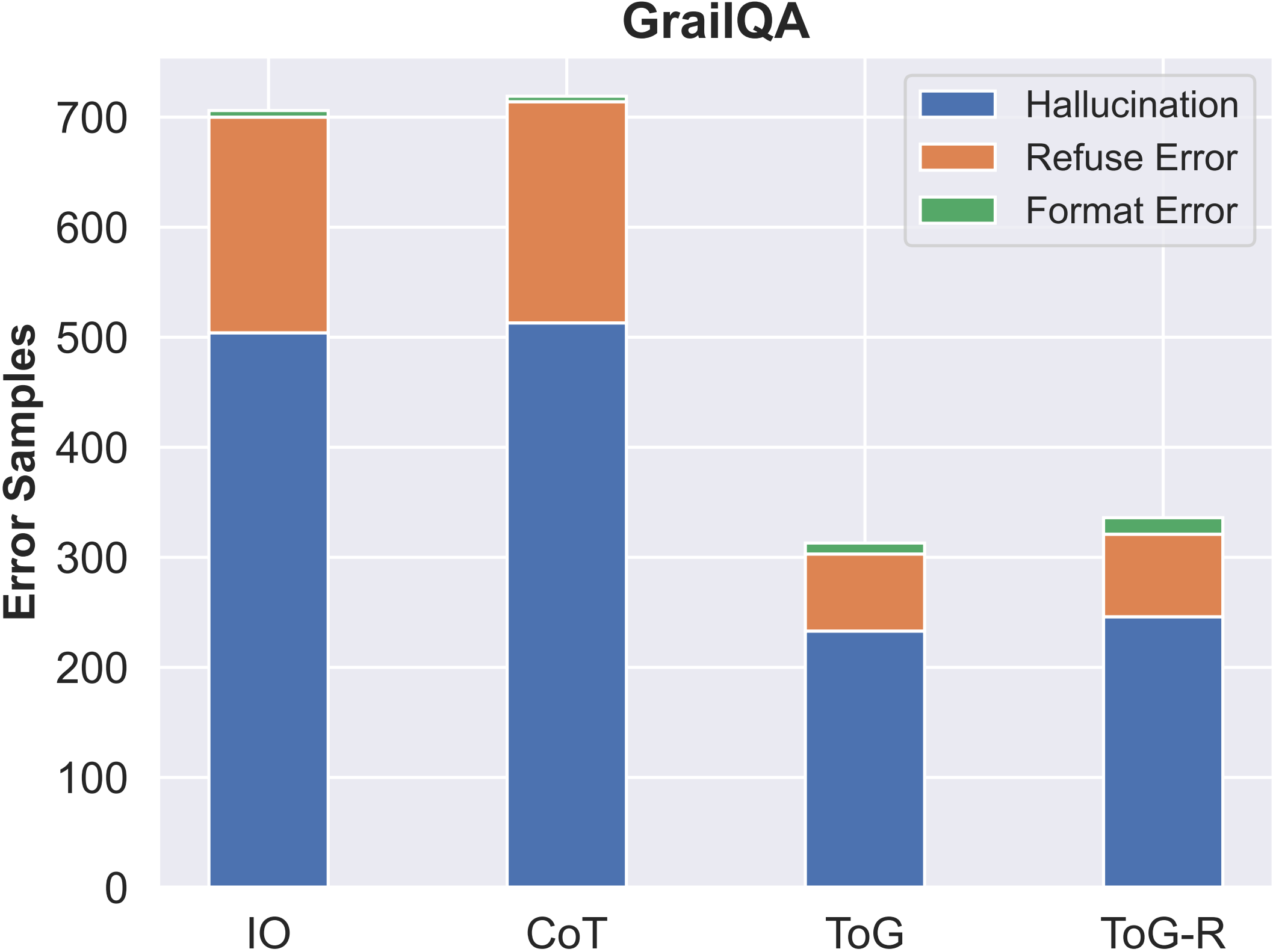}
    \end{subfigure}
    \caption{The erroneous instances and categories in the CWQ, WebQSP, and GrailQA of IO, CoT, and ToG.}
    \label{fig: error_analysis}
\end{figure}

\begin{wraptable}{r}{0.4\textwidth}
\centering
\resizebox{\linewidth}{!}{
\begin{tabular}{llc}
\hline
       \textbf{Search Algorithm}    & \textbf{Dataset} &  \textbf{EM} \\ \hline
\multirow{2}{*}{Naive Beam Search} 

&  	 
 CWQ                    & 30.1		 \\
 &   
 WebQSP                  &  46.1	   \\ \hline
\multirow{2}{*}{TOG-R}                 &  CWQ           &  59.2  \\ 
 &  WebQSP           &  75.1  \\\hline
\multirow{2}{*}{TOG}                 &  CWQ           &  58.8  \\ 
 &  WebQSP           &  76.2  \\\hline
\end{tabular}	
}
\caption{The results of Naive Beam Search, ToG methods on CWQ and WebQSP.}
\label{tab: diff_search_algo}
\end{wraptable}
\paragraph{Difference with Naive Beam Search}
ToG is slightly different from the beam search. ToG uses the top-$N$ reasoning paths as evidence while the naive beam search chooses the most plausible path as the only reasoning path. We conduct naive top1-beam search methods for ToG on CWQ and WebQSP. For each depth of the ToG, we choose the reasoning path with the highest plausibility, to evaluate if the current reasoning path is sufficient to answer the questions. The experiment results are shown in Table \ref{tab: diff_search_algo}. In naive beam search, the calibration error accumulates along the inference, leading to the instability of the final result.
We believe that ToG can partially alleviate this issue by considering the top-$N$ reasoning paths.

\subsection{Result Analysis}\label{sec: analysis}
We conduct a detailed analysis on the answers generated by ToG and ToG-R.

\paragraph{Error Analysis}
We considered three types of errors: (1) Hallucination error, (2) Refuse error \footnote{LLM will refuse to answer due to lack of information.}, and (3) Format error. The distribution is shown in Figure \ref{fig: error_analysis}. Our approach has significantly reduced the hallucination and refusal to answer error types in IO and CoT. For GrailQA, ToG even reduces these types of errors by 50\% and 60\%, respectively. Moreover, in ToG's error samples, there are still many instances of hallucination and refusal to answer errors. This is because the current search depth and width are both set to 3. By increasing the search depth and width, these error instances will further decrease (refer to Section \ref{sec: depth_breadth}). Furthermore, we currently generalize incorrect answers as hallucinations, but there are various categories within hallucinations, which we won't discuss in this paper. Additionally, after applying ToG, there's a slight increase in samples with format errors. This result shows that the explored paths lead to a noticeable increase in the tokens, sometimes even exceeding the maximum output limit. However, the error rate from this issue is negligible (less than 3\%).

\begin{figure}[htbp]
    \centering
    \begin{subfigure}[b]{0.32\textwidth}
        \includegraphics[width=\textwidth]{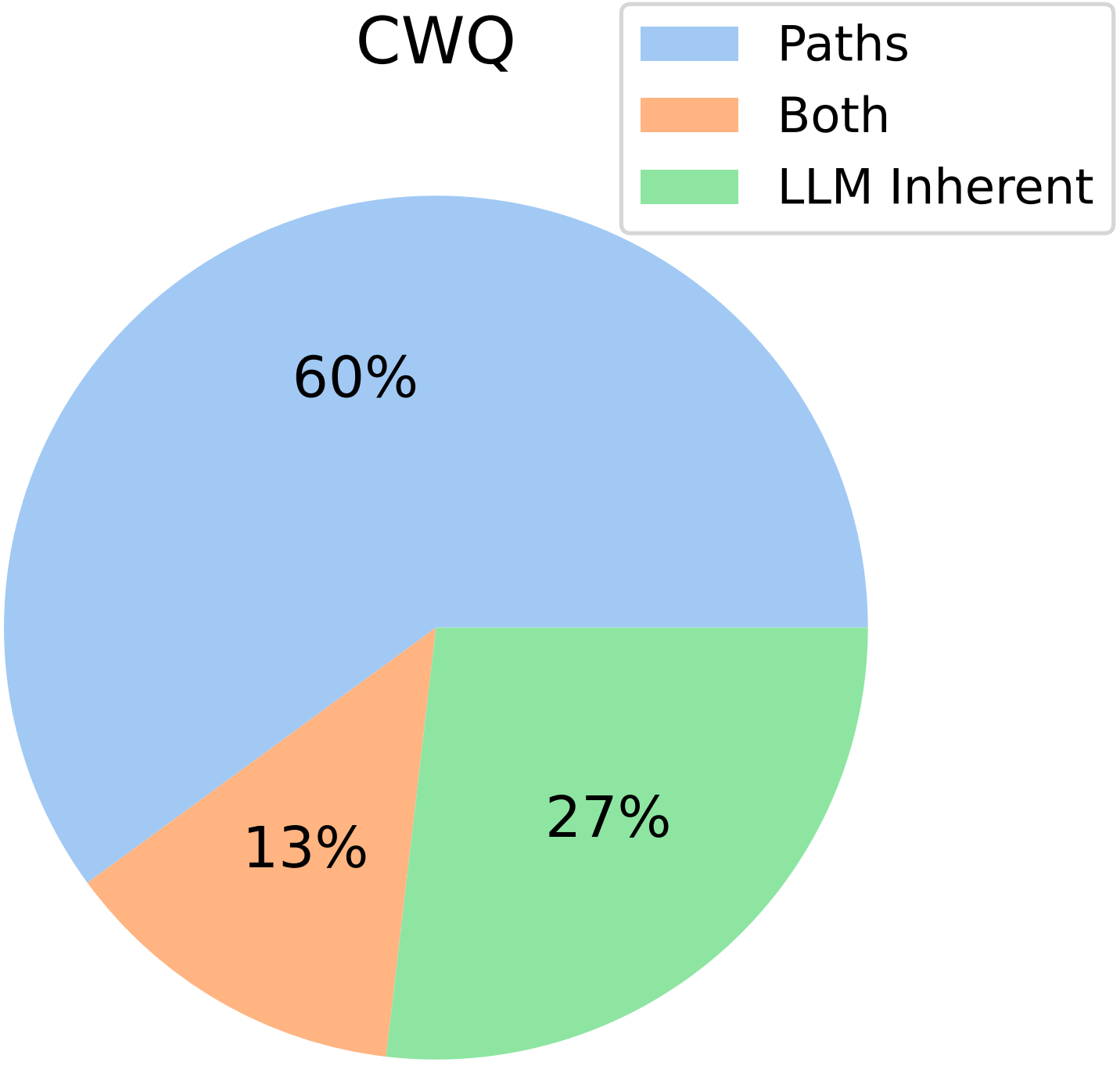}
    \end{subfigure}
    \hfill
    \begin{subfigure}[b]{0.32\textwidth}
        \includegraphics[width=\textwidth]{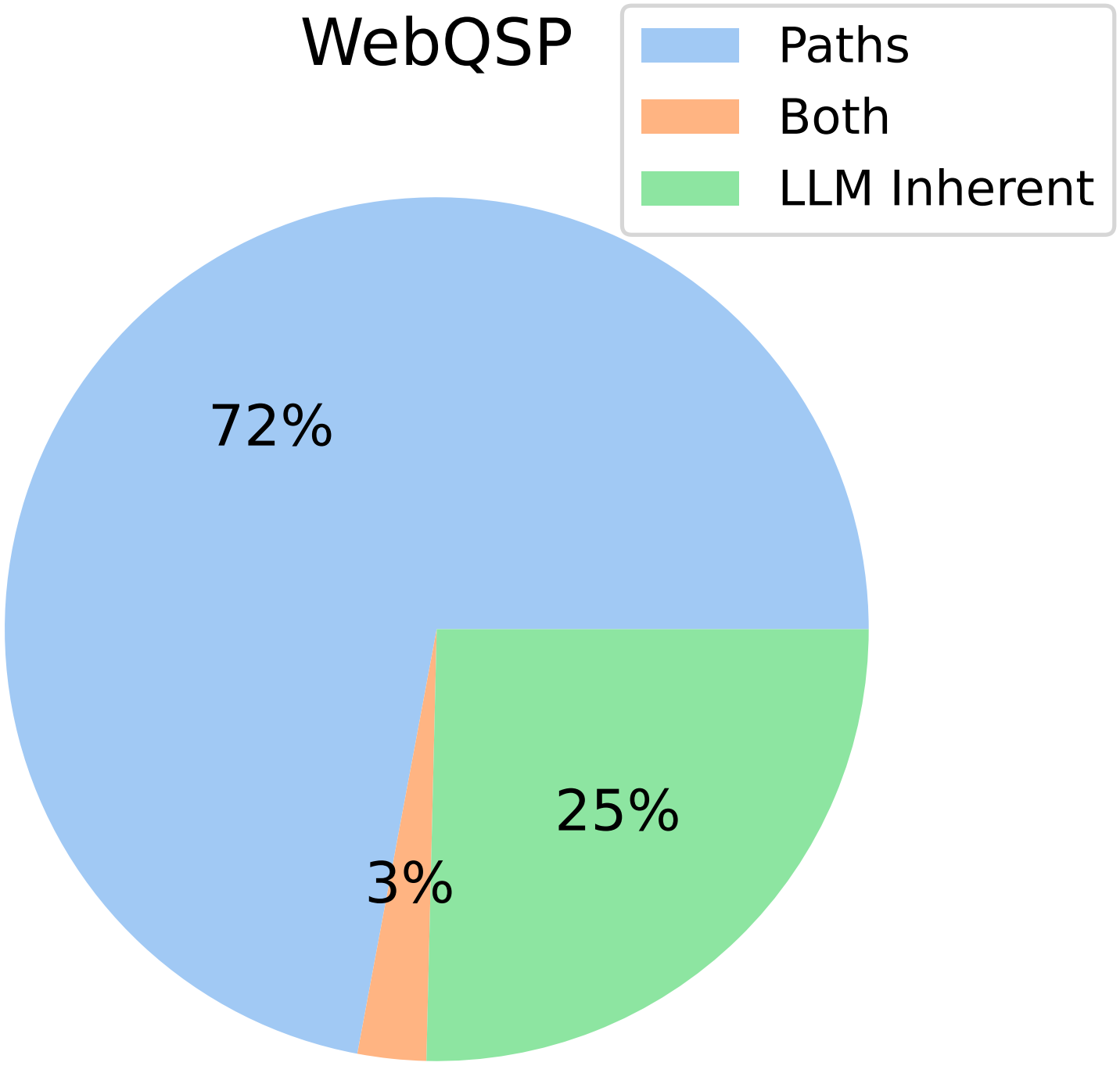}
    \end{subfigure}
    \hfill
    \begin{subfigure}[b]{0.32\textwidth}
        \includegraphics[width=\textwidth]{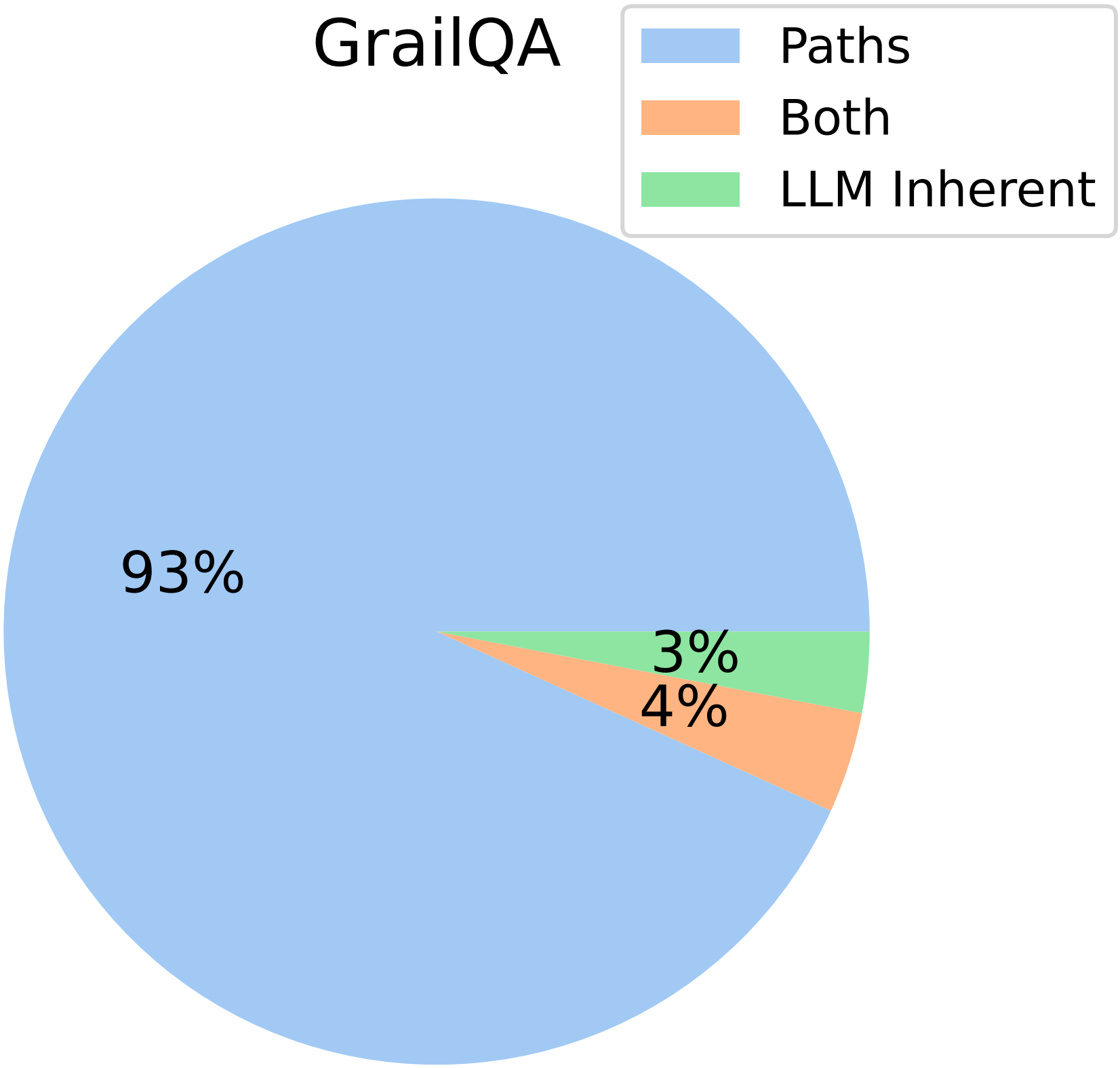}
    \end{subfigure}
    \caption{The proportions of ToG's evidence of answers on CWQ, WebQSP, and GrailQA datasets.}
    \label{fig: evidence_answers}
\end{figure}

\paragraph{Evidence of Answers}
We conducted an analysis of the correctly answered samples in three datasets to investigate the evidence for LLM in generating answers as shown in Figure \ref{fig: evidence_answers}. Evidently, a significant portion of the answers are derived from the paths explored by ToG, while roughly 20\% rely exclusively on the intrinsic knowledge embedded within LLM's parameters for generating responses. It is worth noting that around 7\% of the correctly answered samples require a combination of knowledge from both the explored paths and LLM's inherent knowledge (as elaborated in Appendix Table \ref{table: case3}). This distinction sets our approach apart from traditional graph-based search methods, as it does not necessitate the path to encompass the node containing the correct answer entirely. Instead, the explored paths supplement and reference LLM's inherent knowledge. The distribution of answer types for ToG-R is almost indistinguishable from that of ToG, proving the robustness of our approach.

\begin{figure}[htbp]
    \centering
    \begin{subfigure}[b]{0.32\textwidth}
        \includegraphics[width=\textwidth]{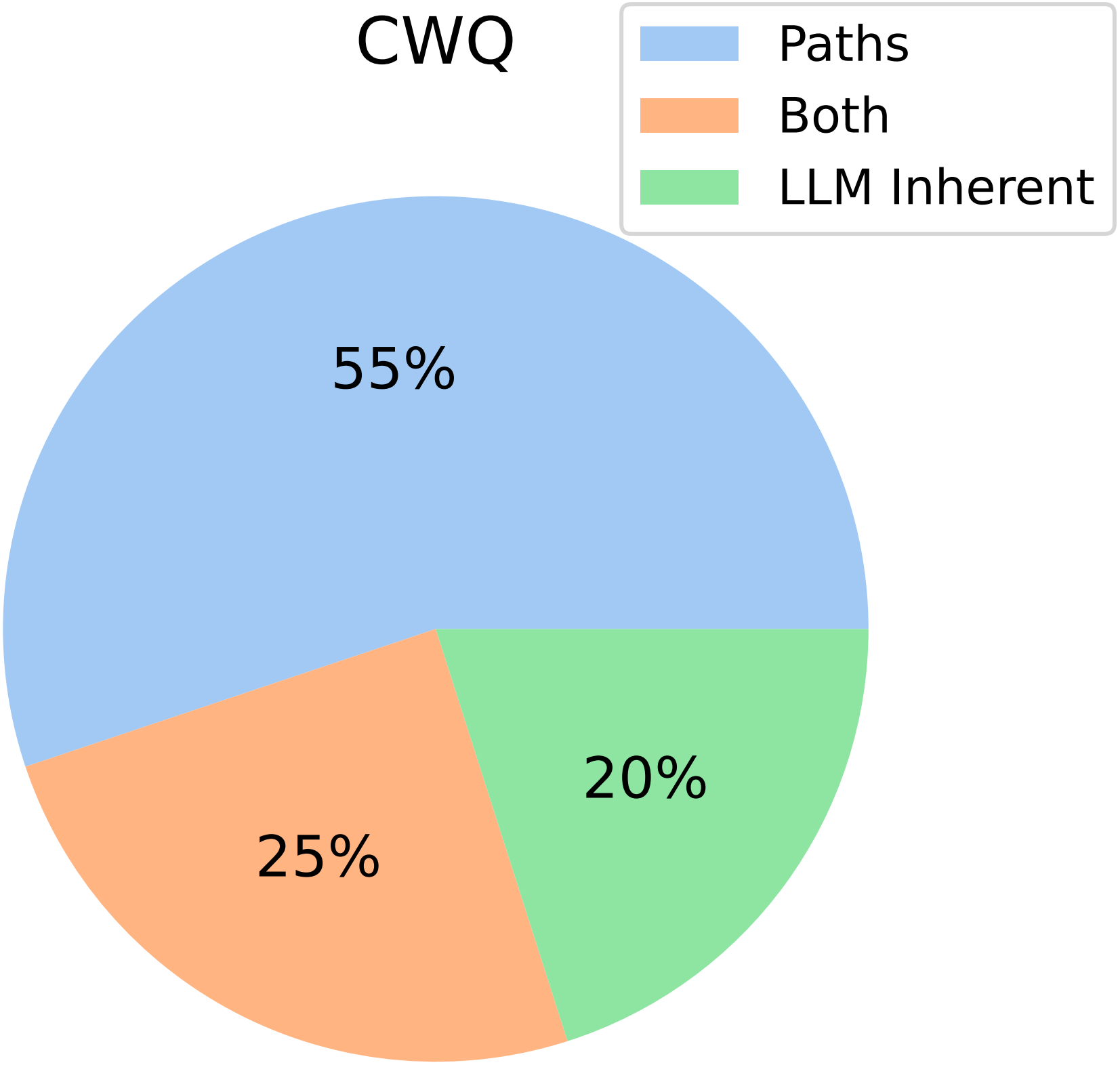}
    \end{subfigure}
    \hfill
    \begin{subfigure}[b]{0.32\textwidth}
        \includegraphics[width=\textwidth]{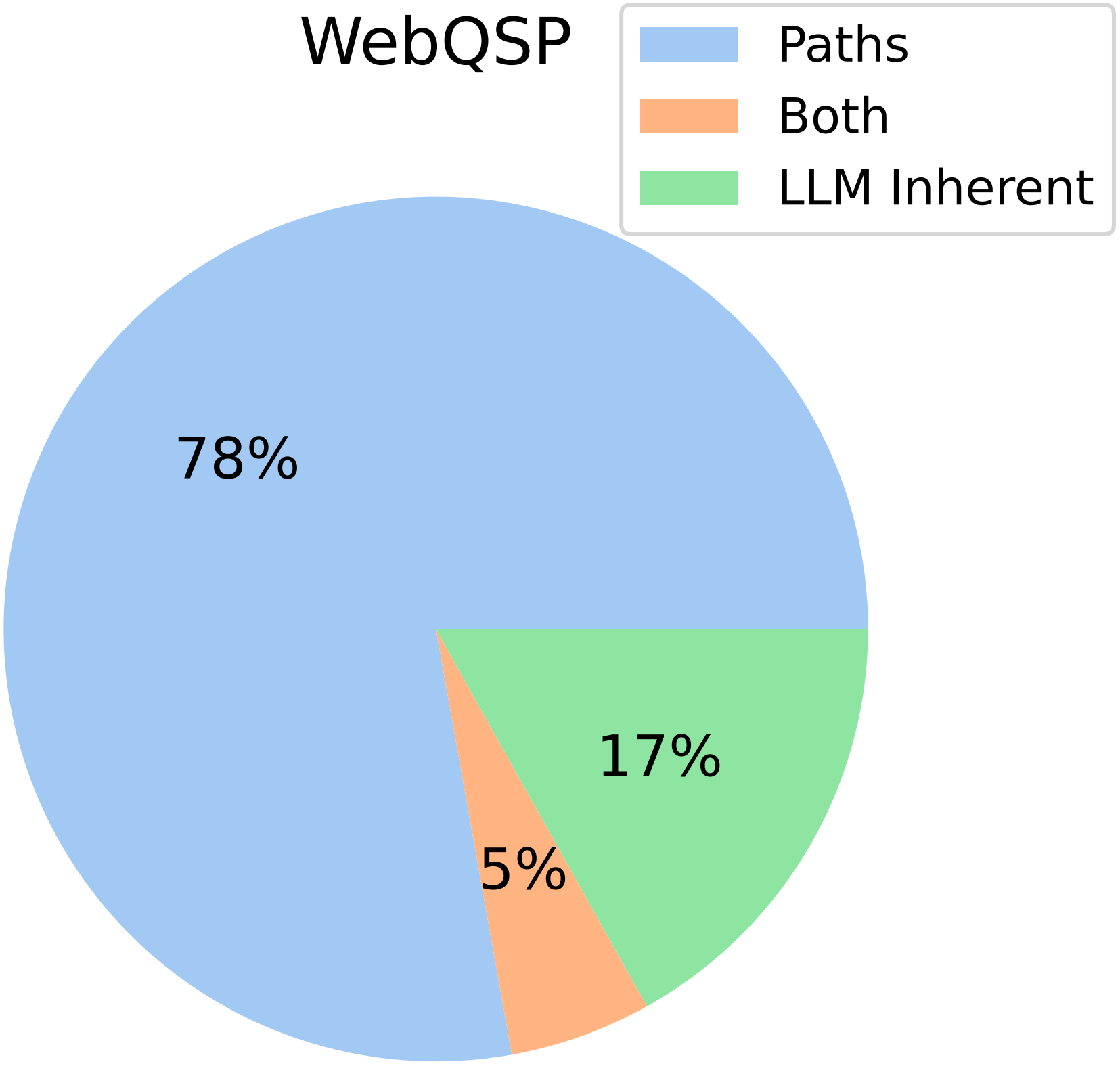}
    \end{subfigure}
    \hfill
    \begin{subfigure}[b]{0.32\textwidth}
        \includegraphics[width=\textwidth]{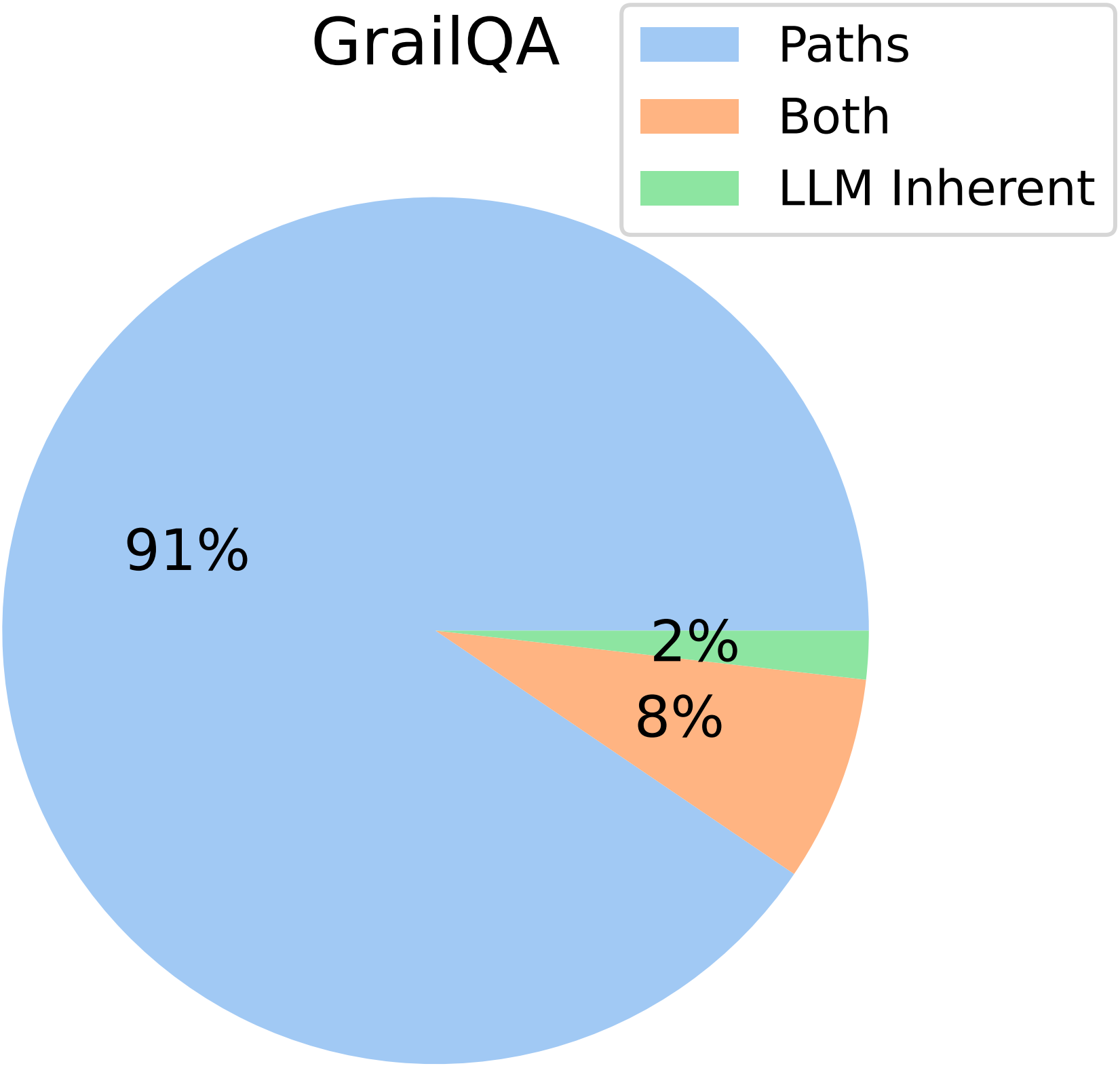}
    \end{subfigure}
    \caption{The explored path overlap ratio of ToG-R on CWQ, WebQSP, and GrailQA datasets.}
    \label{fig: evidence_answers_togr}
\end{figure}

\begin{figure}
  \centering
  \includegraphics[width=0.95\textwidth]{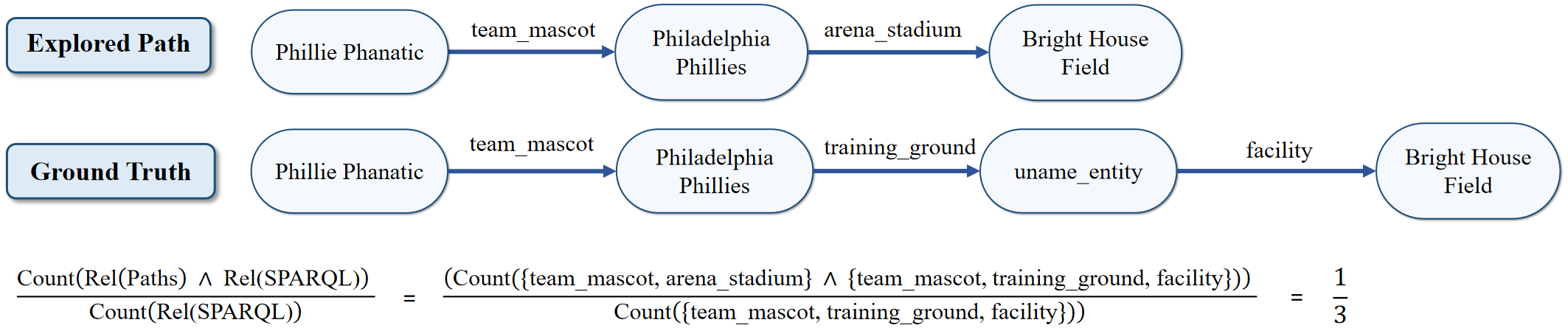}
  \caption{Path schematic to calculate overlap.}
  \label{fig: path_schematic}
\end{figure}

\begin{figure}[htbp]
    \centering
    \begin{subfigure}[b]{0.32\textwidth}
        \includegraphics[width=\textwidth]{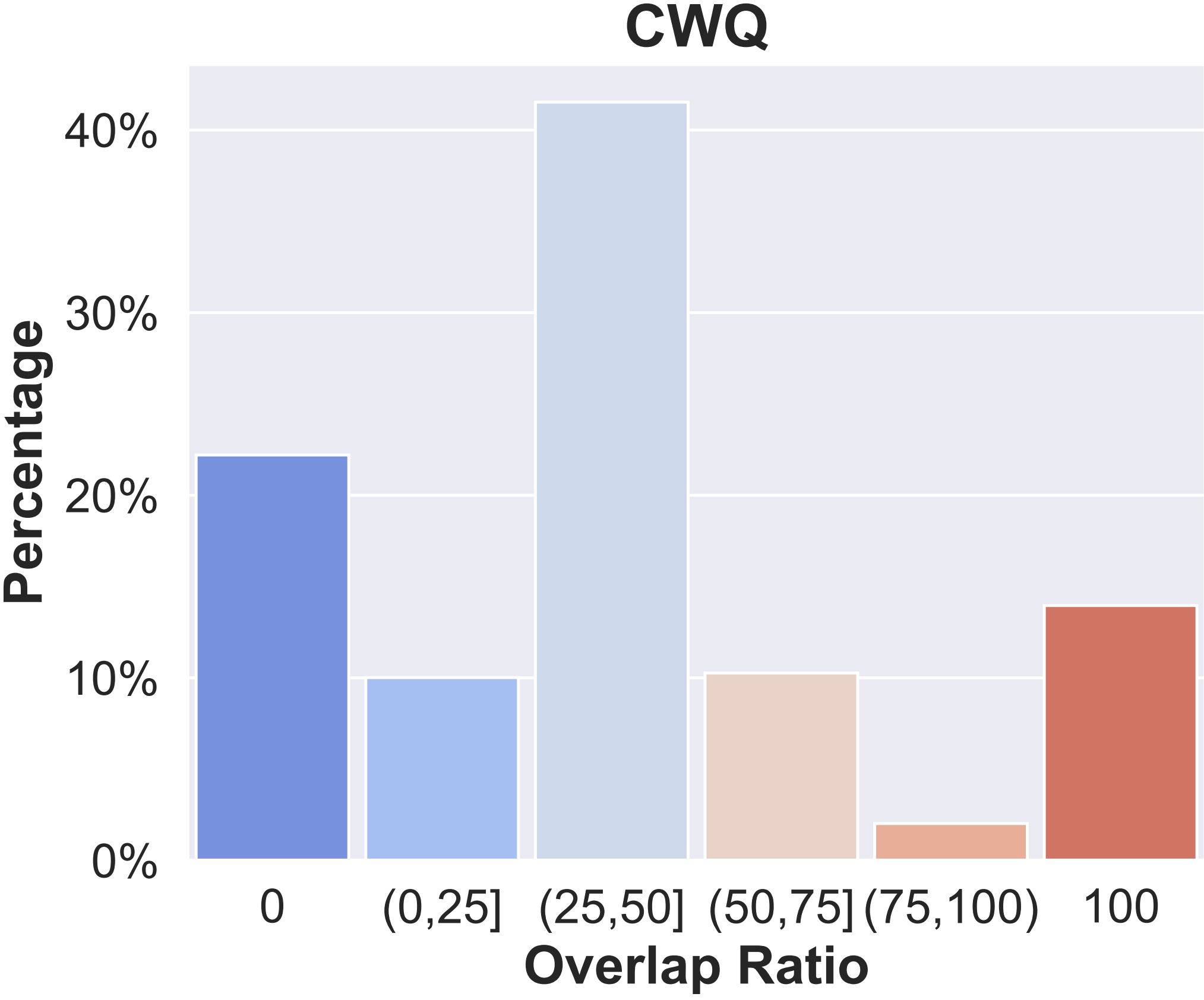}
    \end{subfigure}
    \hfill
    \begin{subfigure}[b]{0.32\textwidth}
        \includegraphics[width=\textwidth]{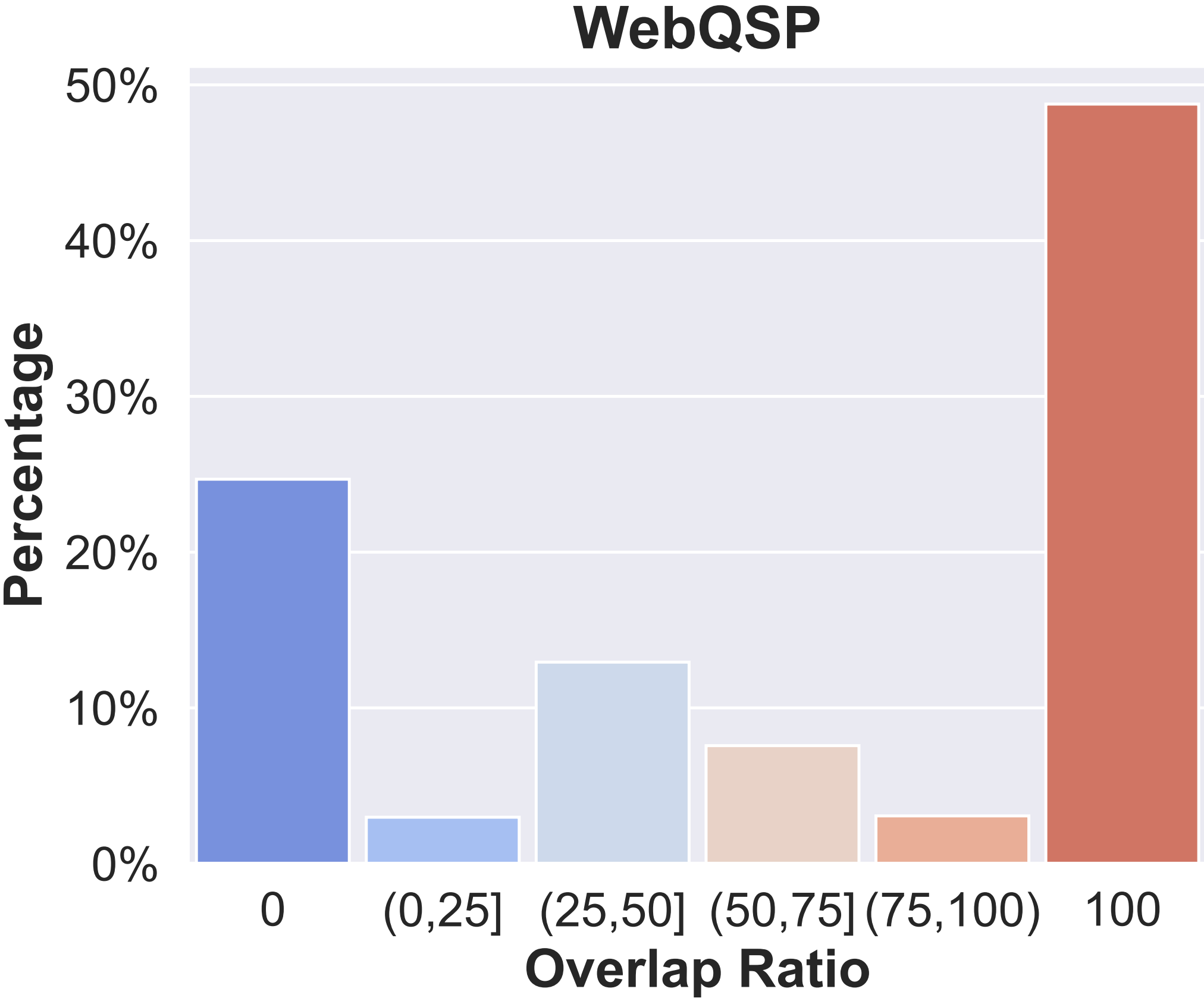}
    \end{subfigure}
    \hfill
    \begin{subfigure}[b]{0.32\textwidth}
        \includegraphics[width=\textwidth]{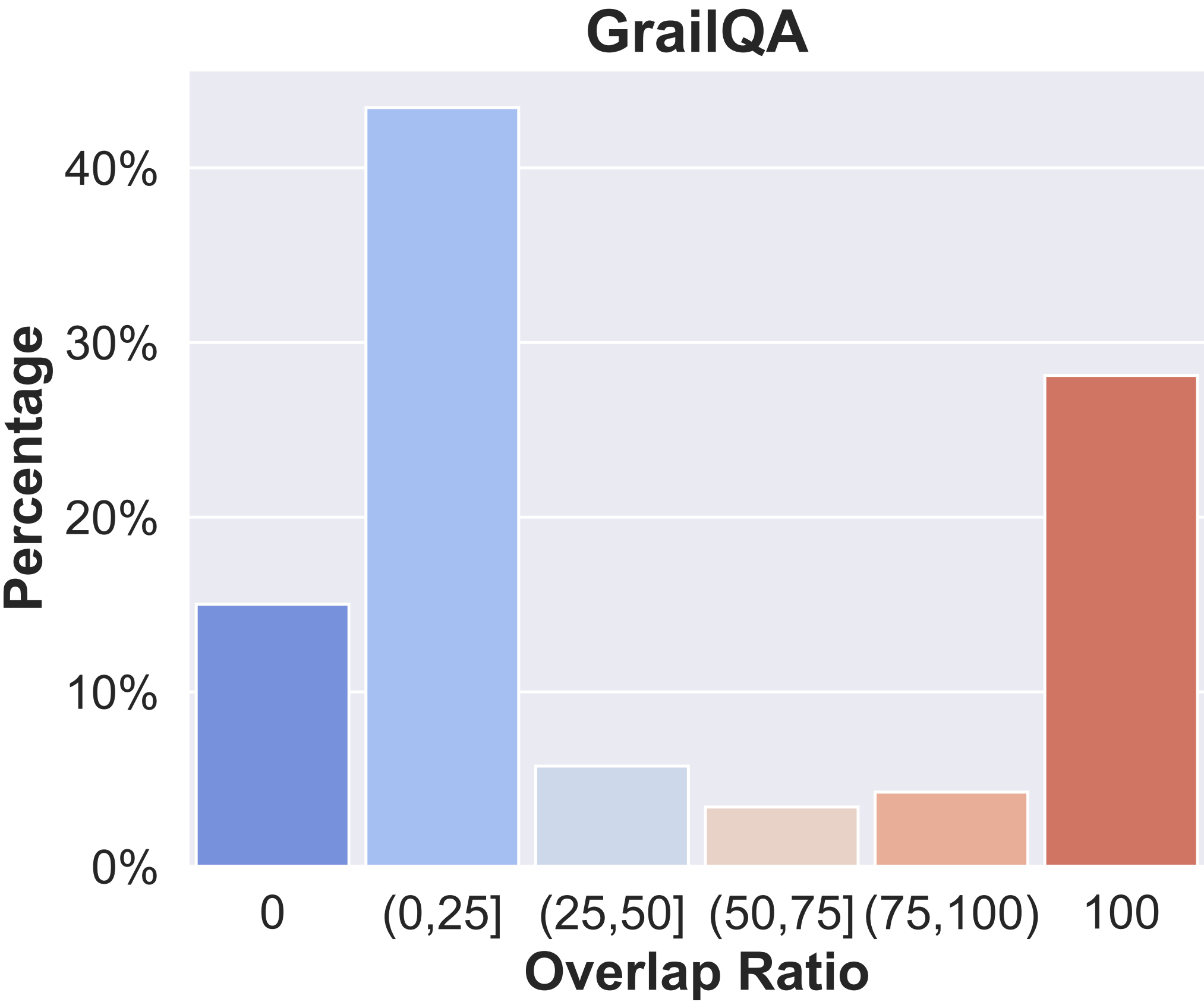}
    \end{subfigure}
    \caption{The explored path overlap ratio of ToG on CWQ, WebQSP, and GrailQA datasets.}
    \label{fig: overlap_ratio}
\end{figure}

\begin{figure}[htbp]
    \centering
    \begin{subfigure}[b]{0.32\textwidth}
        \includegraphics[width=\textwidth]{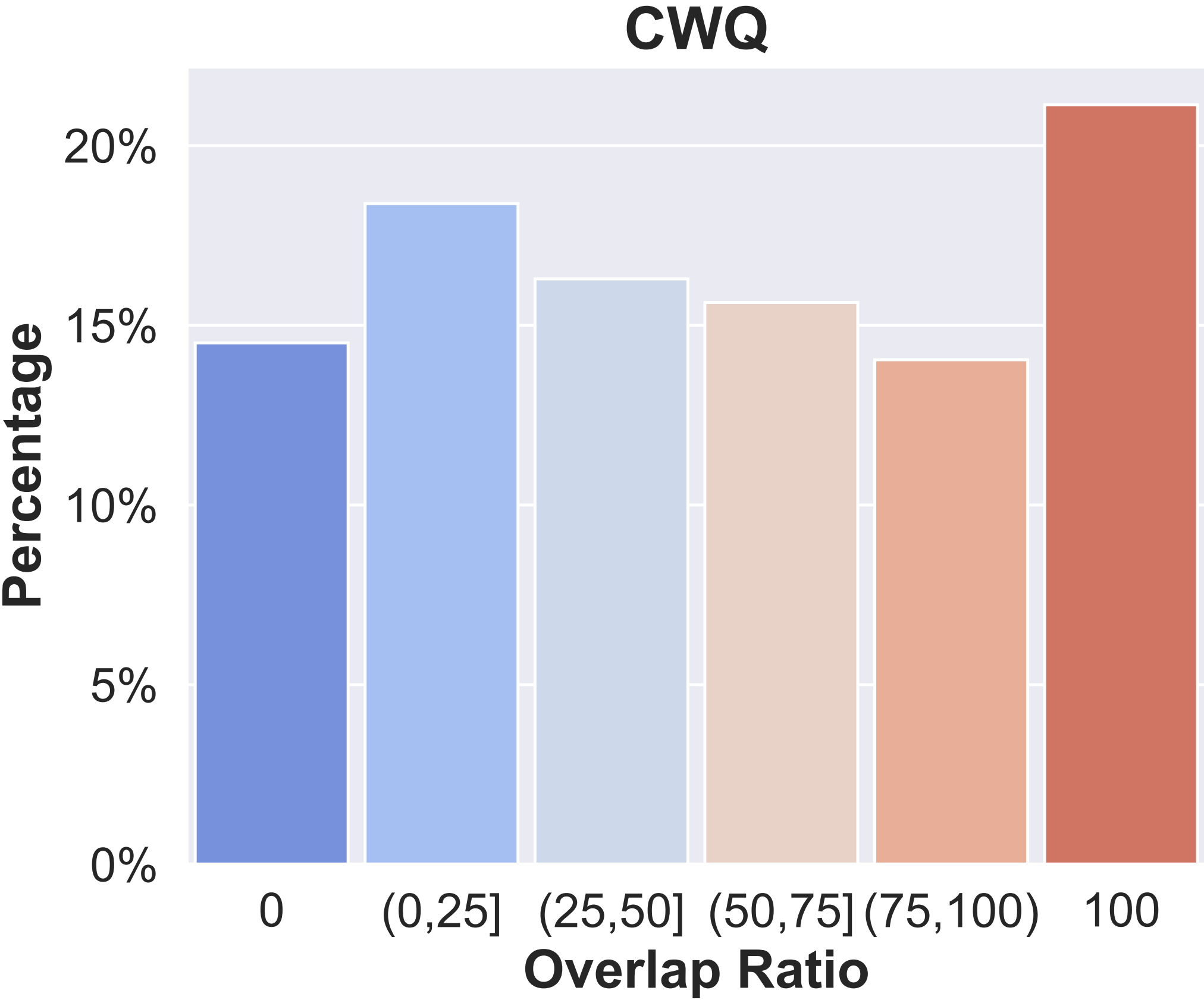}
    \end{subfigure}
    \hfill
    \begin{subfigure}[b]{0.32\textwidth}
        \includegraphics[width=\textwidth]{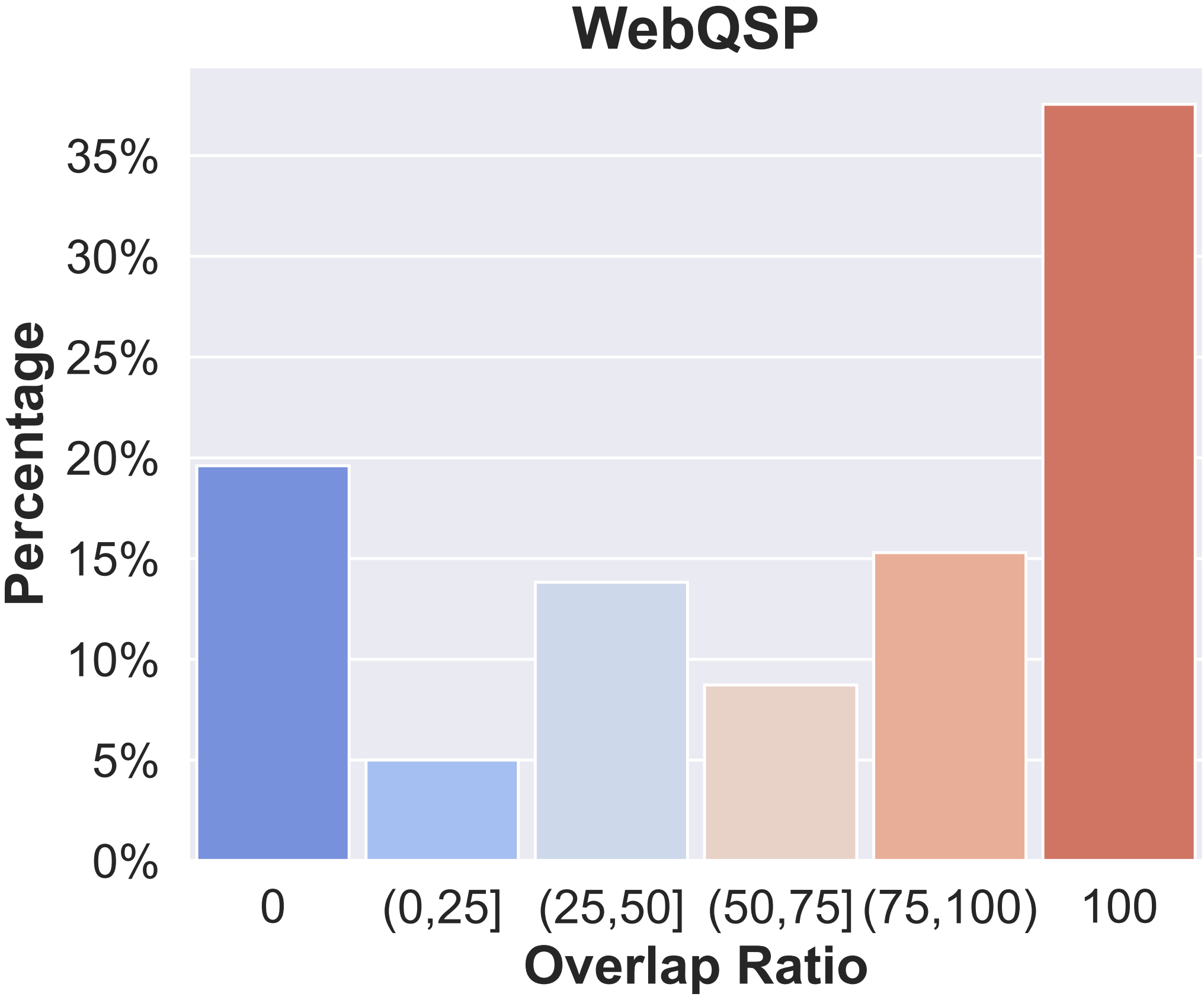}
    \end{subfigure}
    \hfill
    \begin{subfigure}[b]{0.32\textwidth}
        \includegraphics[width=\textwidth]{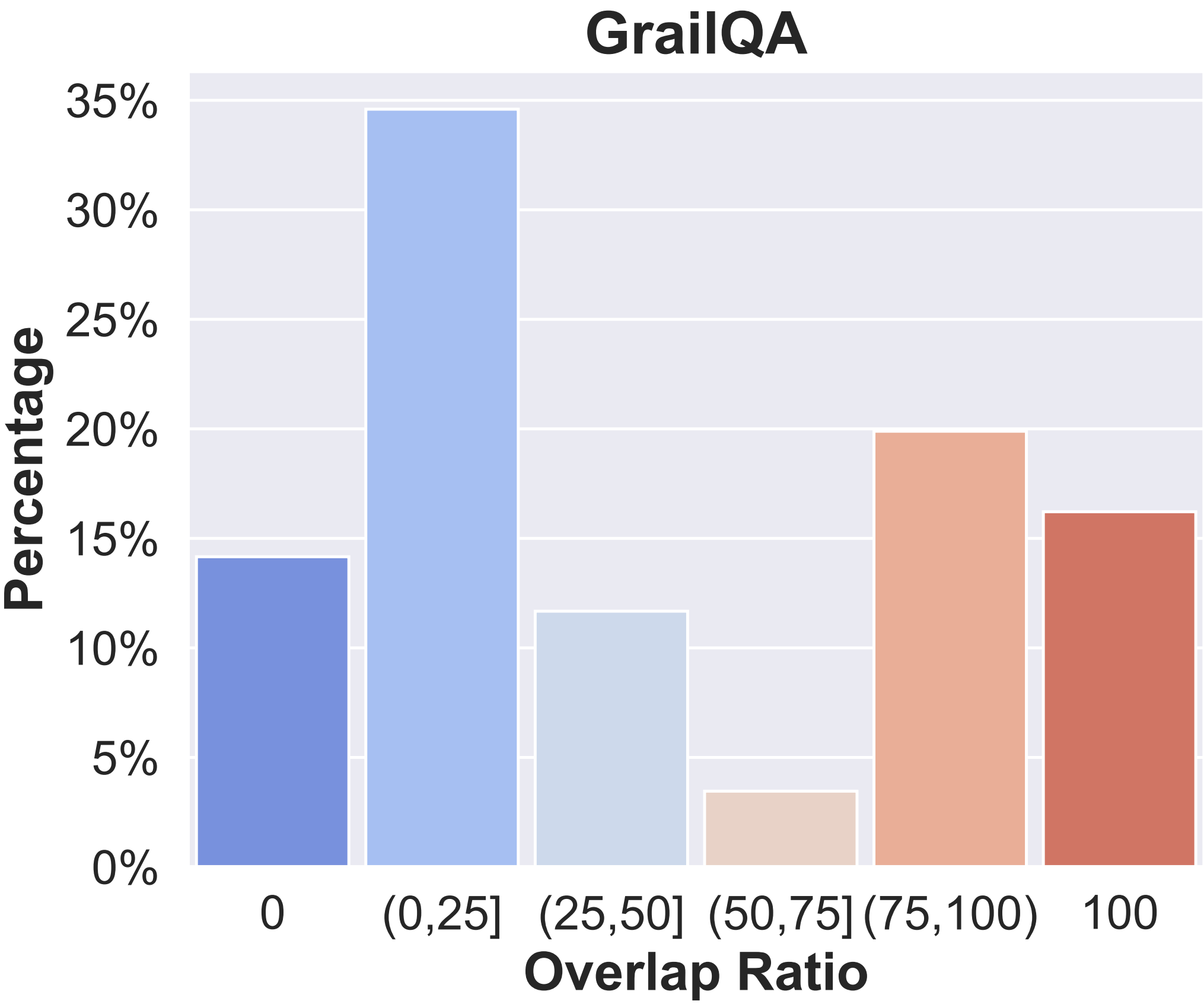}
    \end{subfigure}
    \caption{The path overlap ratio of ToG-R on CWQ, WebQSP, and GrailQA datasets.}
    \label{fig: overlap_ratio_togr}
\end{figure}

\paragraph{The Overlap Ratio between the Explored Paths and Ground-truth Paths}
\label{sec: overlap_ratio}
We also conduct an analysis of the correctly answered samples in three datasets to investigate the ratio of overlap between the paths explored by ToG and the ground-truth path in SPARQL. The definition of overlap ratio is the ratio of overlapping paths to the total number of relations in ground-truth SPARQL:
$$\frac{\text{Count}(\text{Rel}(\text{Paths}) \cap \text{Rel}(\text{SPARQL})) }{\text{Count}(\text{Rel}(\text{SPARQL}))}$$
where \text{Rel}(*) denotes all the unduplicated relations in the "*" and Count(*) denotes the number of "*"\footnote{We approximately calculate the length of a path by counting the number of relations in the ground-truth SPARQL.}. Figure \ref{fig: path_schematic} is a path schematic which takes the case shown in Table \ref{table: case4} for example.
It can be observed from Figure \ref{fig: overlap_ratio} that the paths explored by ToG are identical to the golden paths of an average of 30\% correct samples, while the paths of an average of 21\% correct samples are completely different from the golden path. This indicates that ToG has successfully explored a completely and approximately new path in the knowledge graph space to reach the final answer entity. For ToG-R, the disparity between the two is primarily evident in the CWQ dataset, where the percentage of intervals (25,50] in ToG results is quite significant (nearly 40\%), whereas ToG-R results tend to be more evenly distributed as shown in Figure \ref{fig: overlap_ratio_togr}. We contend that this discrepancy arises from the disregard of entity, thereby enhancing the diversity of explored relations. This represents a significant application of knowledge graph reasoning in academic research.

\paragraph{The Reasoning Depth of Questions}

\begin{figure}
  \centering
  \includegraphics[width=0.7\textwidth]{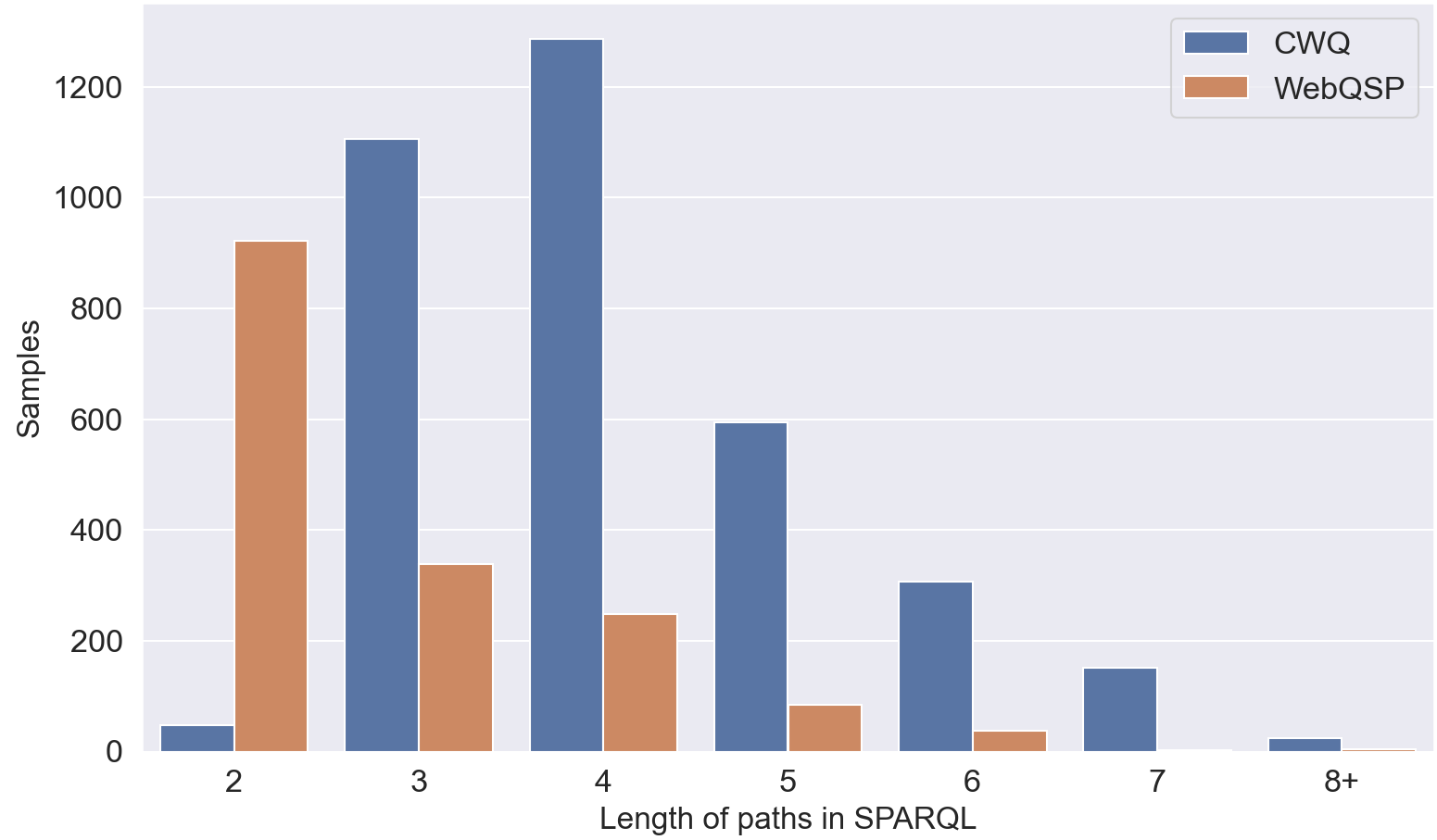}
  \caption{The lengths of the ground-truth SPARQL queries within the CWQ and WebQSP datasets, computed based on relation numbers.}
  \label{fig: cal_links}
\end{figure}

\begin{figure}[t]
    \centering
    \begin{subfigure}[b]{0.44\textwidth}
        \includegraphics[width=\textwidth]{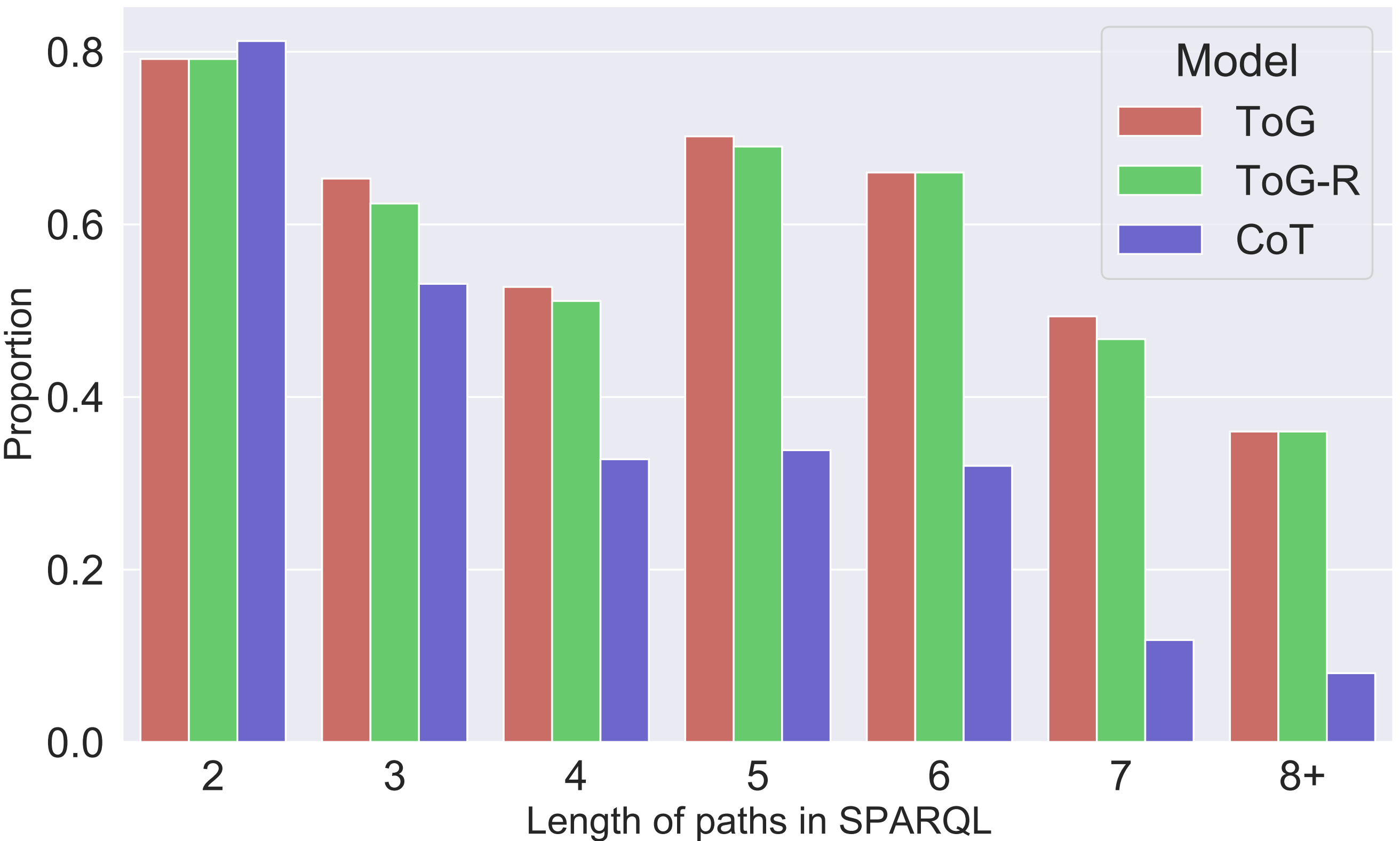}
    \end{subfigure}
    \begin{subfigure}[b]{0.44\textwidth}
        \includegraphics[width=\textwidth]{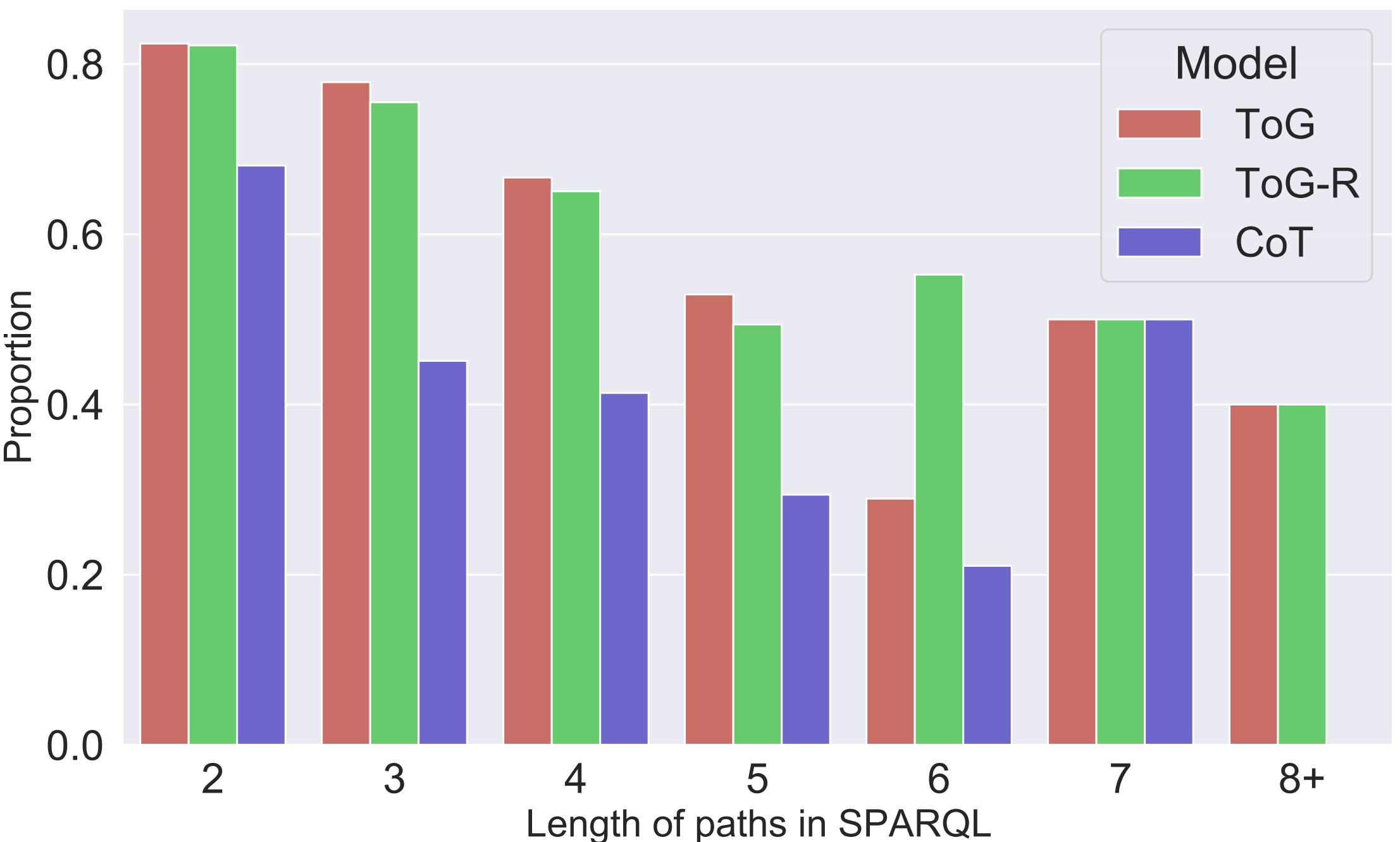}
    \end{subfigure}
    \caption{ToG, ToG-R and CoT's performance among CWQ and WebQSP dataset.}
    \label{fig: right_proportion}
\end{figure}

We calculate the reasoning depth of testing questions based on the number of relations within their ground-truth SPARQL queries on CWQ and WebQSP. The counts of questions with different reasoning depths are shown in Figure \ref{fig: cal_links}. We analyze the performances of ToG, ToG-R, and CoT on testing questions of both datasets with different reasoning depths. As illustrated in Figure~\ref{fig: right_proportion}, the performances of CoT show roughly decreasing trends on both datasets, with the reasoning depth of testing questions increasing. Conversely, ToG and ToG-R can partially counteract the performance degradation caused by the increment of reasoning depths of questions, especially on CWQ. Generally, the performance difference between ToG and CoT becomes more significant on deeper questions.



\subsubsection{Efficiency of ToG}
There are many solutions to improve efficiency and reduce the computational complexity (proportional to the number of calling LLMs) of ToG from the original $O(ND)$ to $O(D)$, where $D$ is the depth (or equivalently length) of the reasoning path, and $N$ is the width of the beam-search (how many paths are remained in the pool in each iteration).

\paragraph{Solution 1}
Reducing computational complexity from $O(ND)$ to $O(D)$ by using lightweight model in pruning. The bottleneck of computation is the pruning step, which contributes to $N*D$ times calling, and it is important to optimize it for computational efficiency. A technical route is to replace LLM with small models such as BM25 and Sentence-BERT in the pruning step since the small models are much faster than LLM calling. In this way, we can reduce the number of LLM calling from $2ND+D+1$ to $D+1$. When $D$=3, for example, there are only 4 times LLM calling. However, this optimization sacrifices the accuracy due to the weaker scoring model in pruning. For instance, as shown in Table \ref{table: tools} of the manuscript, the performance of ToG on WebQSP drops from 76.2\% to 66.3\% after replacing ChatGPT with SentenceBERT for pruning. To alleviate the issue of the performance degradation, we can appropriately increase the search width to compensate the loss because increasing search width can improve the chance of the optimal path to be selected in the pool and it doesn't affect the number of LLM calling. To empirically verify this, we increase the search width from 3 to 5 and reevaluate ToG with SentenceBERT as the pruning model on WebQSP. The accuracy rises to from 66.3\% to 68.5\% and could be further improved with a greater width since the greater width would not cause an increase in the number of LLM calls.

\paragraph{Solution 2}
Reducing computational complexity from $O(ND)$ to $O(D)$ by unifying the prompts in the same pruning step. Another solution on speeding up the pruning step is to employ the LLM at once to score all components of N candidate sets for obtaining top-N candidates, instead of calling the LLM N times to score N candidate sets separately. Through this solution, either entity pruning step or relation pruning step only need 1 LLM call for each iteration. Thus, the maximum number of LLM calls per question needed for ToG and ToG-R would drop to $2D+D+1$ and $D+D+1$.

\paragraph{Solution 3}
Optimizing pruning step to make the actual calls of LLMs much less than the previously estimated $2ND+D+1$ and closer to some common prompting methods such as CoT-SC. For ToG and other LLM-based methods, the computational time (cost or complexity) in the inference phase mainly depends on how many times calling LLM. For each question, ToG needs at most $2ND + D + 1$ times. Meanwhile, ToG-R needs at most $ND+D+1$ times as mentioned in Section \ref{sec: methods}. 

Given the beam search width $N$ and maximal reasoning depth $D$, ToG's initialize the search from the entity mostly aligning with the keyword in question. In each iterative step of the reasoning path, ToG starts from each of the $N$ entities/relations (nodes/edges on knowledge graph) and searches all its neighboring relations/entities. Given the search width $N$, ToG always keep $N$ "most-likely" candidate reasoning paths in the pool, and thus there are always $N$ candidate entity sets $E^D_{cand,n}$ and $N$ candidate relation sets $R^D_{cand,n}$. Consequently, it needs $N$ LLM calls for entity pruning and $N$ calls for relation pruning, respectively, as well as one additional LLM call for reasoning (evaluating if the information from the current candidate paths are enough or not). We have to point it out that, for each of the $N$ starting entities, all its neighbor entities/relations are NOT scored one by one. On the contrary, all its neighbor entities/relations are "translated" into "one" prompt altogether and are sent to LLM, which output the top-$N$ candidates at one-time. Therefore, each starting entity only calls LLM once for pruning and so $N$ starting entities calls LLM $N$ times in one iterative step. Consequently, there are totally $2ND+D$ times calling after reasoning $D$ steps. In the end, there is an additional calling that "translate" the final path to user-understandable language and answer the user. Therefore, ToG requires $2ND+D+1$ LLM calls in total. Since most questions can be answered within 3 hops (means depth of reasoning path is 3), and the performance is usually good enough when the search width $N$=3 as we tested in Figure \ref{fig: depth_width}, the total number of LLM calling is $2\times 3\times 3+3+1=22$. So the computational time is about 21 times longer than that of LLM-only. With a similar performance to ToG, its variant ToG-R only calls LLM for $ND+D+1$ times by using random entity pruning instead of LLM-based entity pruning, saving nearly half of computational time.

$2ND+D+1$ is the maximal computational complexity. In most cases, ToG does not need $2ND+D+1$ LLM calls for a question because the whole reasoning process might be early stopped before the maximum reasoning depth D is reached if LLM determines enough information has been retrieved. Likewise, ToG-R does not really need $ND+D+1$ LLM calls in most cases. As an illustration, Table \ref{tab:llm-calls-part1} and Table \ref{tab:llm-calls-part2} show the average numbers of LLM calls per question needed by ToG on different datasets. It can be seen that in the four multi-hop KBQA datasets, the average numbers of LLM calls (ranging from 10 to 15) are significantly smaller than 22, which is the theoretical maximum number of LLM calls calculated from $2ND+D+1$ when $N$=3 and $D$=3. We can also see that this AVERAGE number gets even smaller (< 10) for single-hop reasoning datasets, such as SimpleQuestion and T-REx.

\begin{table}
\centering
\begin{tabular}{c|c|c|c|c|c}
\hline
Dataset & CWQ & WebQSP & GrailQA & QALD10-EN & SimpleQuestion \\

Average LLM Calls & 14.3 & 11.2 & 10.4 & 11.4 & 8.7 \\
\hline
\end{tabular}
\caption{Average Number of LLM Calls per Question (Part 1)}
\label{tab:llm-calls-part1}
\end{table}

\begin{table}
\centering
\begin{tabular}{c|c|c|c|c}
\hline
Dataset & WebQuestion & T-REx & Zero-Shot RE & Creak \\
Average LLM Calls & 10.5 & 7.7 & 7.6 & 8.0 \\
\hline
\end{tabular}
\caption{Average Number of LLM Calls per Question (Part 2)}
\label{tab:llm-calls-part2}
\end{table}

\section{Dataset}~\label{sec: datasets}
The statistics of the datasets used in this paper are shown in Table \ref{table:description}.
We also provide a detailed result table for each dataset, shown in Table \ref{tab: cwq} to Table \ref{tab: creak}, illustrating the enhancements of ToG compared to the previous fine-tuning-based and prompting-based relevant works.
For QALD10-en, WebQuestions, Zero-Shot RE, and Creak, ChatGPT-based ToG reached a new state-of-the-art. Furthermore, GPT-4-based ToG exceeded the fine-tuning-based approaches on almost all Multi-Hop KBQA datasets, where on CWQ, ToG is close to the state-of-the-art (69.5\%).

\begin{table}[htbp]
\centering
\resizebox{\linewidth}{!}{
\begin{tabular}{lccccc}
\hline
Dataset         & Answer Format    & Train               & Test & Licence     \\ \hline

\multicolumn{1}{l|}{ComplexWebQuestions}           & \multicolumn{1}{c|}{Entity}           & \multicolumn{1}{c|}{27,734}  & \multicolumn{1}{c|}{3,531} & - \\
\multicolumn{1}{l|}{WebQSP}           & \multicolumn{1}{c|}{Entity/Number}           & \multicolumn{1}{c|}{3,098} & \multicolumn{1}{c|}{1,639} &  CC License           \\
\multicolumn{1}{l|}{GrailQA*}          & \multicolumn{1}{c|}{Entity/Number}          & \multicolumn{1}{c|}{44,337} & \multicolumn{1}{c|}{1,000}  & -           \\
 \multicolumn{1}{l|}{QALD-10}            & \multicolumn{1}{c|}{Entity/Number}  & \multicolumn{1}{c|}{-}    & \multicolumn{1}{c|}{333}  & MIT License  \\
\multicolumn{1}{l|}{Simple Quesiton*}           & \multicolumn{1}{c|}{Entity/Number}           & \multicolumn{1}{c|}{14,894} & \multicolumn{1}{c|}{1,000} & CC License\\
 \multicolumn{1}{l|}{WebQuestions}        & \multicolumn{1}{c|}{Entity/Number}           & \multicolumn{1}{c|}{3,778} & \multicolumn{1}{c|}{2,032}  & -           \\
\multicolumn{1}{l|}{T-REx}            & \multicolumn{1}{c|}{Entity}  & \multicolumn{1}{c|}{2,284,168}  & \multicolumn{1}{c|}{5,000} & MIT License           \\
\multicolumn{1}{l|}{Zero-Shot RE}      & \multicolumn{1}{c|}{Entity}         & \multicolumn{1}{c|}{147,909}  & \multicolumn{1}{c|}{3,724} & MIT License  \\
\multicolumn{1}{l|}{Creak}            & \multicolumn{1}{c|}{Bool}  & \multicolumn{1}{c|}{10,176} & \multicolumn{1}{c|}{1,371}  & MIT License  \\\hline
\end{tabular}
}
\caption{The statistics of the datasets used in this paper. * denotes we randomly selected 1,000 samples from the GrailQA and Simple Questions test set to constitute the testing set owing to the abundance of test samples.}
\label{table:description}
\end{table}

\begin{table}[]
\centering
\begin{tabular}{llc}
\hline
       \textbf{Model}    & \textbf{Method} &  \textbf{EM} \\ \hline
\multirow{3}{*}{Fine-Tuning}     
 &  QGG (Query Graph Generator) \citep{lan2020query}                    &  44.1 	   \\
 &   PullNet \citep{sun2019pullnet}                    & 45.9  \\
 &  NSM+h	 \citep{NSM}                       & 53.9	 \\
                              &  CBR-KBQA \citep{cwq1}	& 67.1	\\     
                                 &  	DecAF 	 
 \citep{webqsp1}                    & 70.4		 \\

                                 \hline
\multirow{3}{*}{ChatGPT}          
                                 &  KD-CoT \citep{wang2023knowledgedriven}           &  49.2  \\
                                 &  ToG                 & 57.1  \\
                                  &  ToG-R                 & \textbf{58.9}   \\
                                  \hline

\multirow{2}{*}{Llama2-70B-Chat} 

 &  ToG                   & 53.6   \\
&  ToG-R                   &  \textbf{57.6}  \\
                                 \hline
\multirow{2}{*}{GPT-4}          
                                
                                 &  ToG                 & 67.6   \\  &  ToG-R                     &  \textbf{69.5}  \\ \hline
\end{tabular}

\caption{The statics of Fine-Tuning, prompting-based methods of ComplexWebQuestions dataset.}
\label{tab: cwq}
\end{table}

\begin{table}
\centering
\begin{tabular}{llc}
\hline
       \textbf{Model}    & \textbf{Method} &  \textbf{EM} \\ \hline
\multirow{3}{*}{Fine-Tuning}     &  KD-CoT \citep{wang2023knowledgedriven}              &  73.7 \\
                                &  NSM \citep{NSM}       & 74.3   \\  
                                 &  Program Transfer \citep{program_transfer}                       & 74.6 \\
                                 &  TIARA \citep{grailqa1}                    & 75.2 \\
                                 &  DecAF \citep{webqsp1}                    &  82.1   \\ \hline
Code-davinci-002                 &  KB-BINDER \citep{DB-BLINDER}            &  74.4  \\ \hline
\multirow{3}{*}{ChatGPT}           &  StructGPT \citep{structgpt}              &  72.6  \\       
                               &  ToG-R              &  75.8  \\ 
                               &  ToG               &  \textbf{76.2}  \\ \hline
\multirow{2}{*}{Llama2-70B-Chat} &  ToG-R                   &  \textbf{69.4}  \\
                                 &  ToG                   & 64.1   \\ \hline
\multirow{2}{*}{GPT-4}          &  ToG-R                     &  81.9  \\
                                 &  ToG                 & \textbf{82.6}   \\ \hline
\end{tabular}
\caption{The statics of Fine-Tuning, prompting-based methods of WebQSP dataset.}
\label{tab: webqsp}
\end{table}

\begin{table}[]
\centering
\begin{tabular}{llc}
\hline
       \textbf{Model}    & \textbf{Method} &  \textbf{EM} \\ \hline
\multirow{4}{*}{Fine-Tuning} 

&  	DecAF 
 \citep{webqsp1}                    & 68.4		 \\
 &  UniParser 
 \citep{liu2022uni}                    &  69.5	   \\
 &  TIARA \citep{grailqa1}                    & 73.0 \\

                                 &  Pangu   \citep{gu2023dont}                       & 75.4	 \\

                                  \hline
Code-davinci-002                 &  KB-BINDER \citep{DB-BLINDER}            &  53.2  \\ \hline
\multirow{2}{*}{ChatGPT}          
                                 &  ToG-R                     &  66.4  \\
                                 &  ToG                 & \textbf{68.7}   \\ \hline
\multirow{2}{*}{GPT-4}          
                                 &  ToG-R                     &  80.3  \\
                                 &  ToG                 & \textbf{81.4}   \\ \hline
\end{tabular}

\caption{The statics of Fine-Tuning, prompting-based methods of GrailQA dataset}
\label{tab: grailqa}
\end{table}

\begin{table}[]
\centering
\begin{tabular}{llc}
\hline
       \textbf{Model}    & \textbf{Method} &  \textbf{Acc} \\ \hline
\multirow{1}{*}{Fine-Tuning}     &  SPARQL-QA\citep{qald10-en1} 	& 45.4 	\\ \hline
\multirow{2}{*}{ChatGPT}          &  ToG-R                     &  48.6  \\
                                 &  ToG                 & \textbf{50.2}   \\ \hline
\multirow{2}{*}{GPT-4}          &  ToG                     &   53.8\\
                                 &  ToG-R               & \textbf{54.7}   \\ \hline
\end{tabular}

\caption{The statics of Fine-Tuning, prompting-based methods of QALD10-en dataset.}
\label{tab: qald}
\end{table}

\begin{table}[]
\centering
\begin{tabular}{llc}
\hline
       \textbf{Model}    & \textbf{Method} &  \textbf{EM} \\ \hline
\multirow{3}{*}{Fine-Tuning}     
&  T5-LARGE+KPs \citep{santos2022knowledge} & 58.3	\\
&  Memory Networks
 \citep{simple_questions_dataset}	& 63.9 	
 
   \\  
   &  GETT-QA \citep{banerjee2023gettqa}	& 76.1	\\
  & DiFaR\citep{simple_questions1}	& \textbf{85.8}	\\ \hline
\multirow{2}{*}{ChatGPT}          
                                 &  ToG-R                     &  45.4  \\
                                 &  ToG                 & \textbf{53.6}   \\ \hline
\multirow{2}{*}{GPT-4}                            &  ToG-R                     &  58.6  \\
                                 &  ToG                 & \textbf{66.7}   \\ \hline
\end{tabular}

\caption{The statics of Fine-Tuning, prompting-based methods of SimpleQuetsions dataset.}
\label{tab: simpleqa}
\end{table}

\begin{table}[]
\centering
\begin{tabular}{llc}
\hline
       \textbf{Model}    & \textbf{Method} &  \textbf{EM} \\ \hline
\multirow{3}{*}{Fine-Tuning}     
 &  T5.1.1-XXL+SSM 

 \citep{raffel2020exploring}                    &  43.5 \\
&  PaLM \citep{palm}                    & 43.5 \\
&  RAG \citep{lewis2021retrievalaugmented}                    & 45.2 \\
&  FiDO  \citep{de2022fido}                       & 51.1 \\
&  FiE+PAQ	 \citep{webq1}       & 56.3 \\

                                	    \hline
PALM2                 &  Few-shot \citep{DB-BLINDER}            &  28.2  \\ \hline
\multirow{3}{*}{ChatGPT} 
&  $\text{BeamSearchQA}_{\text{Fine-tuned Retriever}}$ \citep{beamsearchqa}            &  27.3  \\
                                 &  ToG-R                   &  53.2  \\
                                 &  ToG                   & \textbf{54.5}   \\ \hline
\multirow{3}{*}{GPT-4} 
                                 &  ToG-R                   &  57.1  \\
                                 &  ToG                   & \textbf{57.9}   \\ \hline
\end{tabular}

\caption{The statics of Fine-Tuning, prompting-based methods of WebQuestions dataset.}
\label{tab: webq}
\end{table}

\begin{table}[htbp]
\centering
\begin{minipage}{0.45\textwidth}
\centering
\begin{tabular}{llc}
\hline
\textbf{Model} & \textbf{Method} & \textbf{EM} \\ \hline
\multirow{4}{*}{Fine-Tuning} & MetaRAG & 78.7 \\
& Wikipedia & 81.3 \\
& single ngram & 83.7 \\
& KGI\_1 & 84.4 \\
& Re2G \citep{trex1} & 87.7 \\
\hline
\multirow{2}{*}{ChatGPT} & ToG-R & 75.3 \\
& ToG & \textbf{76.8} \\ \hline
\multirow{2}{*}{GPT-4} & ToG-R & 75.5 \\
& ToG & \textbf{77.1} \\ \hline
\end{tabular}
\caption{The statics of Fine-Tuning, prompting-based methods of T-REx dataset, where data are from the leaderboard.}
\label{tab: trex}
\end{minipage}
\hfill
\begin{minipage}{0.45\textwidth}
\centering
\begin{tabular}{llc}
\hline
\textbf{Model} & \textbf{Method} & \textbf{EM} \\ \hline
\multirow{4}{*}{Fine-Tuning} & Multitask DPR + BART & 58.0 \\
& MetaRAG & 71.6 \\
& KGI\_1 & 72.6 \\
& Wikipedia & 74.0 \\
& single ngram  & 74.6 \\
\hline
\multirow{2}{*}{ChatGPT} & ToG-R & 86.5 \\
& ToG & \textbf{88.0} \\ \hline
\multirow{2}{*}{GPT-4} & ToG-R & 86.9 \\
& ToG & \textbf{88.3} \\ \hline
\end{tabular}
\caption{The statics of Fine-Tuning, prompting-based methods of Zero-Shot RE, where data are from the leaderboard.}
\label{tab: zeroshotre}
\end{minipage}
\end{table}

\begin{table}[]
\centering
\begin{tabular}{llc}
\hline
       \textbf{Model}    & \textbf{Method} &  \textbf{EM} \\ \hline
\multirow{3}{*}{Fine-Tuning}     	&   RoBERTa-Large	 \citep{liu2019roberta}         & 80.6	  	\\
                                 &  	T5-3B	 \citep{raffel2020exploring}                    & 85.6	\\
                                 &  RACo-Large		 \citep{creak1}    &    88.2		 \\\hline
\multirow{2}{*}{ChatGPT}          &  ToG-R                     &  \textbf{93.8}  \\
                                 &  ToG                 & 91.2   \\ \hline
\multirow{2}{*}{GPT-4}          &  ToG-R                     &  95.4  \\
                                 &  ToG                 & \textbf{95.6}   \\ \hline
\end{tabular}

\caption{The statics of Fine-Tuning, prompting-based methods of Creak dataset.}
\label{tab: creak}
\end{table}

\newpage

\section{Case Study}
\label{sec: case_study}
In this section, we present a case analysis of the CWQ dataset to evaluate the utility and limitations of the ToG. We compared ToG with IO, CoT and the New Bing search engine\footnote{Accessed version July 2023.}. We have selected four examples for analysis, each with top-3 reasoning paths and normalized scores. 

In the first example in Table \ref{table: case1}, ToG initially identifies "Arthur Miller" and "Lucian", in the question and subsequently expands its reasoning path through the \texttt{Exploration} and \texttt{Reasoning} processes. After conducting two iterations of the search, ToG successfully arrived at the correct answer, as it links the two entities with the reasoning path, which represents the perfect route for locating solutions. Additionally, the presence of \textsl{UnName\_Entity} in the intermediate steps of reasoning paths, reflects the incompleteness of the knowledge graph (i.e., some entities lack the "name" relation). However, ToG is still capable of performing the next reasoning step, as all available relations contain relevant information. We observe that IO and CoT do not answer the query correctly since they lack the appropriate knowledge, and New Bing do not retrieve the appropriate information during the retrieval process.

In the second example shown in Table \ref{table: case2}, IO prompt and CoT even New Bing suffer from a hallucination issue and provide an erroneous answer, "Florida", since the "Renegade" is the mascot of "Florida State Seminoles" instead of "fight song". ToG obtain the reasoning path "Renegade" $\to$ "sports.fight\_song.sports\_team" $\to$ "Pittsburgh Steeler". However, this reasoning path does not lead to a final answer, but combined with LLMs', ToG can answer the correct answer "Pennsylvania".

The third example in Table \ref{table: case3} demonstrates an example of the ToG-R, where ToG ignores the intermediate entities and focuses on the information in the relations instead. After two-hop of reasoning to "Harvard College", combined with LLMs', ToG gives the final result: "Massachusetts". It can be observed that IO and CoT do not have background knowledge, and New Bing answers the question correctly since it retrieves the correct information.

The final example is shown in Table \ref{table: case4}. Where ToG generates a reasoning path to the final question (Path 1). Notably, the Ground-Truth reasoning path for the answer is \textit{sports.sports\_team.team\_mascot} $\to$ \textit{base.schemastaging.team\_training\_ground\_relationship.facility} $\to$ \textit{base.schemastaging.sports\_team\_extra.training\_ground} (retrievable from the SPARQL), which is more hop than ToG. The ToG enables the exploration of new reasoning paths to reach the correct answer, which represents a significant application of knowledge graph reasoning. However, the answer to the current question in the KB, is "Bright House Field", which is incorrect since "Philadelphia Phillies" training stadium is "Spectrum Field" now. This example exemplifies a constraint of ToG, specifically its dependence on the correctness of the KB, where the incorrect KB has negative impact on ToG's reasoning accuracy. However, as depicted in Figure \ref{fig: app}, ToG presents a novel framework to construct automated knowledge infusion to the KG.

\begin{table}[htbp]
\renewcommand\arraystretch{1.5}

\centering
\begin{tabularx}{\textwidth}{p{2cm}|p{11cm}}
\hline
Question & Who influenced Arthur Miller that was influenced by Lucian? \\ \hline
\multirow{1}{*}{\begin{tabularx}{\linewidth}{@{}l@{}}Reasoning \\ Paths \\ \end{tabularx}} & Arthur Miller $\to$ \textit{influence.influence\_node.influenced\_by} $\to$ William Shakespeare $\to$ \textit{influence.influence\_node.influenced\_by} $\to$ Lucian. (\textbf{Path 1}, Score: 0.75) \\
& Lucian $\to$ \textit{influence.influence\_node.influenced\_by} $\to$ Socrates $\to$ \textit{influence.influence\_node.influenced\_by} $\to$ Parmenides. (\textbf{Path 2}, Score: 0.2) \\
& Arthur Miller $\to$ \textit{people.person.education} $\to$ \textbf{UnName\_Entity} $\to$ \textit{education.education.student} $\to$ Arthur Miller. (\textbf{Path 3}, Score: 0.05) \\ \hline
\multirow{1}{*}{\begin{tabularx}{\linewidth}{@{}l@{}} Outputs \\ \end{tabularx}} & \textbf{IO}: \textcolor{red}{Edmund Wilson}. \\
& \textbf{CoT}: Arthur Miller was influenced by the playwright and director \textcolor{red}{Thornton Wilder}, who in turn was influenced by Lucian of Samosata. \\
& \textbf{New Bing}: I'm not sure about Lucian's influence on Arthur Miller. However, Arthur Miller was influenced by \textcolor{red}{Henrik lbsen and Eugene O'Neill}. He was also influenced by the events of the McCarthy era and the Salem witch trials. \\ 
& \textbf{ToG}: Based on the given knowledge triples, we can infer that Arthur Miller was influenced by \textcolor{blue}{William Shakespeare}, who was directly influenced by Lucian. \\ \hline
Ground Truth & \textcolor{blue}{William Shakespeare}. \\ \hline
\end{tabularx}
\caption{Case for ToG. The \textcolor{red}{wrong answers} are highlighted with red color, and the \textcolor{blue}{correct answers} are highlighted with blue color.}
\label{table: case1}
\end{table}

\begin{table}[htbp]
\renewcommand\arraystretch{1.5}

\centering
\begin{tabularx}{\textwidth}{p{2cm}|p{11cm}}
\hline
Question & What is the state where the team whose fight song is "Renegade" is from? \\ \hline
\multirow{1}{*}{\begin{tabularx}{\linewidth}
{@{}l@{}}Reasoning \\ Paths \\ \end{tabularx}} & Renegade $\to$ \textit{sports.fight\_song.sports\_team} $\to$ Pittsburgh Steelers. (\textbf{Path 1}, Score: 0.67) \\
& Renegade $\to$ \textit{sports.sports\_team.fight\_song} $\to$ \textbf{UnName\_Entity}. (\textbf{Path 2}, Score: 0.33) \\ \hline
\multirow{1}{*}{\begin{tabularx}{\linewidth}{@{}l@{}} Outputs \\ \end{tabularx}} & \textbf{IO}: \textcolor{red}{Florida}. \\
& \textbf{CoT}: The team whose fight song is "Renegade" is the Florida State Seminoles, and they are from the state of \textcolor{red}{Florida} in the United States. \\
& \textbf{New Bing}: The team whose fight song is "Renegade" is the Florida State Seminoles. They are from the state of
\textcolor{red}{Florida} in the United States. \\ 
& \textbf{ToG}: Based on the given knowledge triples, the team with the fight song "Renegade" is the Pittsburgh Steelers. The Pittsburgh Steelers are a professional American football team, and they are from the state of \textcolor{blue}{Pennsylvania}. \\ \hline
Ground Truth & \textcolor{blue}{Pittsburgh Pennsylvania}. \\ \hline
\end{tabularx}
\caption{Case for ToG. The \textcolor{red}{wrong answers} are highlighted with red color, and the \textcolor{blue}{correct answers} are highlighted with blue color.}
\label{table: case2}
\end{table}

\begin{table}[htbp]
\renewcommand\arraystretch{1.5}
\centering
\begin{tabularx}{\textwidth}{p{2cm}|p{11cm}}
\hline
Question & What state is the college that Sampson Salter Blowers is a grad student of located? \\ \hline
\multirow{1}{*}{\begin{tabularx}{\linewidth}{@{}l@{}}Reasoning \\ Paths \\ \end{tabularx}} & Sampson Salter Blowers $\to$ \textit{education.education.student} $\to$ \textit{education.education.institution} $\to$ \{Harvard College\}. (\textbf{Path 1}, Score: 0.75) \\
& Sampson Salter Blowers $\to$ \textit{education.education.student} $\to$ \textit{education.educational\_institution.students\_graduates} $\to$ \{\}. (\textbf{Path 2}, Score: 0.2) \\
& Sampson Salter Blowers $\to$ \textit{education.education.student} $\to$ \textit{people.person.education} $\to$ \{\}. (\textbf{Path 3}, Score: 0.05) \\ \hline
\multirow{1}{*}{\begin{tabularx}{\linewidth}{@{}l@{}} Outputs \\ \end{tabularx}} & \textbf{IO}: Unknown. \\
& \textbf{CoT}: I'm sorry, but I do not have enough information to answer your question. Can you please provide more context or details about Sampson Salter Blowers and the college he attended? \\
& \textbf{New Bing}: I'm sorry but I couldn't find any information about Sampson Salter Blowers being a grad student. However, he was a noted North American lawyer, Lovalist and jurist from Nova Scotia who graduated from Harvard College with a ba in 1763 and an ma in 1765. He was admitted as an attorney in the Suffolk lnferior Court in 1766. \\ 
& \textbf{ToG}: Based on the given relations chains, Sampson Salter Blowers is a grad student of Harvard College. Therefore, the answer to the question is Harvard College, which is located in \textcolor{blue}{Massachusetts}. \\ \hline
Ground Truth & \textcolor{blue}{Massachusetts}. \\ \hline
\end{tabularx}
\caption{Case for ToG-R. "Unknown" denotes LLM is unable to generate a valid response. The \textcolor{red}{wrong answers} are highlighted with red color, and the \textcolor{blue}{correct answers} are highlighted with blue color.}
\label{table: case3}
\end{table}

\begin{table}[htbp]
\renewcommand\arraystretch{1.5}
\centering
\begin{tabularx}{\textwidth}{p{2cm}|p{11cm}}
\hline
Question & What is mascot Phillie Phanatic's team's spring training stadium? \\ \hline
\multirow{1}{*}{\begin{tabularx}{\linewidth}{@{}l@{}}Reasoning \\ Paths \\ \end{tabularx}} & Phillie Phanatic $\to$ \textit{sports.sports\_team.team\_mascot} $\to$ Philadelphia Phillies $\to$ \textit{sports.sports\_team.arena\_stadium} $\to$ Bright House Field. (\textbf{Path 1}, Score: 0.8) \\

& Phillie Phanatic $\to$ \textit{sports.sports\_team.team\_mascot} $\to$ Philadelphia Phillies $\to$ \textit{ports.sports\_team\_location.teams} $\to$ Philadelphia. (\textbf{Path 2}, Score: 0.15) \\

& Phillie Phanatic $\to$ \textit{sports.sports\_team.team\_mascot} $\to$ Philadelphia Phillies $\to$ \textit{sports.sports\_team.team} $\to$ Phillie Phanatic. (\textbf{Path 3}, Score: 0.05) \\
 \hline
\multirow{1}{*}{\begin{tabularx}{\linewidth}{@{}l@{}} Outputs \\ \end{tabularx}} & \textbf{IO}: \textcolor{blue}{Spectrum Field}. \\
& \textbf{CoT}: First, the mascot Phillie Phanatic represents the Philadelphia Phillies baseball team. Second, the Philadelphia Phillies' spring training stadium is located in Clearwater, Florida and is known as Spectrum Field. The answer is \textcolor{blue}{Spectrum Field}. \\
& \textbf{New Bing}: The Philadelphia Phillies' spring training stadium is called \textcolor{red}{BayCare Ballpark}. \\ 
& \textbf{ToG}: The spring training stadium for Phillie Phanatic's team, the Philadelphia Phillies, is \textcolor{red}{Bright House Field}. \\ \hline
Ground Truth & \textcolor{red}{Bright House Field} (KB), \textcolor{blue}{Spectrum Field} (Actual). \\ \hline
\end{tabularx}
\caption{Case for ToG. The \textcolor{red}{wrong answers} are highlighted with red color, and the \textcolor{blue}{correct answers} are highlighted with blue color.}
\label{table: case4}
\end{table}

\newpage

\section{SPARQL and Prompts}
In this section, we show all the prompts that need to be used in the main experiments. First, we pre-define SPARQL for Freebase queries, which can be executed by simply filling in the appropriate mid and relation. For Wikidata, we abstain from employing executable SPARQL, rather we directly engage in querying through nine pre-defined service APIs.
\subsection{Pre-defined SPARQL}~\label{sparqls}
\subsubsection{Relation Search}
\lstset{language=SPARQL}
\begin{lstlisting}
PREFIX ns: <http://rdf.freebase.com/ns/>
SELECT ?relation
WHERE {
    ns:|\textcolor{purple}{mid}| ?relation ?x . 
}
\end{lstlisting}

\lstset{language=SPARQL}
\begin{lstlisting}
PREFIX ns: <http://rdf.freebase.com/ns/>
SELECT ?relation
WHERE {
    ?x ?relation ns:|\textcolor{purple}{mid}| .
}
\end{lstlisting}

\subsubsection{Entity Search}
\lstset{language=SPARQL}
\begin{lstlisting}
PREFIX ns: <http://rdf.freebase.com/ns/>
SELECT ?tailEntity
WHERE {
    ns:|\textcolor{purple}{mid}| ns:|\textcolor{cyan}{relation}| ?tailEntity . 
}

\end{lstlisting}

\lstset{language=SPARQL}
\begin{lstlisting}
PREFIX ns: <http://rdf.freebase.com/ns/>
SELECT ?tailEntity
WHERE {
    ?tailEntity ns:|\textcolor{purple}{mid}| ns:|\textcolor{cyan}{relation}|  . 
}
\end{lstlisting}

\subsubsection{Convert Mid to Label}

\lstset{language=SPARQL}
\begin{lstlisting}
PREFIX ns: <http://rdf.freebase.com/ns/>
SELECT DISTINCT ?tailEntity
WHERE {
{
    ?entity ns:type.object.name ?tailEntity .
    FILTER(?entity = ns:|\textcolor{purple}{mid}|)
}
UNION
{
    ?entity <http://www.w3.org/2002/07/owlsameAs> ?tailEntity .
    FILTER(?entity = ns:|\textcolor{purple}{mid}|)
}
}
\end{lstlisting}

\subsection{Pre-defined APIs}~\label{APIs}

\lstset{language=Python}
\begin{lstlisting}
def label2qid(self, label: str) -> str:

def label2pid(self, label: str) -> str:

def pid2label(self, pid: str) -> str:

def qid2label(self, qid: str) -> str:

def get_all_relations_of_an_entity(self, entity_qid: str) 
    -> tp.Dict[str, tp.List]:

def get_tail_entities_given_head_and_relation(self, head_qid: str, relation_pid: str) 
    -> tp.Dict[str, tp.List]:

def get_tail_values_given_head_and_relation(self, head_qid: str, relation_pid: str) -> tp.List[str]:

def get_external_id_given_head_and_relation(self, head_qid: str, relation_pid: str) -> tp.List[str]:

def mid2qid(self, mid: str) -> str:
\end{lstlisting}

\subsection{ToG}
\subsubsection{Relation Prune}~\label{prompts4rel}
Please retrieve $k$ relations (separated by semicolon) that contribute to the question and rate their contribution on a scale from 0 to 1 (the sum of the scores of $k$ relations is 1).

\texttt{In-Context Few-shot}

Q: \{Query\} 

Topic Entity: \{Topic Entity\} 

Relations: \{list of relations\} 

A: 

\subsubsection{Entity Prune}~\label{prompts4ent}
Please score the entities' contribution to the question on a scale from 0 to 1 (the sum of the scores of all entities is 1).

\texttt{In-Context Few-shot}

Q: \{Query\} 

Relation: \{Current Relation\} 

Entites:  \{list of entities\} 

Score:

\subsubsection{Reasoning}~\label{prompts4rea}
Given a question and the associated retrieved knowledge graph triples (entity, relation, entity), you are asked to answer whether it's sufficient for you to answer the question with these triples and your knowledge (Yes or No).

\texttt{In-Context Few-shot}

Q: \{Query\} 

Knowledge triples: \{Explored Paths\} 

A: 

\subsubsection{Generate}~\label{prompts4gen}
Given a question and the associated retrieved knowledge graph triples (entity, relation, entity), you are asked to answer the question with these triples and your own knowledge.

\texttt{In-Context Few-shot}

Q: \{Query\} 

Knowledge triples: \{Explored Paths\} 

A: 

\subsection{ToG-R}

\subsubsection{Reasoning}~\label{prompts4togr}
Please answer the question using Topic Entity,  Relations Chains and their Candidate Entities that contribute to the question, you are asked to answer whether it's sufficient for you to answer the question with these triples and your knowledge (Yes or No).

\texttt{In-Context Few-shot}

Q: \{Query\} 

Topic Entity, with relations chains, and their candidate entities: \{Explored Relation Chains\} 

A: 

\subsection{CoT and IO}
\subsubsection{CoT prompt}
Q: What state is home to the university that is represented in sports by George Washington Colonials men's basketball?

A: First, the education institution has a sports team named George Washington Colonials men's basketball in is George Washington University , Second, George Washington University is in Washington D.C. The answer is Washington, D.C.

Q: Who lists Pramatha Chaudhuri as an influence and wrote Jana Gana Mana?

A: First, Bharoto Bhagyo Bidhata wrote Jana Gana Mana. Second, Bharoto Bhagyo Bidhata lists Pramatha Chaudhuri as an influence. The answer is Bharoto Bhagyo Bidhata.

Q: Who was the artist nominated for an award for You Drive Me Crazy?

A: First, the artist nominated for an award for You Drive Me Crazy is Britney Spears. The answer is Jason Allen Alexander.

Q: What person born in Siegen influenced the work of Vincent Van Gogh?

A: First, Peter Paul Rubens, Claude Monet and etc. influenced the work of Vincent Van Gogh. Second, Peter Paul Rubens born in Siegen. The answer is Peter Paul Rubens.

Q: What is the country close to Russia where Mikheil Saakashvii holds a government position?

A: First, China, Norway, Finland, Estonia and Georgia is close to Russia. Second, Mikheil Saakashvii holds a government position at Georgia. The answer is Georgia.

Q: What drug did the actor who portrayed the character Urethane Wheels Guy overdosed on?

A: First, Mitchell Lee Hedberg portrayed character Urethane Wheels Guy. Second, Mitchell Lee Hedberg overdose Heroin. The answer is Heroin.

Q: \{Query\} 

A: 
\subsubsection{IO prompt}
Q: What state is home to the university that is represented in sports by George Washington Colonials men's basketball?

A: Washington, D.C.

Q: Who lists Pramatha Chaudhuri as an influence and wrote Jana Gana Mana?

A: Bharoto Bhagyo Bidhata.

Q: Who was the artist nominated for an award for You Drive Me Crazy?

A: Jason Allen Alexander.

Q: What person born in Siegen influenced the work of Vincent Van Gogh?

A: Peter Paul Rubens.

Q: What is the country close to Russia where Mikheil Saakashvii holds a government position?

A: Georgia.

Q: What drug did the actor who portrayed the character Urethane Wheels Guy overdosed on?

A: Heroin.

Q: \{Query\} 

A: 
\end{document}

%% file: math_commands.tex
\usepackage{amsmath,amsfonts,bm}

\def\eqref#1{equation~\ref{#1}}

\def\1{\bm{1}}

\DeclareMathAlphabet{\mathsfit}{\encodingdefault}{\sfdefault}{m}{sl}
\SetMathAlphabet{\mathsfit}{bold}{\encodingdefault}{\sfdefault}{bx}{n}













%% file: iclr2024_conference.bbl
\begin{thebibliography}{73}
\providecommand{\natexlab}[1]{#1}
\providecommand{\url}[1]{\texttt{#1}}
\expandafter\ifx\csname urlstyle\endcsname\relax
  \providecommand{\doi}[1]{doi: #1}\else
  \providecommand{\doi}{doi: \begingroup \urlstyle{rm}\Url}\fi

\bibitem[Atif et~al.(2023)Atif, Khatib, and Difallah]{kg_beam}
Farah Atif, Ola~El Khatib, and Djellel~Eddine Difallah.
\newblock Beamqa: Multi-hop knowledge graph question answering with sequence-to-sequence prediction and beam search.
\newblock In Hsin{-}Hsi Chen, Wei{-}Jou~(Edward) Duh, Hen{-}Hsen Huang, Makoto~P. Kato, Josiane Mothe, and Barbara Poblete (eds.), \emph{Proceedings of the 46th International {ACM} {SIGIR} Conference on Research and Development in Information Retrieval, {SIGIR} 2023, Taipei, Taiwan, July 23-27, 2023}, pp.\  781--790. {ACM}, 2023.
\newblock \doi{10.1145/3539618.3591698}.
\newblock URL \url{https://doi.org/10.1145/3539618.3591698}.

\bibitem[Baek et~al.(2023{\natexlab{a}})Baek, Aji, Lehmann, and Hwang]{simple_questions1}
Jinheon Baek, Alham~Fikri Aji, Jens Lehmann, and Sung~Ju Hwang.
\newblock Direct fact retrieval from knowledge graphs without entity linking.
\newblock In Anna Rogers, Jordan~L. Boyd{-}Graber, and Naoaki Okazaki (eds.), \emph{Proceedings of the 61st Annual Meeting of the Association for Computational Linguistics (Volume 1: Long Papers), {ACL} 2023, Toronto, Canada, July 9-14, 2023}, pp.\  10038--10055. Association for Computational Linguistics, 2023{\natexlab{a}}.
\newblock \doi{10.18653/v1/2023.acl-long.558}.
\newblock URL \url{https://doi.org/10.18653/v1/2023.acl-long.558}.

\bibitem[Baek et~al.(2023{\natexlab{b}})Baek, Aji, and Saffari]{knowledgeaugmented}
Jinheon Baek, Alham~Fikri Aji, and Amir Saffari.
\newblock Knowledge-augmented language model prompting for zero-shot knowledge graph question answering, 2023{\natexlab{b}}.

\bibitem[Banerjee et~al.(2023)Banerjee, Nair, Usbeck, and Biemann]{banerjee2023gettqa}
Debayan Banerjee, Pranav~Ajit Nair, Ricardo Usbeck, and Chris Biemann.
\newblock Gett-qa: Graph embedding based t2t transformer for knowledge graph question answering, 2023.

\bibitem[Berant et~al.(2013)Berant, Chou, Frostig, and Liang]{webq_dataset}
Jonathan Berant, Andrew Chou, Roy Frostig, and Percy Liang.
\newblock Semantic parsing on freebase from question-answer pairs.
\newblock In \emph{Proceedings of the 2013 Conference on Empirical Methods in Natural Language Processing, {EMNLP} 2013, 18-21 October 2013, Grand Hyatt Seattle, Seattle, Washington, USA, {A} meeting of SIGDAT, a Special Interest Group of the {ACL}}, pp.\  1533--1544. {ACL}, 2013.
\newblock URL \url{https://aclanthology.org/D13-1160/}.

\bibitem[Besta et~al.(2023)Besta, Blach, Kubicek, Gerstenberger, Gianinazzi, Gajda, Lehmann, Podstawski, Niewiadomski, Nyczyk, and Hoefler]{got}
Maciej Besta, Nils Blach, Ales Kubicek, Robert Gerstenberger, Lukas Gianinazzi, Joanna Gajda, Tomasz Lehmann, Michal Podstawski, Hubert Niewiadomski, Piotr Nyczyk, and Torsten Hoefler.
\newblock Graph of thoughts: Solving elaborate problems with large language models, 2023.

\bibitem[Bollacker et~al.(2008)Bollacker, Evans, Paritosh, Sturge, and Taylor]{freebase}
Kurt~D. Bollacker, Colin Evans, Praveen~K. Paritosh, Tim Sturge, and Jamie Taylor.
\newblock Freebase: a collaboratively created graph database for structuring human knowledge.
\newblock In \emph{SIGMOD Conference}, 2008.

\bibitem[Bordes et~al.(2015)Bordes, Usunier, Chopra, and Weston]{simple_questions_dataset}
Antoine Bordes, Nicolas Usunier, Sumit Chopra, and Jason Weston.
\newblock Large-scale simple question answering with memory networks.
\newblock \emph{CoRR}, abs/1506.02075, 2015.
\newblock URL \url{http://arxiv.org/abs/1506.02075}.

\bibitem[Brown et~al.(2020{\natexlab{a}})Brown, Mann, Ryder, Subbiah, Kaplan, Dhariwal, Neelakantan, Shyam, Sastry, Askell, Agarwal, Herbert{-}Voss, Krueger, Henighan, Child, Ramesh, Ziegler, Wu, Winter, Hesse, Chen, Sigler, Litwin, Gray, Chess, Clark, Berner, McCandlish, Radford, Sutskever, and Amodei]{fewshotlearner}
Tom~B. Brown, Benjamin Mann, Nick Ryder, Melanie Subbiah, Jared Kaplan, Prafulla Dhariwal, Arvind Neelakantan, Pranav Shyam, Girish Sastry, Amanda Askell, Sandhini Agarwal, Ariel Herbert{-}Voss, Gretchen Krueger, Tom Henighan, Rewon Child, Aditya Ramesh, Daniel~M. Ziegler, Jeffrey Wu, Clemens Winter, Christopher Hesse, Mark Chen, Eric Sigler, Mateusz Litwin, Scott Gray, Benjamin Chess, Jack Clark, Christopher Berner, Sam McCandlish, Alec Radford, Ilya Sutskever, and Dario Amodei.
\newblock Language models are few-shot learners.
\newblock In Hugo Larochelle, Marc'Aurelio Ranzato, Raia Hadsell, Maria{-}Florina Balcan, and Hsuan{-}Tien Lin (eds.), \emph{Advances in Neural Information Processing Systems 33: Annual Conference on Neural Information Processing Systems 2020, NeurIPS 2020, December 6-12, 2020, virtual}, 2020{\natexlab{a}}.
\newblock URL \url{https://proceedings.neurips.cc/paper/2020/hash/1457c0d6bfcb4967418bfb8ac142f64a-Abstract.html}.

\bibitem[Brown et~al.(2020{\natexlab{b}})Brown, Mann, Ryder, Subbiah, Kaplan, Dhariwal, Neelakantan, Shyam, Sastry, Askell, Agarwal, Herbert{-}Voss, Krueger, Henighan, Child, Ramesh, Ziegler, Wu, Winter, Hesse, Chen, Sigler, Litwin, Gray, Chess, Clark, Berner, McCandlish, Radford, Sutskever, and Amodei]{io_prompt}
Tom~B. Brown, Benjamin Mann, Nick Ryder, Melanie Subbiah, Jared Kaplan, Prafulla Dhariwal, Arvind Neelakantan, Pranav Shyam, Girish Sastry, Amanda Askell, Sandhini Agarwal, Ariel Herbert{-}Voss, Gretchen Krueger, Tom Henighan, Rewon Child, Aditya Ramesh, Daniel~M. Ziegler, Jeffrey Wu, Clemens Winter, Christopher Hesse, Mark Chen, Eric Sigler, Mateusz Litwin, Scott Gray, Benjamin Chess, Jack Clark, Christopher Berner, Sam McCandlish, Alec Radford, Ilya Sutskever, and Dario Amodei.
\newblock Language models are few-shot learners.
\newblock In Hugo Larochelle, Marc'Aurelio Ranzato, Raia Hadsell, Maria{-}Florina Balcan, and Hsuan{-}Tien Lin (eds.), \emph{Advances in Neural Information Processing Systems 33: Annual Conference on Neural Information Processing Systems 2020, NeurIPS 2020, December 6-12, 2020, virtual}, 2020{\natexlab{b}}.
\newblock URL \url{https://proceedings.neurips.cc/paper/2020/hash/1457c0d6bfcb4967418bfb8ac142f64a-Abstract.html}.

\bibitem[Cao et~al.(2022)Cao, Shi, Yao, Lv, Yu, Hou, Li, Liu, and Xiao]{program_transfer}
Shulin Cao, Jiaxin Shi, Zijun Yao, Xin Lv, Jifan Yu, Lei Hou, Juanzi Li, Zhiyuan Liu, and Jinghui Xiao.
\newblock Program transfer for answering complex questions over knowledge bases.
\newblock In Smaranda Muresan, Preslav Nakov, and Aline Villavicencio (eds.), \emph{Proceedings of the 60th Annual Meeting of the Association for Computational Linguistics (Volume 1: Long Papers), {ACL} 2022, Dublin, Ireland, May 22-27, 2022}, pp.\  8128--8140. Association for Computational Linguistics, 2022.
\newblock \doi{10.18653/v1/2022.acl-long.559}.
\newblock URL \url{https://doi.org/10.18653/v1/2022.acl-long.559}.

\bibitem[Chowdhery et~al.(2022)Chowdhery, Narang, Devlin, Bosma, Mishra, Roberts, Barham, Chung, Sutton, Gehrmann, Schuh, Shi, Tsvyashchenko, Maynez, Rao, Barnes, Tay, Shazeer, Prabhakaran, Reif, Du, Hutchinson, Pope, Bradbury, Austin, Isard, Gur-Ari, Yin, Duke, Levskaya, Ghemawat, Dev, Michalewski, Garcia, Misra, Robinson, Fedus, Zhou, Ippolito, Luan, Lim, Zoph, Spiridonov, Sepassi, Dohan, Agrawal, Omernick, Dai, Pillai, Pellat, Lewkowycz, Moreira, Child, Polozov, Lee, Zhou, Wang, Saeta, Diaz, Firat, Catasta, Wei, Meier-Hellstern, Eck, Dean, Petrov, and Fiedel]{palm}
Aakanksha Chowdhery, Sharan Narang, Jacob Devlin, Maarten Bosma, Gaurav Mishra, Adam Roberts, Paul Barham, Hyung~Won Chung, Charles Sutton, Sebastian Gehrmann, Parker Schuh, Kensen Shi, Sasha Tsvyashchenko, Joshua Maynez, Abhishek Rao, Parker Barnes, Yi~Tay, Noam Shazeer, Vinodkumar Prabhakaran, Emily Reif, Nan Du, Ben Hutchinson, Reiner Pope, James Bradbury, Jacob Austin, Michael Isard, Guy Gur-Ari, Pengcheng Yin, Toju Duke, Anselm Levskaya, Sanjay Ghemawat, Sunipa Dev, Henryk Michalewski, Xavier Garcia, Vedant Misra, Kevin Robinson, Liam Fedus, Denny Zhou, Daphne Ippolito, David Luan, Hyeontaek Lim, Barret Zoph, Alexander Spiridonov, Ryan Sepassi, David Dohan, Shivani Agrawal, Mark Omernick, Andrew~M. Dai, Thanumalayan~Sankaranarayana Pillai, Marie Pellat, Aitor Lewkowycz, Erica Moreira, Rewon Child, Oleksandr Polozov, Katherine Lee, Zongwei Zhou, Xuezhi Wang, Brennan Saeta, Mark Diaz, Orhan Firat, Michele Catasta, Jason Wei, Kathy Meier-Hellstern, Douglas Eck, Jeff Dean, Slav Petrov, and Noah Fiedel.
\newblock Palm: Scaling language modeling with pathways, 2022.

\bibitem[Das et~al.(2021)Das, Zaheer, Thai, Godbole, Perez, Lee, Tan, Polymenakos, and McCallum]{cwq1}
Rajarshi Das, Manzil Zaheer, Dung Thai, Ameya Godbole, Ethan Perez, Jay-Yoon Lee, Lizhen Tan, Lazaros Polymenakos, and Andrew McCallum.
\newblock Case-based reasoning for natural language queries over knowledge bases, 2021.

\bibitem[de~Jong et~al.(2022)de~Jong, Zemlyanskiy, Ainslie, FitzGerald, Sanghai, Sha, and Cohen]{de2022fido}
Michiel de~Jong, Yury Zemlyanskiy, Joshua Ainslie, Nicholas FitzGerald, Sumit Sanghai, Fei Sha, and William Cohen.
\newblock Fido: Fusion-in-decoder optimized for stronger performance and faster inference.
\newblock \emph{arXiv preprint arXiv:2212.08153}, 2022.

\bibitem[dos Santos et~al.(2022)dos Santos, Dong, Cer, Nham, Shakeri, Ni, and hsuan Sung]{santos2022knowledge}
Cicero~Nogueira dos Santos, Zhe Dong, Daniel Cer, John Nham, Siamak Shakeri, Jianmo Ni, and Yun hsuan Sung.
\newblock Knowledge prompts: Injecting world knowledge into language models through soft prompts, 2022.

\bibitem[ElSahar et~al.(2018)ElSahar, Vougiouklis, Remaci, Gravier, Hare, Laforest, and Simperl]{trex_dataset}
Hady ElSahar, Pavlos Vougiouklis, Arslen Remaci, Christophe Gravier, Jonathon~S. Hare, Fr{\'{e}}d{\'{e}}rique Laforest, and Elena Simperl.
\newblock T-rex: {A} large scale alignment of natural language with knowledge base triples.
\newblock In Nicoletta Calzolari, Khalid Choukri, Christopher Cieri, Thierry Declerck, Sara Goggi, K{\^{o}}iti Hasida, Hitoshi Isahara, Bente Maegaard, Joseph Mariani, H{\'{e}}l{\`{e}}ne Mazo, Asunci{\'{o}}n Moreno, Jan Odijk, Stelios Piperidis, and Takenobu Tokunaga (eds.), \emph{Proceedings of the Eleventh International Conference on Language Resources and Evaluation, {LREC} 2018, Miyazaki, Japan, May 7-12, 2018}. European Language Resources Association {(ELRA)}, 2018.
\newblock URL \url{http://www.lrec-conf.org/proceedings/lrec2018/summaries/632.html}.

\bibitem[Fu et~al.(2023)Fu, Peng, Sabharwal, Clark, and Khot]{complex_cot}
Yao Fu, Hao Peng, Ashish Sabharwal, Peter Clark, and Tushar Khot.
\newblock Complexity-based prompting for multi-step reasoning.
\newblock In \emph{The Eleventh International Conference on Learning Representations, {ICLR} 2023, Kigali, Rwanda, May 1-5, 2023}. OpenReview.net, 2023.
\newblock URL \url{https://openreview.net/pdf?id=yf1icZHC-l9}.

\bibitem[Glass et~al.(2022)Glass, Rossiello, Chowdhury, Naik, Cai, and Gliozzo]{trex1}
Michael Glass, Gaetano Rossiello, Md~Faisal~Mahbub Chowdhury, Ankita Naik, Pengshan Cai, and Alfio Gliozzo.
\newblock {R}e2{G}: Retrieve, rerank, generate.
\newblock In \emph{Proceedings of the 2022 Conference of the North American Chapter of the Association for Computational Linguistics: Human Language Technologies}, pp.\  2701--2715, Seattle, United States, July 2022. Association for Computational Linguistics.
\newblock \doi{10.18653/v1/2022.naacl-main.194}.
\newblock URL \url{https://aclanthology.org/2022.naacl-main.194}.

\bibitem[Gu et~al.(2021)Gu, Kase, Vanni, Sadler, Liang, Yan, and Su]{garilqa_dataset}
Yu~Gu, Sue Kase, Michelle Vanni, Brian~M. Sadler, Percy Liang, Xifeng Yan, and Yu~Su.
\newblock Beyond {I.I.D.:} three levels of generalization for question answering on knowledge bases.
\newblock In Jure Leskovec, Marko Grobelnik, Marc Najork, Jie Tang, and Leila Zia (eds.), \emph{{WWW} '21: The Web Conference 2021, Virtual Event / Ljubljana, Slovenia, April 19-23, 2021}, pp.\  3477--3488. {ACM} / {IW3C2}, 2021.
\newblock \doi{10.1145/3442381.3449992}.
\newblock URL \url{https://doi.org/10.1145/3442381.3449992}.

\bibitem[Gu et~al.(2023)Gu, Deng, and Su]{gu2023dont}
Yu~Gu, Xiang Deng, and Yu~Su.
\newblock Don't generate, discriminate: A proposal for grounding language models to real-world environments, 2023.

\bibitem[He et~al.(2021)He, Lan, Jiang, Zhao, and Wen]{NSM}
Gaole He, Yunshi Lan, Jing Jiang, Wayne~Xin Zhao, and Ji{-}Rong Wen.
\newblock Improving multi-hop knowledge base question answering by learning intermediate supervision signals.
\newblock In Liane Lewin{-}Eytan, David Carmel, Elad Yom{-}Tov, Eugene Agichtein, and Evgeniy Gabrilovich (eds.), \emph{{WSDM} '21, The Fourteenth {ACM} International Conference on Web Search and Data Mining, Virtual Event, Israel, March 8-12, 2021}, pp.\  553--561. {ACM}, 2021.
\newblock \doi{10.1145/3437963.3441753}.
\newblock URL \url{https://doi.org/10.1145/3437963.3441753}.

\bibitem[Hu et~al.(2023)Hu, Liu, Zhao, Hou, Nie, and Li]{KELLMsurvey}
Linmei Hu, Zeyi Liu, Ziwang Zhao, Lei Hou, Liqiang Nie, and Juanzi Li.
\newblock A survey of knowledge enhanced pre-trained language models.
\newblock \emph{IEEE Transactions on Knowledge and Data Engineering}, 2023.

\bibitem[Huang et~al.(2024)Huang, Wei, Qu, Xie, Mao, and Chen]{huang2024joint}
Rikui Huang, Wei Wei, Xiaoye Qu, Wenfeng Xie, Xianling Mao, and Dangyang Chen.
\newblock Joint multi-facts reasoning network for complex temporal question answering over knowledge graph.
\newblock \emph{arXiv preprint arXiv:2401.02212}, 2024.

\bibitem[Jiang et~al.(2023)Jiang, Zhou, Dong, Ye, Zhao, and Wen]{structgpt}
Jinhao Jiang, Kun Zhou, Zican Dong, Keming Ye, Wayne~Xin Zhao, and Ji-Rong Wen.
\newblock Structgpt: A general framework for large language model to reason over structured data, 2023.

\bibitem[Jurafsky \& Martin(2009)Jurafsky and Martin]{beamsearch}
Dan Jurafsky and James~H. Martin.
\newblock \emph{Speech and language processing: an introduction to natural language processing, computational linguistics, and speech recognition, 2nd Edition}.
\newblock Prentice Hall series in artificial intelligence. Prentice Hall, Pearson Education International, 2009.
\newblock ISBN 9780135041963.
\newblock URL \url{https://www.worldcat.org/oclc/315913020}.

\bibitem[Kedia et~al.(2022)Kedia, Zaidi, and Lee]{webq1}
Akhil Kedia, Mohd~Abbas Zaidi, and Haejun Lee.
\newblock Fie: Building a global probability space by leveraging early fusion in encoder for open-domain question answering.
\newblock In Yoav Goldberg, Zornitsa Kozareva, and Yue Zhang (eds.), \emph{Proceedings of the 2022 Conference on Empirical Methods in Natural Language Processing, {EMNLP} 2022, Abu Dhabi, United Arab Emirates, December 7-11, 2022}, pp.\  4246--4260. Association for Computational Linguistics, 2022.
\newblock \doi{10.18653/v1/2022.emnlp-main.285}.
\newblock URL \url{https://doi.org/10.18653/v1/2022.emnlp-main.285}.

\bibitem[Kojima et~al.(2022)Kojima, Gu, Reid, Matsuo, and Iwasawa]{zero_shot_cot}
Takeshi Kojima, Shixiang~Shane Gu, Machel Reid, Yutaka Matsuo, and Yusuke Iwasawa.
\newblock Large language models are zero-shot reasoners.
\newblock In \emph{NeurIPS}, 2022.
\newblock URL \url{http://papers.nips.cc/paper\_files/paper/2022/hash/8bb0d291acd4acf06ef112099c16f326-Abstract-Conference.html}.

\bibitem[Lan \& Jiang(2020)Lan and Jiang]{lan2020query}
Yunshi Lan and Jing Jiang.
\newblock Query graph generation for answering multi-hop complex questions from knowledge bases.
\newblock Association for Computational Linguistics, 2020.

\bibitem[Lan et~al.(2022)Lan, He, Jiang, Jiang, Zhao, and Wen]{complexKBQA}
Yunshi Lan, Gaole He, Jinhao Jiang, Jing Jiang, Wayne~Xin Zhao, and Ji-Rong Wen.
\newblock Complex knowledge base question answering: A survey.
\newblock \emph{IEEE Transactions on Knowledge and Data Engineering}, 2022.

\bibitem[Lewis et~al.(2021)Lewis, Perez, Piktus, Petroni, Karpukhin, Goyal, Küttler, Lewis, tau Yih, Rocktäschel, Riedel, and Kiela]{lewis2021retrievalaugmented}
Patrick Lewis, Ethan Perez, Aleksandra Piktus, Fabio Petroni, Vladimir Karpukhin, Naman Goyal, Heinrich Küttler, Mike Lewis, Wen tau Yih, Tim Rocktäschel, Sebastian Riedel, and Douwe Kiela.
\newblock Retrieval-augmented generation for knowledge-intensive nlp tasks, 2021.

\bibitem[Li et~al.(2023{\natexlab{a}})Li, Ma, Zhuang, Gu, Su, and Chen]{DB-BLINDER}
Tianle Li, Xueguang Ma, Alex Zhuang, Yu~Gu, Yu~Su, and Wenhu Chen.
\newblock Few-shot in-context learning on knowledge base question answering.
\newblock In Anna Rogers, Jordan~L. Boyd{-}Graber, and Naoaki Okazaki (eds.), \emph{Proceedings of the 61st Annual Meeting of the Association for Computational Linguistics (Volume 1: Long Papers), {ACL} 2023, Toronto, Canada, July 9-14, 2023}, pp.\  6966--6980. Association for Computational Linguistics, 2023{\natexlab{a}}.
\newblock \doi{10.18653/v1/2023.acl-long.385}.
\newblock URL \url{https://doi.org/10.18653/v1/2023.acl-long.385}.

\bibitem[Li et~al.(2023{\natexlab{b}})Li, Wei, Qu, Mao, Yuan, Xie, and Chen]{li2023trea}
Wendi Li, Wei Wei, Xiaoye Qu, Xian-Ling Mao, Ye~Yuan, Wenfeng Xie, and Dangyang Chen.
\newblock Trea: Tree-structure reasoning schema for conversational recommendation.
\newblock In \emph{Proceedings of the 61st Annual Meeting of the Association for Computational Linguistics (Volume 1: Long Papers)}, pp.\  2970--2982, 2023{\natexlab{b}}.

\bibitem[Li et~al.(2023{\natexlab{c}})Li, Zhao, Chia, Ding, Bing, Joty, and Poria]{cok}
Xingxuan Li, Ruochen Zhao, Yew~Ken Chia, Bosheng Ding, Lidong Bing, Shafiq Joty, and Soujanya Poria.
\newblock Chain of knowledge: A framework for grounding large language models with structured knowledge bases, 2023{\natexlab{c}}.

\bibitem[Liu et~al.(2020)Liu, Qu, Dong, and Zhou]{liu2020reasoning}
Daizong Liu, Xiaoye Qu, Jianfeng Dong, and Pan Zhou.
\newblock Reasoning step-by-step: Temporal sentence localization in videos via deep rectification-modulation network.
\newblock In \emph{Proceedings of the 28th International Conference on Computational Linguistics}, pp.\  1841--1851, 2020.

\bibitem[Liu et~al.(2024)Liu, Zeng, He, Jiang, and He]{liu2024what}
Wei Liu, Weihao Zeng, Keqing He, Yong Jiang, and Junxian He.
\newblock What makes good data for alignment? a comprehensive study of automatic data selection in instruction tuning.
\newblock In \emph{The Twelfth International Conference on Learning Representations}, 2024.
\newblock URL \url{https://openreview.net/forum?id=BTKAeLqLMw}.

\bibitem[Liu et~al.(2022)Liu, Yavuz, Meng, Radev, Xiong, and Zhou]{liu2022uni}
Ye~Liu, Semih Yavuz, Rui Meng, Dragomir Radev, Caiming Xiong, and Yingbo Zhou.
\newblock Uni-parser: Unified semantic parser for question answering on knowledge base and database.
\newblock \emph{arXiv preprint arXiv:2211.05165}, 2022.

\bibitem[Liu et~al.(2019)Liu, Ott, Goyal, Du, Joshi, Chen, Levy, Lewis, Zettlemoyer, and Stoyanov]{liu2019roberta}
Yinhan Liu, Myle Ott, Naman Goyal, Jingfei Du, Mandar Joshi, Danqi Chen, Omer Levy, Mike Lewis, Luke Zettlemoyer, and Veselin Stoyanov.
\newblock Roberta: A robustly optimized bert pretraining approach, 2019.

\bibitem[Luo et~al.(2024)Luo, Li, Haffari, and Pan]{luo2024rog}
Linhao Luo, Yuan-Fang Li, Gholamreza Haffari, and Shirui Pan.
\newblock Reasoning on graphs: Faithful and interpretable large language model reasoning.
\newblock In \emph{International Conference on Learning Representations}, 2024.

\bibitem[Moiseev et~al.(2022)Moiseev, Dong, Alfonseca, and Jaggi]{knowledge_infusion}
Fedor Moiseev, Zhe Dong, Enrique Alfonseca, and Martin Jaggi.
\newblock {SKILL:} structured knowledge infusion for large language models.
\newblock In Marine Carpuat, Marie{-}Catherine de~Marneffe, and Iv{\'{a}}n Vladimir~Meza Ru{\'{\i}}z (eds.), \emph{Proceedings of the 2022 Conference of the North American Chapter of the Association for Computational Linguistics: Human Language Technologies, {NAACL} 2022, Seattle, WA, United States, July 10-15, 2022}, pp.\  1581--1588. Association for Computational Linguistics, 2022.
\newblock \doi{10.18653/v1/2022.naacl-main.113}.
\newblock URL \url{https://doi.org/10.18653/v1/2022.naacl-main.113}.

\bibitem[Onoe et~al.(2021)Onoe, Zhang, Choi, and Durrett]{creak_dataset}
Yasumasa Onoe, Michael J.~Q. Zhang, Eunsol Choi, and Greg Durrett.
\newblock {CREAK:} {A} dataset for commonsense reasoning over entity knowledge.
\newblock In Joaquin Vanschoren and Sai{-}Kit Yeung (eds.), \emph{Proceedings of the Neural Information Processing Systems Track on Datasets and Benchmarks 1, NeurIPS Datasets and Benchmarks 2021, December 2021, virtual}, 2021.
\newblock URL \url{https://datasets-benchmarks-proceedings.neurips.cc/paper/2021/hash/5737c6ec2e0716f3d8a7a5c4e0de0d9a-Abstract-round2.html}.

\bibitem[OpenAI(2023)]{gpt4}
OpenAI.
\newblock Gpt-4 technical report, 2023.

\bibitem[Ouyang et~al.(2022)Ouyang, Wu, Jiang, Almeida, Wainwright, Mishkin, Zhang, Agarwal, Slama, Ray, Schulman, Hilton, Kelton, Miller, Simens, Askell, Welinder, Christiano, Leike, and Lowe]{gpt3.5}
Long Ouyang, Jeff Wu, Xu~Jiang, Diogo Almeida, Carroll~L. Wainwright, Pamela Mishkin, Chong Zhang, Sandhini Agarwal, Katarina Slama, Alex Ray, John Schulman, Jacob Hilton, Fraser Kelton, Luke Miller, Maddie Simens, Amanda Askell, Peter Welinder, Paul~F. Christiano, Jan Leike, and Ryan Lowe.
\newblock Training language models to follow instructions with human feedback.
\newblock \emph{arXiv Preprint}, 2022.
\newblock \doi{10.48550/arXiv.2203.02155}.
\newblock URL \url{https://doi.org/10.48550/arXiv.2203.02155}.

\bibitem[Pan et~al.(2023)Pan, Luo, Wang, Chen, Wang, and Wu]{pan2023unifying}
Shirui Pan, Linhao Luo, Yufei Wang, Chen Chen, Jiapu Wang, and Xindong Wu.
\newblock Unifying large language models and knowledge graphs: A roadmap.
\newblock \emph{arXiv preprint arXiv:2306.08302}, 2023.

\bibitem[Perevalov et~al.(2022)Perevalov, Diefenbach, Usbeck, and Both]{qald10-en_dataset}
A.~Perevalov, D.~Diefenbach, R.~Usbeck, and A.~Both.
\newblock Qald-9-plus: A multilingual dataset for question answering over dbpedia and wikidata translated by native speakers.
\newblock In \emph{2022 IEEE 16th International Conference on Semantic Computing (ICSC)}, pp.\  229--234, Los Alamitos, CA, USA, jan 2022. IEEE Computer Society.
\newblock \doi{10.1109/ICSC52841.2022.00045}.
\newblock URL \url{https://doi.ieeecomputersociety.org/10.1109/ICSC52841.2022.00045}.

\bibitem[Peters et~al.(2019)Peters, Neumann, Logan, Schwartz, Joshi, Singh, and Smith]{Knowbert}
Matthew~E. Peters, Mark Neumann, Robert Logan, Roy Schwartz, Vidur Joshi, Sameer Singh, and Noah~A. Smith.
\newblock Knowledge enhanced contextual word representations.
\newblock In \emph{Proceedings of the 2019 Conference on Empirical Methods in Natural Language Processing and the 9th International Joint Conference on Natural Language Processing (EMNLP-IJCNLP)}, pp.\  43--54, Hong Kong, China, November 2019. Association for Computational Linguistics.
\newblock \doi{10.18653/v1/D19-1005}.
\newblock URL \url{https://aclanthology.org/D19-1005}.

\bibitem[Petroni et~al.(2021)Petroni, Piktus, Fan, Lewis, Yazdani, Cao, Thorne, Jernite, Karpukhin, Maillard, Plachouras, Rocktäschel, and Riedel]{kilt}
Fabio Petroni, Aleksandra Piktus, Angela Fan, Patrick Lewis, Majid Yazdani, Nicola~De Cao, James Thorne, Yacine Jernite, Vladimir Karpukhin, Jean Maillard, Vassilis Plachouras, Tim Rocktäschel, and Sebastian Riedel.
\newblock Kilt: a benchmark for knowledge intensive language tasks, 2021.

\bibitem[Raffel et~al.(2020)Raffel, Shazeer, Roberts, Lee, Narang, Matena, Zhou, Li, and Liu]{raffel2020exploring}
Colin Raffel, Noam Shazeer, Adam Roberts, Katherine Lee, Sharan Narang, Michael Matena, Yanqi Zhou, Wei Li, and Peter~J Liu.
\newblock Exploring the limits of transfer learning with a unified text-to-text transformer.
\newblock \emph{The Journal of Machine Learning Research}, 21\penalty0 (1):\penalty0 5485--5551, 2020.

\bibitem[Santana et~al.(2022)Santana, Cuteri, Ricca, and Barbara]{qald10-en1}
Manuel Alejandro~Borroto Santana, Bernardo Cuteri, Francesco Ricca, and Vito Barbara.
\newblock {SPARQL-QA} enters the {QALD} challenge.
\newblock In Xi~Yan, Meriem Beloucif, and Ricardo Usbeck (eds.), \emph{Proceedings of the 7th Natural Language Interfaces for the Web of Data (NLIWoD) co-located with the 19th European Semantic Web Conference {(ESWC} 2022), Hersonissos, Greece, May 29th, 2022}, volume 3196 of \emph{{CEUR} Workshop Proceedings}, pp.\  25--31. CEUR-WS.org, 2022.
\newblock URL \url{https://ceur-ws.org/Vol-3196/paper3.pdf}.

\bibitem[Shu et~al.(2022)Shu, Yu, Li, Karlsson, Ma, Qu, and Lin]{grailqa1}
Yiheng Shu, Zhiwei Yu, Yuhan Li, B{\"o}rje Karlsson, Tingting Ma, Yuzhong Qu, and Chin-Yew Lin.
\newblock {TIARA}: Multi-grained retrieval for robust question answering over large knowledge base.
\newblock In \emph{Proceedings of the 2022 Conference on Empirical Methods in Natural Language Processing}, pp.\  8108--8121, Abu Dhabi, United Arab Emirates, December 2022. Association for Computational Linguistics.
\newblock \doi{10.18653/v1/2022.emnlp-main.555}.
\newblock URL \url{https://aclanthology.org/2022.emnlp-main.555}.

\bibitem[Sun et~al.(2019)Sun, Bedrax-Weiss, and Cohen]{sun2019pullnet}
Haitian Sun, Tania Bedrax-Weiss, and William~W. Cohen.
\newblock Pullnet: Open domain question answering with iterative retrieval on knowledge bases and text, 2019.

\bibitem[Sun et~al.(2023{\natexlab{a}})Sun, Liu, Gong, Dong, Lu, Zhang, Jiang, Yang, Majumder, and Duan]{beamsearchqa}
Hao Sun, Xiao Liu, Yeyun Gong, Anlei Dong, Jingwen Lu, Yan Zhang, Daxin Jiang, Linjun Yang, Rangan Majumder, and Nan Duan.
\newblock Beamsearchqa: Large language models are strong zero-shot {QA} solver.
\newblock \emph{CoRR}, abs/2305.14766, 2023{\natexlab{a}}.
\newblock \doi{10.48550/arXiv.2305.14766}.
\newblock URL \url{https://doi.org/10.48550/arXiv.2305.14766}.

\bibitem[Sun et~al.(2023{\natexlab{b}})Sun, Luo, Gong, Lin, Shen, Guo, and Duan]{iter-cot}
Jiashuo Sun, Yi~Luo, Yeyun Gong, Chen Lin, Yelong Shen, Jian Guo, and Nan Duan.
\newblock Enhancing chain-of-thoughts prompting with iterative bootstrapping in large language models, 2023{\natexlab{b}}.

\bibitem[Talmor \& Berant(2018)Talmor and Berant]{cwq_dataset}
Alon Talmor and Jonathan Berant.
\newblock The web as a knowledge-base for answering complex questions.
\newblock In Marilyn~A. Walker, Heng Ji, and Amanda Stent (eds.), \emph{Proceedings of the 2018 Conference of the North American Chapter of the Association for Computational Linguistics: Human Language Technologies, {NAACL-HLT} 2018, New Orleans, Louisiana, USA, June 1-6, 2018, Volume 1 (Long Papers)}, pp.\  641--651. Association for Computational Linguistics, 2018.
\newblock \doi{10.18653/v1/n18-1059}.
\newblock URL \url{https://doi.org/10.18653/v1/n18-1059}.

\bibitem[Talmor et~al.(2019)Talmor, Herzig, Lourie, and Berant]{commonsenseqa}
Alon Talmor, Jonathan Herzig, Nicholas Lourie, and Jonathan Berant.
\newblock Commonsenseqa: {A} question answering challenge targeting commonsense knowledge.
\newblock In Jill Burstein, Christy Doran, and Thamar Solorio (eds.), \emph{Proceedings of the 2019 Conference of the North American Chapter of the Association for Computational Linguistics: Human Language Technologies, {NAACL-HLT} 2019, Minneapolis, MN, USA, June 2-7, 2019, Volume 1 (Long and Short Papers)}, pp.\  4149--4158, 2019.
\newblock \doi{10.18653/v1/n19-1421}.
\newblock URL \url{https://doi.org/10.18653/v1/n19-1421}.

\bibitem[Tan et~al.(2023)Tan, Min, Li, Li, Hu, Chen, and Qi]{LLMKBQA}
Yiming Tan, Dehai Min, Yu~Li, Wenbo Li, Nan Hu, Yongrui Chen, and Guilin Qi.
\newblock Evaluation of chatgpt as a question answering system for answering complex questions.
\newblock \emph{arXiv preprint arXiv:2303.07992}, 2023.

\bibitem[Thoppilan et~al.(2022)Thoppilan, Freitas, Hall, Shazeer, Kulshreshtha, Cheng, Jin, Bos, Baker, Du, Li, Lee, Zheng, Ghafouri, Menegali, Huang, Krikun, Lepikhin, Qin, Chen, Xu, Chen, Roberts, Bosma, Zhou, Chang, Krivokon, Rusch, Pickett, Meier{-}Hellstern, Morris, Doshi, Santos, Duke, Soraker, Zevenbergen, Prabhakaran, Diaz, Hutchinson, Olson, Molina, Hoffman{-}John, Lee, Aroyo, Rajakumar, Butryna, Lamm, Kuzmina, Fenton, Cohen, Bernstein, Kurzweil, y~Arcas, Cui, Croak, Chi, and Le]{lamda}
Romal Thoppilan, Daniel~De Freitas, Jamie Hall, Noam Shazeer, Apoorv Kulshreshtha, Heng{-}Tze Cheng, Alicia Jin, Taylor Bos, Leslie Baker, Yu~Du, YaGuang Li, Hongrae Lee, Huaixiu~Steven Zheng, Amin Ghafouri, Marcelo Menegali, Yanping Huang, Maxim Krikun, Dmitry Lepikhin, James Qin, Dehao Chen, Yuanzhong Xu, Zhifeng Chen, Adam Roberts, Maarten Bosma, Yanqi Zhou, Chung{-}Ching Chang, Igor Krivokon, Will Rusch, Marc Pickett, Kathleen~S. Meier{-}Hellstern, Meredith~Ringel Morris, Tulsee Doshi, Renelito~Delos Santos, Toju Duke, Johnny Soraker, Ben Zevenbergen, Vinodkumar Prabhakaran, Mark Diaz, Ben Hutchinson, Kristen Olson, Alejandra Molina, Erin Hoffman{-}John, Josh Lee, Lora Aroyo, Ravi Rajakumar, Alena Butryna, Matthew Lamm, Viktoriya Kuzmina, Joe Fenton, Aaron Cohen, Rachel Bernstein, Ray Kurzweil, Blaise~Ag{\"{u}}era y~Arcas, Claire Cui, Marian Croak, Ed~H. Chi, and Quoc Le.
\newblock Lamda: Language models for dialog applications.
\newblock \emph{CoRR}, 2022.
\newblock URL \url{https://arxiv.org/abs/2201.08239}.

\bibitem[Touvron et~al.(2023)Touvron, Martin, Stone, Albert, Almahairi, Babaei, Bashlykov, Batra, Bhargava, Bhosale, Bikel, Blecher, Ferrer, Chen, Cucurull, Esiobu, Fernandes, Fu, Fu, Fuller, Gao, Goswami, Goyal, Hartshorn, Hosseini, Hou, Inan, Kardas, Kerkez, Khabsa, Kloumann, Korenev, Koura, Lachaux, Lavril, Lee, Liskovich, Lu, Mao, Martinet, Mihaylov, Mishra, Molybog, Nie, Poulton, Reizenstein, Rungta, Saladi, Schelten, Silva, Smith, Subramanian, Tan, Tang, Taylor, Williams, Kuan, Xu, Yan, Zarov, Zhang, Fan, Kambadur, Narang, Rodriguez, Stojnic, Edunov, and Scialom]{llama2}
Hugo Touvron, Louis Martin, Kevin Stone, Peter Albert, Amjad Almahairi, Yasmine Babaei, Nikolay Bashlykov, Soumya Batra, Prajjwal Bhargava, Shruti Bhosale, Dan Bikel, Lukas Blecher, Cristian~Canton Ferrer, Moya Chen, Guillem Cucurull, David Esiobu, Jude Fernandes, Jeremy Fu, Wenyin Fu, Brian Fuller, Cynthia Gao, Vedanuj Goswami, Naman Goyal, Anthony Hartshorn, Saghar Hosseini, Rui Hou, Hakan Inan, Marcin Kardas, Viktor Kerkez, Madian Khabsa, Isabel Kloumann, Artem Korenev, Punit~Singh Koura, Marie-Anne Lachaux, Thibaut Lavril, Jenya Lee, Diana Liskovich, Yinghai Lu, Yuning Mao, Xavier Martinet, Todor Mihaylov, Pushkar Mishra, Igor Molybog, Yixin Nie, Andrew Poulton, Jeremy Reizenstein, Rashi Rungta, Kalyan Saladi, Alan Schelten, Ruan Silva, Eric~Michael Smith, Ranjan Subramanian, Xiaoqing~Ellen Tan, Binh Tang, Ross Taylor, Adina Williams, Jian~Xiang Kuan, Puxin Xu, Zheng Yan, Iliyan Zarov, Yuchen Zhang, Angela Fan, Melanie Kambadur, Sharan Narang, Aurelien Rodriguez, Robert Stojnic, Sergey Edunov, and Thomas
  Scialom.
\newblock Llama 2: Open foundation and fine-tuned chat models, 2023.

\bibitem[Vrande\v{c}i\'{c} \& Kr\"{o}tzsch(2014)Vrande\v{c}i\'{c} and Kr\"{o}tzsch]{Wikidata}
Denny Vrande\v{c}i\'{c} and Markus Kr\"{o}tzsch.
\newblock Wikidata: A free collaborative knowledgebase.
\newblock \emph{Commun. ACM}, 57\penalty0 (10):\penalty0 78–85, sep 2014.
\newblock ISSN 0001-0782.
\newblock \doi{10.1145/2629489}.
\newblock URL \url{https://doi.org/10.1145/2629489}.

\bibitem[Wang et~al.(2023{\natexlab{a}})Wang, Sun, Chen, Li, and Gao]{wang2023boosting}
Jianing Wang, Qiushi Sun, Nuo Chen, Xiang Li, and Ming Gao.
\newblock Boosting language models reasoning with chain-of-knowledge prompting, 2023{\natexlab{a}}.

\bibitem[Wang et~al.(2023{\natexlab{b}})Wang, Duan, Wang, Li, Xian, Yin, Rong, and Xiong]{wang2023knowledgedriven}
Keheng Wang, Feiyu Duan, Sirui Wang, Peiguang Li, Yunsen Xian, Chuantao Yin, Wenge Rong, and Zhang Xiong.
\newblock Knowledge-driven cot: Exploring faithful reasoning in llms for knowledge-intensive question answering, 2023{\natexlab{b}}.

\bibitem[Wang et~al.(2023{\natexlab{c}})Wang, Wei, Schuurmans, Le, Chi, Narang, Chowdhery, and Zhou]{self_consistency}
Xuezhi Wang, Jason Wei, Dale Schuurmans, Quoc~V. Le, Ed~H. Chi, Sharan Narang, Aakanksha Chowdhery, and Denny Zhou.
\newblock Self-consistency improves chain of thought reasoning in language models.
\newblock In \emph{The Eleventh International Conference on Learning Representations, {ICLR} 2023, Kigali, Rwanda, May 1-5, 2023}. OpenReview.net, 2023{\natexlab{c}}.
\newblock URL \url{https://openreview.net/pdf?id=1PL1NIMMrw}.

\bibitem[Wei et~al.(2022)Wei, Wang, Schuurmans, Bosma, Chi, Le, and Zhou]{COT}
Jason Wei, Xuezhi Wang, Dale Schuurmans, Maarten Bosma, Ed~H. Chi, Quoc Le, and Denny Zhou.
\newblock Chain of thought prompting elicits reasoning in large language models.
\newblock \emph{arXiv Preprint}, 2022.
\newblock URL \url{https://arxiv.org/abs/2201.11903}.

\bibitem[Xie et~al.(2022)Xie, Wu, Shi, Zhong, Scholak, Yasunaga, Wu, Zhong, Yin, Wang, Zhong, Wang, Li, Boyle, Ni, Yao, Radev, Xiong, Kong, Zhang, Smith, Zettlemoyer, and Yu]{skg}
Tianbao Xie, Chen~Henry Wu, Peng Shi, Ruiqi Zhong, Torsten Scholak, Michihiro Yasunaga, Chien-Sheng Wu, Ming Zhong, Pengcheng Yin, Sida~I. Wang, Victor Zhong, Bailin Wang, Chengzu Li, Connor Boyle, Ansong Ni, Ziyu Yao, Dragomir Radev, Caiming Xiong, Lingpeng Kong, Rui Zhang, Noah~A. Smith, Luke Zettlemoyer, and Tao Yu.
\newblock {U}nified{SKG}: Unifying and multi-tasking structured knowledge grounding with text-to-text language models.
\newblock In \emph{Proceedings of the 2022 Conference on Empirical Methods in Natural Language Processing}, pp.\  602--631, Abu Dhabi, United Arab Emirates, December 2022. Association for Computational Linguistics.
\newblock URL \url{https://aclanthology.org/2022.emnlp-main.39}.

\bibitem[Xie et~al.(2023)Xie, Kawaguchi, Zhao, Zhao, Kan, He, and Xie]{self_eval}
Yuxi Xie, Kenji Kawaguchi, Yiran Zhao, Xu~Zhao, Min-Yen Kan, Junxian He, and Qizhe Xie.
\newblock Decomposition enhances reasoning via self-evaluation guided decoding, 2023.

\bibitem[Yang et~al.(2023)Yang, Chen, Li, Ding, and Wu]{yang2023chatgpt}
Linyao Yang, Hongyang Chen, Zhao Li, Xiao Ding, and Xindong Wu.
\newblock Chatgpt is not enough: Enhancing large language models with knowledge graphs for fact-aware language modeling, 2023.

\bibitem[Yao et~al.(2022)Yao, Zhao, Yu, Du, Shafran, Narasimhan, and Cao]{yao2022react}
Shunyu Yao, Jeffrey Zhao, Dian Yu, Nan Du, Izhak Shafran, Karthik Narasimhan, and Yuan Cao.
\newblock React: Synergizing reasoning and acting in language models.
\newblock \emph{arXiv preprint arXiv:2210.03629}, 2022.

\bibitem[Yao et~al.(2023)Yao, Yu, Zhao, Shafran, Griffiths, Cao, and Narasimhan]{tot}
Shunyu Yao, Dian Yu, Jeffrey Zhao, Izhak Shafran, Thomas~L. Griffiths, Yuan Cao, and Karthik Narasimhan.
\newblock Tree of thoughts: Deliberate problem solving with large language models, 2023.

\bibitem[Yih et~al.(2016)Yih, Richardson, Meek, Chang, and Suh]{webqsp_dataset}
Wen{-}tau Yih, Matthew Richardson, Christopher Meek, Ming{-}Wei Chang, and Jina Suh.
\newblock The value of semantic parse labeling for knowledge base question answering.
\newblock In \emph{Proceedings of the 54th Annual Meeting of the Association for Computational Linguistics, {ACL} 2016, August 7-12, 2016, Berlin, Germany, Volume 2: Short Papers}. The Association for Computer Linguistics, 2016.
\newblock \doi{10.18653/v1/p16-2033}.
\newblock URL \url{https://doi.org/10.18653/v1/p16-2033}.

\bibitem[Yu et~al.(2023)Yu, Zhang, Ng, Zhu, Li, Wang, Hu, Wang, Wang, and Xiang]{webqsp1}
Donghan Yu, Sheng Zhang, Patrick Ng, Henghui Zhu, Alexander~Hanbo Li, Jun Wang, Yiqun Hu, William Wang, Zhiguo Wang, and Bing Xiang.
\newblock Decaf: Joint decoding of answers and logical forms for question answering over knowledge bases, 2023.

\bibitem[Yu et~al.(2022)Yu, Zhu, Zhang, Wang, Zhang, Fang, and Jiang]{creak1}
Wenhao Yu, Chenguang Zhu, Zhihan Zhang, Shuohang Wang, Zhuosheng Zhang, Yuwei Fang, and Meng Jiang.
\newblock Retrieval augmentation for commonsense reasoning: A unified approach.
\newblock In \emph{Proceedings of the 2022 Conference on Empirical Methods in Natural Language Processing}, pp.\  4364--4377, Abu Dhabi, United Arab Emirates, December 2022. Association for Computational Linguistics.
\newblock \doi{10.18653/v1/2022.emnlp-main.294}.
\newblock URL \url{https://aclanthology.org/2022.emnlp-main.294}.

\bibitem[Zhang et~al.(2021)Zhang, Gong, Shen, Li, Lv, Duan, and Chen]{zhang2021poolingformer}
Hang Zhang, Yeyun Gong, Yelong Shen, Weisheng Li, Jiancheng Lv, Nan Duan, and Weizhu Chen.
\newblock Poolingformer: Long document modeling with pooling attention.
\newblock In \emph{International Conference on Machine Learning}, pp.\  12437--12446. PMLR, 2021.

\bibitem[Zhang et~al.(2023)Zhang, Gong, He, Liu, Guo, Lv, and Guo]{zhang2023noisy}
Hang Zhang, Yeyun Gong, Xingwei He, Dayiheng Liu, Daya Guo, Jiancheng Lv, and Jian Guo.
\newblock Noisy pair corrector for dense retrieval.
\newblock \emph{arXiv preprint arXiv:2311.03798}, 2023.

\bibitem[Zhang et~al.(2022)Zhang, Zhang, Li, and Smola]{auto}
Zhuosheng Zhang, Aston Zhang, Mu~Li, and Alex Smola.
\newblock Automatic chain of thought prompting in large language models.
\newblock \emph{arXiv Preprint}, 2022.
\newblock \doi{10.48550/arXiv.2210.03493}.
\newblock URL \url{https://doi.org/10.48550/arXiv.2210.03493}.

\end{thebibliography}
